\documentclass[10pt]{article} 
\usepackage[accepted]{tmlr}

\usepackage{amsmath,amsfonts,bm}

\def\eqref#1{equation~\ref{#1}}

\def\1{\bm{1}}

\DeclareMathAlphabet{\mathsfit}{\encodingdefault}{\sfdefault}{m}{sl}
\SetMathAlphabet{\mathsfit}{bold}{\encodingdefault}{\sfdefault}{bx}{n}

\usepackage{hyperref}
\usepackage{url}
\usepackage{graphicx}
\usepackage{booktabs}
\usepackage{enumitem}
\usepackage{multirow}
\usepackage{tcolorbox}
\usepackage{makecell}
\usepackage{subcaption}
\usepackage{array}
\newcolumntype{C}[1]{>{\centering\arraybackslash}p{#1}}

\usepackage{xcolor}
\usepackage{soul}
\usepackage{calc}
\usepackage{colortbl} 

\if
\newcommand{\colorcell}[1]{%
    \ifdim #1pt < -0.2pt
        \cellcolor[HTML]{80B88F} #1 
    \else\ifdim #1pt < -0.1pt
        \cellcolor[HTML]{A0C9A5} #1 
    \else\ifdim #1pt < -0.05pt
        \cellcolor[HTML]{C4DBC7} #1 
    \else\ifdim #1pt < 0pt
        \cellcolor[HTML]{E0EAE0} #1 
    \else\ifdim #1pt = 0pt
        \cellcolor[HTML]{FFFFFF} #1 
    \else\ifdim #1pt < 0.05pt
        \cellcolor[HTML]{FAE0E0} #1 
    \else\ifdim #1pt < 0.15pt
        \cellcolor[HTML]{F3A8A8} #1 
    \else\ifdim #1pt < 0.2pt
        \cellcolor[HTML]{E08080} #1 
    \else
        \cellcolor[HTML]{D04040} #1 
    \fi\fi\fi\fi\fi\fi\fi\fi
}
\fi
\newcommand{\colorcell}[1]{%
    \ifdim #1pt < -0.2pt
        \cellcolor[HTML]{2166AC}\textcolor{white}{#1}%
    \else\ifdim #1pt < -0.1pt
        \cellcolor[HTML]{4393C3} #1%
    \else\ifdim #1pt < -0.05pt
        \cellcolor[HTML]{92C5DE} #1%
    \else\ifdim #1pt < 0pt
        \cellcolor[HTML]{D1E5F0} #1%
    \else\ifdim #1pt = 0pt
        \cellcolor[HTML]{F7F7F7} #1%
    \else\ifdim #1pt < 0.05pt
        \cellcolor[HTML]{FDDBC7} #1%
    \else\ifdim #1pt < 0.1pt
        \cellcolor[HTML]{F4A582} #1%
    \else\ifdim #1pt < 0.2pt
        \cellcolor[HTML]{D6604D} #1%
    \else
        \cellcolor[HTML]{B2182B}\textcolor{white}{#1}%
    \fi\fi\fi\fi\fi\fi\fi\fi
}

\newcommand{\dcell}[4][]{\makecell{$#2$#1\\[-2pt]{\scriptsize$[#3,\,#4]$}}}

\title{Mitigating Social Desirability Bias in Random Silicon \\Sampling}

\author{\name Sashank Chapala$^*$ \email s.chapala@student.tue.nl \\
      Eindhoven University of Technology, The Netherlands
      \AND
      \name Maksym Mironov$^*$ \email m.mironov@student.tue.nl \\
      Eindhoven University of Technology, The Netherlands
      \AND
      \name Songgaojun Deng$^{\dagger}$  \email s.deng@tue.nl\\
      Eindhoven University of Technology, The Netherlands\\
      $^{*}$Equal contribution \quad $^{\dagger}$Corresponding author
      }

\def\month{09}  
\def\year{2026} 
\def\openreview{\url{https://openreview.net/forum?id=1njhuA8B3r}} 

\begin{document}

\maketitle

\begin{abstract}
Large Language Models (LLMs) are increasingly used to simulate population responses, a method known as ``Silicon Sampling''. 
However, responses to socially sensitive questions frequently exhibit social desirability bias, diverging from real human data toward socially desirable answers. Existing studies on social desirability bias in LLM-based sampling remain limited. In this work, we investigate whether minimal, psychologically grounded prompt wording changes can mitigate this bias and improve alignment between silicon and human samples.
We conduct a study using data from the American National Election Studies (ANES) on three LLMs spanning the Llama and GPT model families.
We first re-implement a baseline silicon sampling study, confirming persistent social desirability bias. 
We then test four prompt-based mitigation methods: \emph{reformulation} (neutral, third-person phrasing), \emph{reverse-coding} (semantic inversion), and two meta-instructions, \emph{priming} and \emph{preamble}, respectively encouraging analytical reasoning and sincerity.
We evaluate alignment with ANES using Jensen-Shannon divergence, and introduce a signed desirability gap that measures the direction of the bias, distinguishing reduction of social desirability bias from generic distributional changes.

Results show that reformulated prompts improve alignment, reducing concentration on socially desirable answers and yielding distributions closer to ANES for two of the three LLMs. 
Reverse-coding produced mixed results across eligible items, while Priming and Preamble showed no systematic benefit for bias mitigation.
Treating items as the unit of analysis, most aggregate effects do not generalize beyond our item set. The exception is an ablation applying third-person framing alone, which reliably reduces both metrics for Llama-3.1-8B and the desirability gap for GPT-4.1-mini.
We further show that reformulation performs similarly across survey waves and test it on a different survey covering three Western populations outside the U.S. 
Our findings support the use of prompt-based framing controls to mitigate social desirability bias in LLMs, providing a practical way to bring silicon samples closer to the human benchmark on the items where the bias is present.
\end{abstract}

\section{Introduction}

Large Language Models (LLMs)~\citep{radford2018improving,kojima2022large} can simulate human emotions and opinions, from subjective labeling of Twitter posts \citep{törnberg2023,yang2024oasis} and participation in psychological studies \citep{aher2023,qiu2024interactive} to producing behavior changes consistent with personality frameworks \citep{serapio2023personality,besta2025psychologicallyenhancedaiagents}.
These non-trivial capabilities naturally led researchers to explore whether LLMs can be used to simulate entire populations for social sciences~\citep{Argyle2023,yang2024oasis,gao2023s3}, polling~\citep{yu2024towards,zhang2024electionsim}, or marketing research studies \citep{sarstedt2024using,arora2025ai}. Using LLM-simulated respondents in these setups could help tackle the limitations of large sample sizes, high costs, and long execution times associated with human respondents. This led to the idea of \textit{silicon sampling}, which refers to the use of LLM-generated agents with demographic conditioning to simulate population-level survey responses \citep{Argyle2023, sun2024}.

Prior work found remarkable alignment between silicon and human samples on some topics. However, they diverged more sharply when sensitive topics or groups were involved \citep{sun2024}. This divergence likely reflects \emph{Social Desirability Bias (SDB)}, i.e., the tendency of LLMs to generate socially approved rather than demographically representative answers \citep{salecha2024large}.

To make silicon sampling a viable method, reliability across topics is of key importance, which makes investigating this divergence an important research direction. We believe that the social desirability bias may largely stem from the otherwise benign decision of model developers to suppress humans' harmful biases and stereotypes learned by the model, a finding that is repeatedly confirmed by research that finds almost no explicit stereotypes in responses of different models \citep{liang2021towards,lin2024towards,bai2025explicitly, zhao2025explicit, li2025actions}. When performing silicon sampling, however, it is desirable to elicit a model's knowledge about stereotypes held by different groups of people to yield more accurate results. 
Although models have been trained to avoid expressing any \textit{explicit} stereotypes, prior work shows that \textit{implicit} stereotypes can still emerge when socially sensitive content is elicited more indirectly~\citep{bai2025explicitly,zhao2025explicit}. This suggests that prompt design~\citep{white2023prompt} impacts model responses and may help reduce social desirability bias, yielding more representative responses. 
We therefore hypothesize that it is possible to reduce the social desirability bias by using other methods drawn from previous LLM and psychological research. 

The main goal of this study lies in the systematic exploration of prompt engineering techniques for reducing the social desirability bias in large population silicon sampling~\citep{sun2024}. We test four prompt design strategies to evaluate their effectiveness in mitigating bias and bringing a model's responses closer to human populations. This way, we hope to provide future silicon sampling studies with a method of achieving closer alignment with human samples, particularly on sensitive topics.
The main contributions of this work are:
\begin{itemize}[noitemsep, topsep=0pt]
\item We present a systematic evaluation of prompt-based mitigation strategies for social desirability bias in large-scale silicon sampling, re-implementing prior silicon-sampling results and extending them with four prompt conditions across three instruction-tuned LLMs.
\item We introduce a signed desirability gap alongside Jensen-Shannon divergence, and a joint criterion that counts a change as bias mitigation only when divergence and the gap both fall on an item whose baseline gap is positive. This separates reduction of social desirability bias from distributional changes that improve alignment for unrelated reasons.
\item We find that item reformulation improves distributional alignment and narrows the desirability gap for two of the three models, while reverse-coding, priming, and preamble yield limited or inconsistent benefits. Treating items as the unit of analysis, most of these aggregate effects do not generalize beyond the items we study.
\item We isolate third-person framing from the other edits reformulation makes, and find that framing alone reduces both metrics reliably across items for Llama-3.1-8B and the desirability gap for GPT-4.1-mini, where the full reformulation does not.
\item We evaluate reformulation across survey waves (ANES 2020 and 2024) and on a different survey covering three non-U.S. Western populations (WVS Wave 7), and show that it transfers on a subset of items rather than uniformly.
\item We report a demographic conditioning analysis showing that the profile shapes the response distribution primarily through the two political attributes.
\end{itemize}

\section{Related Work}

\textbf{Silicon sampling and synthetic populations.}
Researchers have examined population-level simulations using LLMs, an approach known as \emph{silicon sampling} \citep{Argyle2023, lee2024can,sarstedt2024using}. 
\citet{lee2024can} investigated the algorithmic fidelity and biases of LLMs in simulating public opinions regarding global warming.
Instead,
\citet{sun2024} used demographic distributions from the American National Election Studies (ANES) to construct synthetic respondents and compared their political survey responses with those of real participants.
While there was significant alignment between silicon and human samples on objective or politically neutral items, a larger disagreement was observed for socially sensitive questions, especially about race diversity, gender equality, or identity politics \citep{sun2024}. The results indicate that LLMs may tend to produce responses that align with socially desirable or politically neutral positions rather than the human opinions in real populations.

\textbf{Bias and social desirability in LLMs.}\label{sec:related-bias}
There is a broader body of research on bias in modern LLMs~\citep{schramowski2022large,gallegos2024bias} that provides insight on diverging answers in silicon sampling when dealing with sensitive subjects. Many LLMs undergo the process of alignment training, in which the models are fine-tuned to follow human values and avoid harmful or biased outputs \citep{bai2025explicitly, zhao2025explicit, li2025actions}. 
While this process reduces explicit bias, implicit biases can remain embedded in internal representations and emerge only under indirect or carefully crafted queries. 
%
Prior work shows that prompting models with demographic personas can elicit stereotyped, essentializing portrayals rather than faithful representations of a group, an effect most pronounced for marginalized and intersectional groups~\citep{cheng2023marked}.
Conversely, models can also move toward the opposite extreme, exhibiting social desirability bias by producing socially approved rather than statistically representative answers~\citep{salecha2024large}.
It is further revealed that the extent of social desirability bias depends on the context. In particular, models show stronger social desirability bias in the context of being ``judged'', mirroring findings from human psychology, where respondents offer more socially desirable answers when under the impression of being evaluated \citep{crowne1960new, paulhus1991enhancement}.

\textbf{Mitigation strategies and cognitive framing.}
Research in Psychology and Computer Science suggests approaches to reduce social desirability bias through changes in framing and context. In human respondents, bias can be mitigated by rephrasing evaluative questions into neutral or third-person formats, and by assuring anonymity and non-judgment \citep{fisher1993social}. Similarly, in LLMs, prompt conditioning and framing can influence the response mode of the model \citep{park2023generative,besta2025psychologicallyenhancedaiagents}. 
For example, by pre-conditioning models to adopt a more analytical persona, researchers have been able to produce more consistency in models' strategies in a game setting, with less adaptation to social circumstances. The results were also more representative of the real-world answers and less constrained by social expectations \citep{besta2025psychologicallyenhancedaiagents}.
Beyond prompt framing, human survey methodology offers further corrections (measurement-based control via desirability scales and post-hoc statistical adjustment against population benchmarks) that do not transfer cleanly to LLMs.
Treating the LLM as an unsupervised respondent without such calibration signals, we focus on prompt-level interventions and detail these alternatives, alongside the broader bias-mitigation landscape, in Appendix~\ref{sec:additional-related-work}.

\textbf{Summary and research gap.}
Prior work shows that LLMs can simulate human-like personas and that silicon sampling can approximate population-level survey responses. 
However, responses to socially sensitive questions often diverge from empirical data due to social desirability bias.
Although alignment training helps reduce explicit bias, it may also suppress implicit biases that contribute to realistic human behavior in simulations. 
Despite this, little is known about how prompt framing, evaluation context, or cognitive conditioning influences SDB in LLM-based population sampling.
We address this gap by systematically evaluating psychologically grounded prompt interventions, such as  third-person framing, neutral phrasing, and rational-mode prompting, and their effects on alignment between silicon and human responses.

\section{Methodology}\label{sec:methods}
\begin{figure}[t]
  \centering
\includegraphics[width=\linewidth]{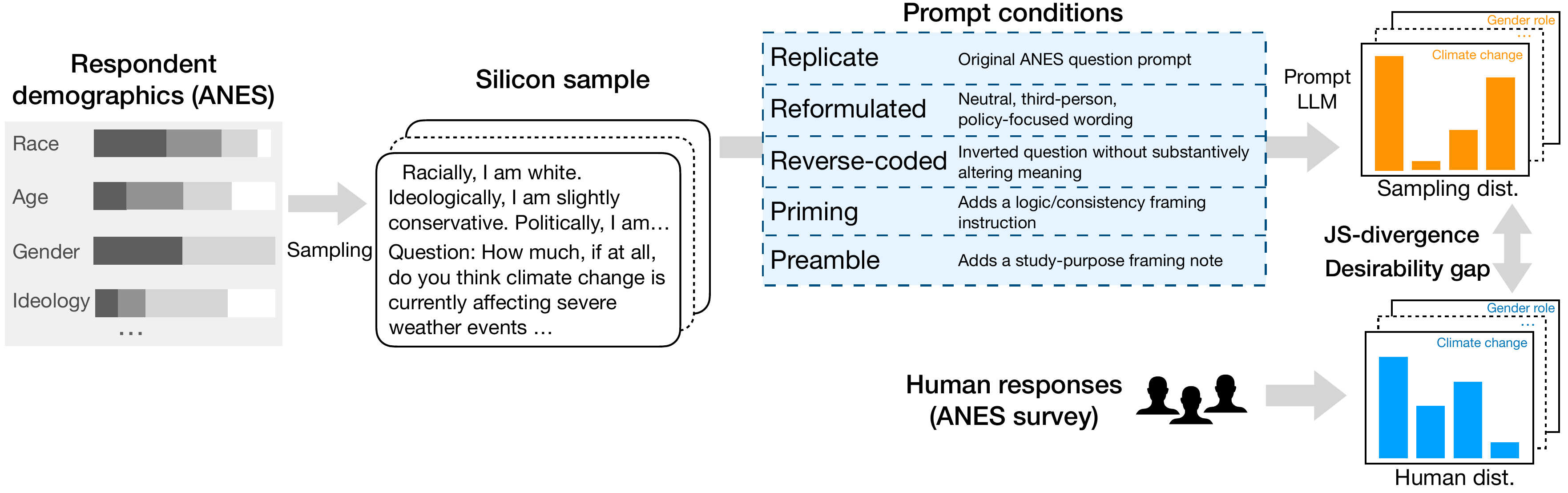}\vspace{-5pt}
\caption{Overview of the experimental pipeline. We draw a random ``silicon sample'' of personas from the ANES respondent demographic distribution and query an LLM under five prompt conditions, then compare the resulting model response distribution (orange) against real human ANES responses (blue) via per-item Jensen-Shannon divergence and the signed desirability gap. The Replicate condition follows the item wording of \citet{sun2024}.}
  \label{fig:pipeline}
\end{figure}

Our methodology follows a four-stage pipeline: (1) extracting demographic distributions from a reference survey dataset; (2) generating a synthetic (silicon) population by conditioning LLMs on these demographic profiles; (3) collecting survey responses using zero-shot prompting under five prompt conditions; and (4) evaluating alignment between silicon and human responses using divergence-based metrics. The experimental pipeline is illustrated in \autoref{fig:pipeline}.
Data, code, and results are available at \url{https://github.com/songgaojundeng/mitigating-social-desirability-bias}.

\subsection{Data and sampling}\label{sec:data-sample}
\textbf{Survey data.} We mainly use the American National Election Studies (ANES) 2020 pre-election survey dataset, which includes 5,441 respondents following a prior study~\citep{sun2024}.\footnote{\url{https://electionstudies.org/data-center/2020-time-series-study/}} The dataset contains both demographic information and responses to a wide range of political and social questions. We focus on ten multiple-choice questions/items covering different social and political topics, including \textit{Race Diversity}, \textit{Gender Role}, \textit{Current Economy}, \textit{Drug Addiction}, \textit{Climate Change}, \textit{Gay Marriage}, \textit{Refugee Allowing}, \textit{Health Insurance}, \textit{Gun Regulation}, and \textit{Income Inequality}.
We additionally use the ANES 2024 pre-election survey for temporal consistency analysis and World Values Survey (WVS) Wave~7 data for European countries to test effectiveness beyond ANES; details are given in the following sections.

\textbf{Answer-option categorization.} For each question, we classify its discrete response options into three types according to social-desirability valence, following the survey-methodology view that response options differ in their perceived social desirability~\citep{stocke2004interdependence,tourangeau2007sensitive}. A \textit{socially desirable (acceptable)} option expresses support for a position generally framed as socially inclusive or supportive in the survey context; a \textit{socially undesirable} option expresses the more restrictive or opposing position; and a \textit{safe (neutral or moderate)} option avoids strong judgment or policy change, typically a status-quo or middle response. 
For instance, the Health Insurance item asks:
\begin{quote}
    Do you favor an increase, decrease, or no change in government spending to help people pay for health insurance when people cannot pay for it all themselves? 
    
    1. Increase 2. Decrease 3. No change
\end{quote}
Here, \textit{Increase} is the socially desirable option, \textit{Decrease} the socially undesirable option, and \textit{No change} the safe option. 
The full set of items with their response options is given in Appendices (\ref{sec:all-questions} and \ref{sec:all-questions-wvs}), where the most socially desirable option for each item is underlined. 
The annotation procedure is detailed in Appendix~\ref{app:annotation}: 9 of the 10 ANES 2020 items and 8 of the 10 WVS items were classified as directional.\footnote{These labels do not reflect a universal moral standard. We interpret them as indicators of population-level social pressure rather than as claims about the values of every subgroup.}
Because the socially desirable option may coincide with the politically progressive one, we additionally measure each item's left-right difference from the human data (Appendix~\ref{sec:politics-analysis}). The desirable pole aligns with the more left-leaning response on all directional items, but the strength varies widely. Most ANES items divide left and right respondents strongly, whereas \textit{Drug Addiction} and \textit{Gender Role} divide them only weakly, and the WVS items are consistently less divided.

\textbf{Silicon sampling.} To generate synthetic respondents, we follow the random silicon sampling procedure in \citet{sun2024}. Let $\mathcal{D}_{\text{ANES}}$ denote the empirical ANES population. From $\mathcal{D}_{\text{ANES}}$, we estimate empirical marginal distributions over $K=8$ demographic variables $\{A_1, A_2,\ldots, A_K\}$, corresponding to \textit{race}, \textit{age}, \textit{gender}, \textit{ideology}, \textit{party identification}, \textit{church attendance}, \textit{political interest}, and \textit{political discussion frequency}. 
We then construct a silicon sample $\{R_i\}_{i=1}^{N}$ with $N=5{,}441$, matching the size of the human dataset. Each synthetic respondent $R_i$ is defined by a demographic profile sampled independently from these empirical marginals:
\begin{equation}\label{eq:sampling}
R_i = \{a^{(i)}_k \sim A_k\}_{k=1}^K.
\end{equation}
The demographic attributes of each $R_i$ are rendered in natural language and provided to the language model as conditioning context, together with a survey question. Answer options are presented in a fixed order across all experiments. Because option positions are identical for matched items, any position bias~\citep{zheng2024large} is largely constant across conditions and is unlikely to confound our comparisons. The language model then generates a response for that synthetic individual. Repeating this procedure across all $R_i$ yields an aggregate response distribution from random silicon subjects.\footnote{Sampling each $A_k$ independently does not preserve empirical dependencies among attributes (e.g., age-party-ideology), so some profiles may be rare or implausible. We adopt this design to obtain broad, systematic coverage of the demographic space for probing model responses in a controlled manner. \citet{sun2024} further show that random silicon sampling reproduces the aggregate ANES distribution more closely than full joint replication (their Table~1). Nevertheless, sampling that better preserves correlations is a promising direction we consider for future work.}

All comparisons in this study are made at the level of the aggregate distribution rather than the individual respondent $R_i$. We evaluate how closely the aggregate silicon distribution matches the \textit{marginal} human distribution, and do not claim that any single profile predicts a real person's response. Because profiles are sampled independently from the empirical marginals, the same profile may be drawn more than once. We find that such duplication is rare and does not meaningfully affect the aggregate results (Appendix~\ref{sec:duplication}).
Details of demographic variables are in Appendix~\ref{sec:demographic-vars}.

\textbf{Estimand and data treatment.} We compare the responses against the unweighted ANES 2020 respondent distribution and do not apply the survey weights. Silicon profiles are sampled from the marginals of that same unweighted data. Our results are therefore about the ANES respondent sample rather than the U.S. adult population.
Human response distributions are computed over the substantive answer options only, with non-substantive codes excluded per item rather than per respondent. The same treatment is applied to the demographic variables used for sampling. Human cells are therefore roughly 10\% smaller than the corresponding silicon cells, ranging from 0.3\% to 13.7\%
across items. Appendix~\ref{sec:non-response} gives the full treatment, and Appendix~\ref{app:non-response-analysis} reports additional analysis.
Each run draws its own profile sample, so a small part of any difference between runs reflects sampling rather than the condition being compared. We report a single run covering all respondents and items, and do not repeat it under multiple seeds.\footnote{Redrawing the profile sample gives a standard deviation of 0.0008 in the aggregate JS-divergence, measured in the demographic ablation of Section~\ref{sec:conditioning}.}

\subsection{Model selection}
Our analysis employs three instruction-tuned LLMs to compare performance across scale and licensing: the open-source Llama-3.1-8B-Instruct (Llama-8B) and Llama-3.1-70B-Instruct (Llama-70B) \citep{grattafiori2024llama},
and the closed-source GPT-4.1-mini (released on 2025-04-14).
All three are alignment- and instruction-tuned, consistent with our focus on the deployed systems practitioners use for silicon sampling rather than raw pretraining checkpoints; we discuss this scope decision and the role of pre-alignment base models in Limitations. 
The closed-source model is a substantial advance over the GPT-3.5-turbo model (released on 2023-06-13) used in the prior silicon sampling study~\citep{sun2024}. We selected GPT-4.1-mini over the GPT-5 family (released on 2025-08-07) for its optimal performance-cost balance and faster inference speed, making it better suited for high-throughput survey simulations. 
Together, the selected models allow us to evaluate the trade-off among model scales, capability, computational cost, and reproducibility.
All models were configured under stochastic decoding. More implementation details are provided in Appendix~\ref{sec:implementation-details}.

\subsection{Prompt-based mitigation strategies}
The core of silicon sampling involves generating responses by conditioning the LLM on unique demographic profiles derived from the ANES dataset. To mitigate SDB, we design five experimental conditions by systematically varying the question structure, language, and contextual instructions.

\paragraph{Replicate Condition (0)}
We re-implement the experimental setup of prior work \cite{sun2024}. We keep the prompts unchanged, and ask questions directly copied from the ANES dataset. However, we use different models and different evaluation metrics than the original study \citep{sun2024}, which thus constitutes a re-implementation of the pipeline. This condition is used as a baseline to compare other conditions, as well as to measure SDB prior to any interventions.
%

\paragraph{Reformulated Condition (1)}
This condition is motivated by evidence that LLMs produce more socially desirable answers when the context suggests they are being evaluated~\citep{salecha2024large}, and by survey research showing that neutral wording and indirect, third-person questioning reduce socially desirable responding in human respondents~\citep{fisher1993social}. Nevertheless, reduced evaluation awareness is only one of several possible mechanisms, and a further discussion is available in Section~\ref{sec:discussion-reform}. We apply several modifications to the phrasing of the survey questions while ensuring that their meaning remains unchanged. 

\begin{itemize}[noitemsep, topsep=0pt]
    \item For all 10 items in the ANES dataset, we re-formulate the questions to avoid direct ``what do you think'' phrasing, using a more neutral third-person formulation (e.g., ``what would this respondent think'').
    \item In 4 ANES items (\textit{Current Economy}, \textit{Race Diversity}, \textit{Refugee Allowing}, and \textit{Income Inequality}), we neutralize the questions by using less evaluative language (e.g., avoiding evaluative words like ``good'', ``positive'', ``bad'').
    \item For the \textit{Refugee Allowing} item, the question is re-formulated to ask about one’s opinion about a policy rather than a social phenomenon or a group.
\end{itemize}
    
For several items this intervention also entails relabeling answer options, so the condition combines changes to question framing and to answer option wording.

An example of a typical prompt under Reformulated Condition: 
\begin{quote}
\textit{Racially, the respondent is black. The respondent doesn't like to discuss politics with their family and friends. Ideologically, the respondent is slightly liberal. Politically, the respondent is an independent. The respondent does not attend church. The respondent is 43 years old. The respondent is a man. The respondent is somewhat interested in politics}. 

\textit{How would this respondent assess if there should be an increase, decrease, or no change in government spending to help people pay for health insurance when people can't pay for it all themselves? }

\textit{1. Increase 2. Decrease 3. No change}

\end{quote}
\paragraph{Reverse-coded Condition (2)} 
Prior work found that reverse-coding of questions can reduce SDB~\citep{salecha2024large}. To test whether it can alleviate social desirability on sensitive topics, we include reverse-coded versions \emph{where this does not substantively alter the item's semantic meaning}. For example,
\begin{quote}
    Original: How much, if at all, do you think climate change is currently affecting severe weather events or temperature patterns in the United States?

    \textit{Reverse-coded: How much do you agree with the statement that climate change is unrelated to severe weather events or temperature patterns in the United States?}
\end{quote}
For reverse-coded questions, the resulting answer options were inverted numerically to ensure consistency in the final distribution analysis. Items where reverse-coding alters the construct are omitted.

\paragraph{Priming Condition (3)}
This condition is grounded in studies showing that psychological priming of LLMs can influence behavior~\citep{besta2025psychologicallyenhancedaiagents}. We use the conditioning prompts to induce more of a ``Thinking'' agent, which is added before the survey question. The verbatim instructions are:
\begin{quote}
You value logic, objectivity, and internal consistency. When responding, you prioritize reasoning over emotion, and aim to base your answers on evidence, structure, and rational analysis. Your goal is to provide clear, well-reasoned, and intellectually honest answers that reflect careful thought. The best answer is that which reflects your reasoning process.
\end{quote}

\paragraph{Preamble Condition (4)} Adding a preamble encouraging sincere answers and promising no judgment is standard in studies on sensitive topics as it reduces pressure to conform among human respondents. 
We use the following verbatim preamble:
\begin{quote}
In this study, we are exploring how people naturally respond to various questions. To ensure meaningful results, it is important that responses reflect your genuine thoughts and feelings. There are no correct or desirable answers, and your responses will not be evaluated or judged. Please answer honestly and without concern for how your answers might be perceived. Your sincerity helps us better understand authentic human responses.
\end{quote}

\noindent For each condition, the same distribution of demographic variables was used, and each new response was produced with a fresh and isolated LLM session to preclude any memory or carry-over of context. 
To standardize data collection, all demographic attributes were provided in the system prompt, while the user prompt contained the survey question, listed the discrete answer options, and explicitly instructed the model to \textit{respond with a single number only}. For the Priming and Preamble conditions, the corresponding text was added at the beginning of the system prompt.
All questions and prompt conditions, including reformulated and reverse-coded variants are in Appendix~\ref{sec:all-questions}. LLM prompting examples are in Appendix~\ref{sec:prompt-eg}.

\subsection{Evaluation}\label{sec:evaluation}
\paragraph{Jensen-Shannon divergence (JS-divergence).}

To quantify alignment between LLM-generated (``silicon'') and human survey responses, we compute the JS-divergence between the empirical human distribution from ANES and the corresponding silicon distribution. JS-divergence is derived from the Kullback-Leibler divergence (KL-divergence), defined as:
\begin{equation}
    \mathrm{JSD}(P \Vert  Q) = \frac{1}{2}\mathrm{KLD}(P \Vert  M) + \frac{1}{2}\mathrm{KLD}(Q \Vert  M),
\end{equation}
\begin{equation}
\mathrm{KLD}(P \Vert M) = \sum_{o \in \mathcal{O}} P(o) \log \frac{P(o)}{M(o)},
\end{equation}

where $P$ and $Q$ are the human and silicon response distributions over answer options $o \in \mathcal{O}$ and $M = \tfrac{1}{2}(P + Q)$. Lower values indicate closer alignment. We adopt JS-divergence over KL-divergence or a Chi-squared test for three reasons. 
It is \emph{symmetric}, suiting a comparison of two samples rather than a comparison against a ``true'' reference. It is \emph{bounded} on $[0,1]$ (base-2 logarithms) and finite under non-overlapping support, whereas $\mathrm{KLD}(P \Vert Q)$ diverges whenever the silicon distribution assigns zero probability to a human-selected option, common for categorical items with finite samples.  Unlike a Chi-squared statistic \citep{sun2024}, which scales with sample size, it yields a normalized distance comparable across items and conditions. 

For each question $X$, we evaluate five experimental conditions $c \in \{0, 1, 2, 3, 4\}$,
with $c_0$ the Replicate baseline.
Let \( P_X \) denote the human response distribution for question \( X \), and \( Q_{c,X} \) the silicon response distribution under condition \( c \).
This yields five JS-divergence values per question:
\[
\mathrm{JSD}(X, c)=\mathrm{JSD}(P_X \Vert Q_{c,X}).
\]

\paragraph{Signed desirability gap.}
JS-divergence only indicates that the model’s choices differ from human choices. It does not show the direction of the difference, e.g., whether the model chooses the socially desirable option more or less often than humans do, which is the key prediction of SDB.
To capture the direction of the difference, we introduce a \emph{signed desirability gap}, computed using the desirability labels from Section~\ref{sec:data-sample} and the same distributions as JS-divergence.
For an item $X$ with options $\mathcal{O}$, we summarize each distribution by a desirability-weighted score $S(P)=\sum_{o\in\mathcal{O}} w_{o}\,P(o)$ and define the gap under condition $c$:
\begin{equation}\label{equ:signed_gap}
    D(X, c) = \frac{S(Q_{c,X}) - S(P_X)}{Z_X}.
\end{equation}
The weights $w_{o}$ encode only the direction of defined social desirability. 
For categorical items, whose options either have no natural order or are not coded monotonically from one pole to the other, $w_{o}=1$ for the desirable option and $0$ otherwise, so $S(\cdot)$ is the probability mass on that option and $Z_X=1$. For ordinal items, $w_{o}$ is based on the option's scale position, oriented to increase toward the desirable pole (higher values are more desirable). Appendix~\ref{app:weight-sensitivity} shows the ordinal results are largely robust to a binary alternative.
$S(\cdot)$ is then the mean response, and $Z_X=o_{\max}-o_{\min}$ normalizes by the scale range. In both cases $D(X,c)\in[-1,1]$. A positive gap means the silicon distribution is tilted more toward the desirable pole than the human one.
Under SDB, we therefore expect \textit{a positive gap at baseline}, $D(X,c_0)>0$; successful mitigation should then reduce it toward zero. Items with no desirable direction are excluded here but retained in the JS-divergence comparison.

Since the desirable pole aligns with the politically progressive response on most directional items, $D$ may capture social desirability and ideological priors jointly on strongly divided items. While the impact of ideological priors is not measured directly, previous work \citep{li2025actions, zhao2025explicit} shows that LLMs underwent alignment processes aimed at avoiding bias and providing responses similar to our classification of socially desirable when asked about race, ethnicity, and gender. 
This suggests that at least some part of the desirability gap is driven by a model's tendency to select socially appropriate responses in line with its post-training rather than being entirely attributable to ideology. 
Nevertheless, the weakly divided items of Appendix~\ref{sec:politics-analysis} provide comparatively cleaner tests.

We read each JS-divergence reduction together with the change in the desirability gap. For item $X$ and condition $c$ relative to the baseline $c_0$, we consider two changes jointly: the change in divergence, $\Delta\mathrm{JSD}_{c,X} = \mathrm{JSD}(X, c) - \mathrm{JSD}(X, c_0)$, and the change in the absolute gap, $\Delta|D|_{c,X} = |D(X, c)| - |D(X, c_0)|$, each defined so that negative values denote movement toward the human distribution. A negative $\Delta|D|$ means the desirability-weighted score has moved toward the human one, which corresponds to selecting the socially desirable option less often only when the baseline signed gap is \textit{positive}. Where the baseline gap is negative, the same movement means the model selects that option more often than at baseline, and we report such cases as calibration improvements rather than as SDB reduction. We therefore read a divergence reduction that coincides with a narrowing gap ($\Delta\mathrm{JSD}_{c,X} < 0$ and $\Delta|D|_{c,X} < 0$) on an item with $D(X,c_0) > 0$ as the model choosing the socially desirable answer less often, rather than its answers changing for some unrelated reason that improves alignment with human data. We refer to this as the joint criterion.

\paragraph{Uncertainty estimation and statistical comparison.}

To quantify uncertainty and test whether differences between conditions are statistically meaningful, we use a paired non-parametric bootstrap that resamples both the silicon and the human (ANES) responses. Both metrics are computed from the same resampled draws, so the divergence and the desirability gap are estimated on identical data at every replicate. For each question $X$ and replicate $j = 1, \dots, n$:

\begin{enumerate}[noitemsep, topsep=0pt]

    \item we resample the human responses with replacement to form $P_X^{(j)}$, shared across all conditions at replicate $j$;

    \item for each condition $c$, we resample its silicon responses to form $Q_{c,X}^{(j)}$;

    \item from this pair we compute the divergence $\mathrm{JSD}^{(j)}(X, c) = \mathrm{JSD}\!\big(P_X^{(j)} \Vert Q_{c,X}^{(j)}\big)$ and, for directional items, the desirability-weighted scores giving the signed gap $D^{(j)}(X, c)$.

\end{enumerate}

Resampling the human side incorporates ANES sampling error rather than treating the reference distribution as fixed, and sharing $P_X^{(j)}$ within a replicate correlates the paired terms and reduces the variance of their differences. We report the 95\% percentile interval of each condition's bootstrap distribution, and of the baseline gap $D(X, c_0)$, for description only. Point estimates in all tables are computed on the observed responses rather than as bootstrap means.
Significance is assessed using a two-sided bootstrap test on the paired differences relative to the baseline $c_0$, applied to the empirical distributions of $\Delta\mathrm{JSD}_{c,X}^{(j)} = \mathrm{JSD}^{(j)}(X, c) - \mathrm{JSD}^{(j)}(X, c_0)$ and of $\Delta|D|_{c,X}^{(j)}$. To account for multiple comparisons (e.g., 10 items $\times$ 4 conditions per model), we control the false discovery rate at $q = 0.05$ using the Benjamini-Hochberg procedure~\citep{benjamini1995controlling}, yielding adjusted $p$-values $p^{\mathrm{BH}}$. A comparison is significant when $p^{\mathrm{BH}} \le q$. The family is defined as all item-condition comparisons within a given model and dataset, with $\Delta\mathrm{JSD}$ and $\Delta|D|$ corrected in separate families. All bootstrap analyses use $n = 10{,}000$ replicates.

\section{Experimental Results}
\subsection{Results across experimental conditions} 

\begin{figure}[!t] 
    \centering
    \begin{subfigure}[t]{0.45\textwidth} 
        \centering
        \includegraphics[width=\linewidth]{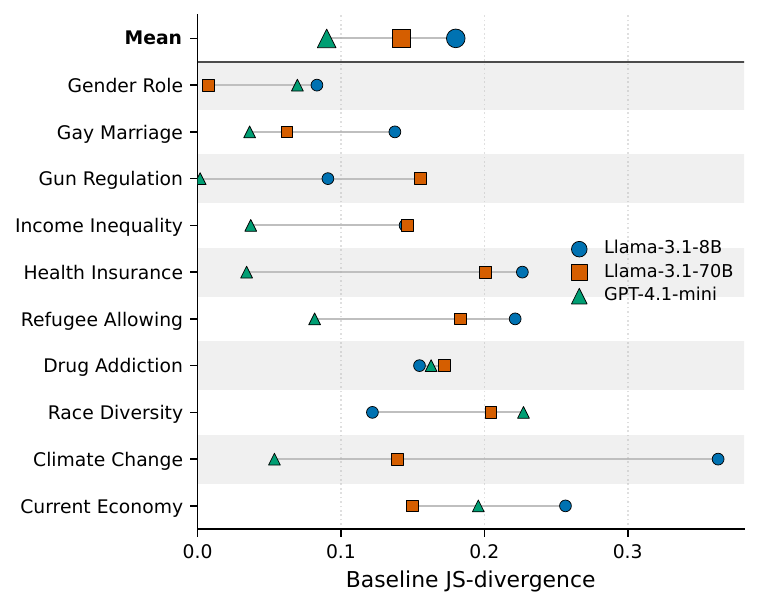}\vspace{-5pt}
    \caption{JS-divergence}
        \label{fig:baseline-jsd}
    \end{subfigure}\hfill
    \begin{subfigure}[t]{0.45\textwidth} 
        \centering
        \includegraphics[width=\linewidth]{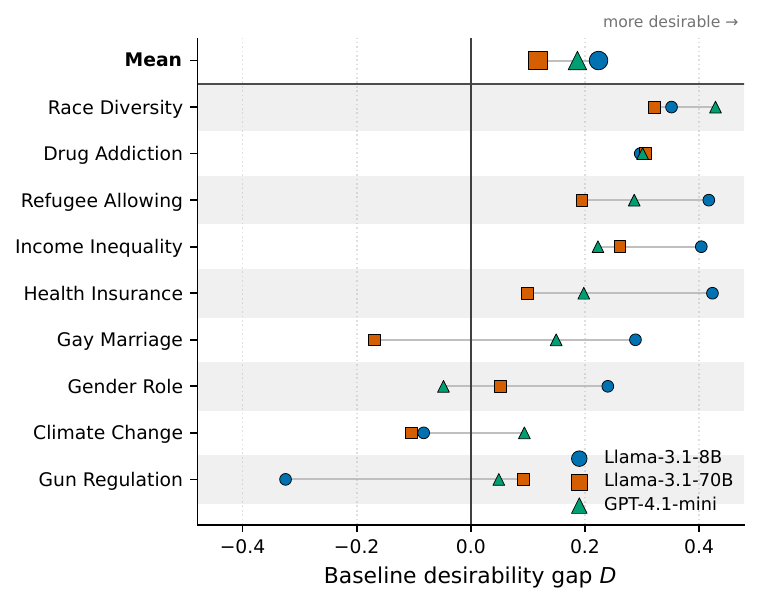}\vspace{-5pt}
    \caption{Signed desirability gap $D$}
        \label{fig:baseline-gap}
    \end{subfigure}
    \vspace{-5pt}
    \caption{Baseline (\emph{Replicate}) alignment with the human ANES~2020
distribution, per item and per model, under stochastic decoding.
\textbf{(a)} JS-divergence over all ten items; lower is closer.
\textbf{(b)} Signed desirability gap over the nine directional items (\textit{Current Economy} excluded).
$D>0$ marks a shift toward the socially desirable pole, and all three models
show a positive mean gap, consistent with social desirability bias.}
\label{fig:baseline}

\end{figure}

\paragraph{Replicate Condition (0)}
We re-implemented the base silicon-sampling pipeline of~\citet{sun2024} using our selected LLMs, conditioning each sample on standard demographic attributes and the original survey question. 
\autoref{tab:baseline} gives the model-level averages.
All three models show a positive mean desirability gap ($\overline{D}$) at baseline, placing more mass on the socially desirable pole than human respondents. 
This inclination toward the desirable option is the defining feature of social desirability bias, which the gap is built to measure.
It holds across models despite their differing alignment: GPT-4.1-mini is closest to ANES in JS-divergence
($\overline{\text{JSD}}=0.090$ over all ten items), followed by Llama-70B ($0.142$) and Llama-8B ($0.180$).

\autoref{fig:baseline} reports per-item and per-model alignment with ANES~2020: JS-divergence (panel a) and the signed desirability gap
$D$ (panel b). 
The models differ in how the bias appears. Llama-8B combines the largest mean signed
gap ($\overline{D}$) with the highest divergence, and its distributions concentrate heavily on the desirable option relative to the more dispersed human responses on sensitive items such as \textit{Refugee Allowing}, \textit{Health Insurance}, and
\textit{Gay Marriage}. 
Llama-70B has the smallest mean signed gap ($\overline{D}=0.117$), but this reflects cancellation rather than closer agreement: its items fall on both sides of zero (\autoref{fig:baseline-gap}), so positive and negative deviations offset in the mean. The absolute gap is larger ($\overline{|D|}=0.178$): strongly positive on \textit{Race Diversity} and \textit{Drug Addiction}, negative on \textit{Gay Marriage}, near zero on \textit{Gender Role}, \textit{Climate Change}, and \textit{Gun Regulation}.
This mix of opposing deviations suggests a polarized response pattern, with mass on the extremes rather than the moderate options humans more often choose.
GPT-4.1-mini has the tightest alignment overall but the second-largest gap, with signed and absolute means close ($\overline{D}=0.187$, $\overline{|D|}=0.198$). 
Its shift toward the desirable pole ($D$) is positive on all items except \textit{Gender Role}, producing distributions that are narrower and more concentrated on the desirable option than ANES.

On average, all three models have a positive baseline gap, over-representing socially desirable responses relative to humans.
Llama-8B and GPT-4.1-mini concentrate on the desirable option, while Llama-70B splits mass across both the desirable and undesirable options. 
Model architecture, parameter scale, and instruction-tuning can all drive these behavioral variations, stemming from the different models' underlying representation of population knowledge and reasoning capabilities.
Appendix \ref{sec:full-distr-replicate} provides a more detailed per-model discussion with the full response distributions. These patterns motivate the mitigation strategies evaluated in the following sections.

 \begin{table}[t]
\centering
\caption{Baseline ({Replicate}) alignment and desirability, averaged over the
nine directional items for each model. $\overline{\text{JSD}}$ is the mean
JS-divergence from the human distribution. $\overline{D}$ and $\overline{|D|}$ are
the mean signed and absolute desirability gaps, 
where $D > 0$ indicates that the model places more mass on the socially desirable pole than human respondents do.
Averaged over all ten items (adding the non-directional \textit{Current Economy} item),
$\overline{\text{JSD}}$ is $0.180$, $0.142$, and $0.090$ respectively.}
\label{tab:baseline}
\begin{tabular}{lccc}
\toprule
Model & $\overline{\text{JSD}}$ & $\overline{D}$ & $\overline{|D|}$ \\
\midrule
Llama-3.1-8B  & $+0.172$ & $+0.224$ & $+0.314$ \\
Llama-3.1-70B & $+0.141$ & $+0.117$ & $+0.178$ \\
GPT-4.1-mini  & $+0.078$ & $+0.187$ & $+0.198$ \\
\bottomrule
\end{tabular}
\end{table}

\begin{table}[t]\centering
\caption{Effect of each prompt condition relative to Replicate, per model (ANES~2020). Cells report the mean change in JS-divergence $\overline{\Delta \mathrm{JSD}}$ and absolute desirability gap $\overline{\Delta |D|}$ (condition $-$ baseline) with 95\% intervals across items; negative denotes improvement, and bold marks intervals excluding zero. Reformulated, Priming, and Preamble average over the nine directional items; Reverse-coded over the six eligible items against its own baseline. Both item sets include items with a \textit{negative} baseline gap, so $\overline{\Delta |D|}$ describes distance from the human distribution rather than SDB reduction alone. These intervals are computed over the set of per-item changes (one value per item); respondent-level bootstrap intervals and per-item significance are in Appendix~\ref{sec:full-jsd-gap-tables}.}
\label{tab:all-conditions}
\small
\setlength{\tabcolsep}{3pt}
\renewcommand{\cellalign}{cc}
\begin{tabular}{lcccccc}
\toprule
& \multicolumn{2}{c}{Llama-3.1-8B} & \multicolumn{2}{c}{Llama-3.1-70B} & \multicolumn{2}{c}{GPT-4.1-mini} \\
\cmidrule(lr){2-3}\cmidrule(lr){4-5}\cmidrule(lr){6-7}
Condition & $\overline{\Delta \mathrm{JSD}}$ & $\overline{\Delta |D|}$ & $\overline{\Delta \mathrm{JSD}}$ & $\overline{\Delta |D|}$ & $\overline{\Delta \mathrm{JSD}}$ & $\overline{\Delta |D|}$ \\
\midrule
Replicate (baseline) & \multicolumn{2}{c}{0.172 / 0.314} & \multicolumn{2}{c}{0.141 / 0.178} & \multicolumn{2}{c}{0.078 / 0.198} \\
\midrule
Reformulated & \dcell{-0.026}{-0.058}{+0.005} & \dcell{\textbf{-0.065}}{-0.124}{-0.006} & \dcell{+0.005}{-0.054}{+0.064} & \dcell{+0.043}{-0.052}{+0.138} & \dcell{-0.019}{-0.051}{+0.013} & \dcell{-0.050}{-0.101}{+0.000} \\
Reverse-coded & \dcell{+0.026}{-0.182}{+0.234} & \dcell{+0.012}{-0.256}{+0.279} & \dcell{+0.121}{-0.189}{+0.430} & \dcell{+0.157}{-0.207}{+0.520} & \dcell{+0.146}{-0.210}{+0.502} & \dcell{+0.114}{-0.312}{+0.541} \\
Priming & \dcell{+0.009}{-0.039}{+0.056} & \dcell{-0.051}{-0.147}{+0.046} & \dcell{+0.015}{-0.020}{+0.049} & \dcell{-0.018}{-0.056}{+0.020} & \dcell{\textbf{+0.042}}{+0.002}{+0.082} & \dcell{\textbf{+0.042}}{+0.006}{+0.078} \\
Preamble & \dcell{-0.003}{-0.048}{+0.041} & \dcell{-0.043}{-0.141}{+0.055} & \dcell{+0.007}{-0.005}{+0.020} & \dcell{+0.020}{-0.020}{+0.060} & \dcell{\textbf{+0.043}}{+0.003}{+0.082} & \dcell{\textbf{+0.065}}{+0.016}{+0.115} \\
\bottomrule
\end{tabular}
\end{table}

\begin{figure}[t]
    \centering
    \begin{subfigure}[t]{0.5\textwidth}
        \centering
        \includegraphics[width=\linewidth]{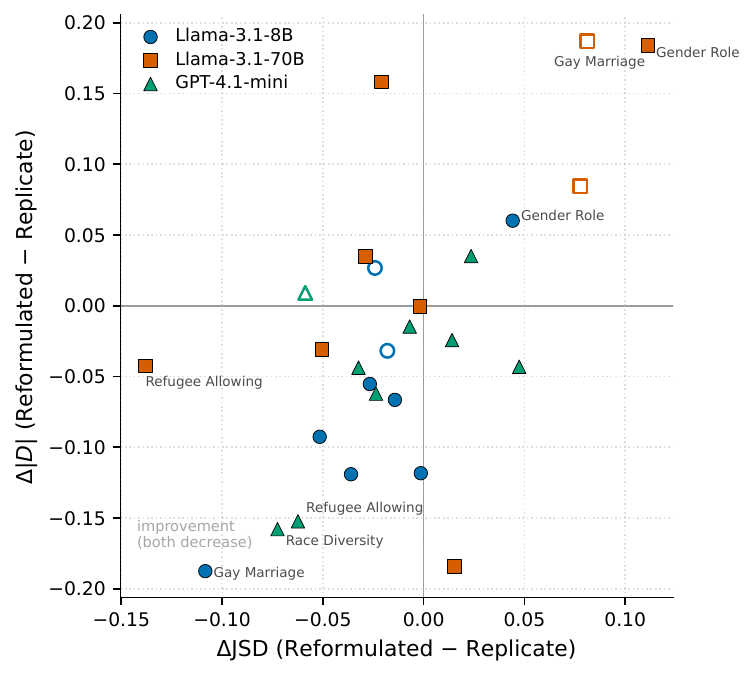}
        \caption{$\Delta$JSD vs.\ $\Delta|D|$}
        \label{fig:reformulated-mechanism}
    \end{subfigure}\hfill
    \begin{subfigure}[t]{0.42\textwidth}
        \centering
        \raisebox{7.5mm}{\includegraphics[width=\linewidth]{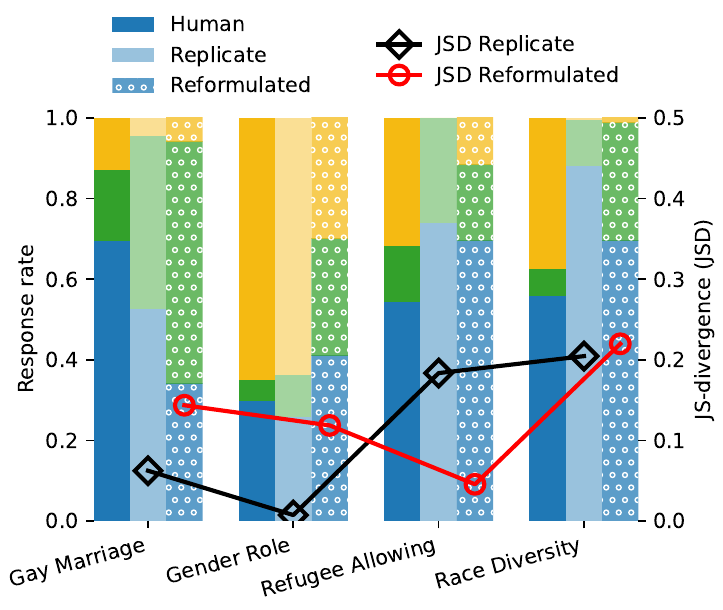}}
        \caption{Llama-3.1-70B distributions}
        \label{fig:reformulated-70b-bars-sel}
    \end{subfigure}
\caption{Reformulated condition. \textbf{(a)} Per-item change (Reformulated $-$ Replicate) in JS-divergence and absolute desirability gap $|D|$. The lower-left quadrant denotes improvement on both. Open markers denote the five model-item pairs with a negative baseline gap, where a narrowing gap reflects improved calibration rather than reduced SDB: \textit{Climate Change} and \textit{Gun Regulation} for Llama-8B, \textit{Climate Change} and \textit{Gay Marriage} for Llama-70B, and \textit{Gender Role} for GPT-4.1-mini.  \textbf{(b)} Human, Replicate, and Reformulated response distributions for Llama-3.1-70B on four selected items, with per-item JS-divergence.}
    \label{fig:reformulated}
\end{figure}

\paragraph{Reformulated Condition (1)}
In this condition, items are neutralized and expressed in a third-person perspective. \autoref{tab:all-conditions} reports the mean change in JS-divergence and the absolute desirability gap relative to the Replicate baseline, and \autoref{fig:reformulated-mechanism} relates the two per item. The confidence intervals in \autoref{tab:all-conditions} treat each item as the unit of analysis, so they indicate whether an effect generalizes beyond the nine items.
Table~\ref{tab:all-conditions} shows that Reformulation reduces both quantities for Llama-8B and GPT-4.1-mini, by the largest margins of any condition, but the across-item interval excludes zero only for $\Delta|D|$ on Llama-8B. We therefore do not claim that the divergence reduction generalizes across items for either model, or that the gap reduction does so for GPT-4.1-mini.

The per-item pattern is nonetheless consistent. In Figure~\ref{fig:reformulated-mechanism}, most Llama-8B and GPT-4.1-mini items fall in the lower-left quadrant, where both divergence and gap decrease: the items that improve most on JS-divergence are those whose desirability gap shrinks most. 
The effect is clearest on socially sensitive items, such as \textit{Race Diversity} and \textit{Refugee Allowing} for GPT-4.1-mini (both $\Delta|D|<-0.15$), where neutral third-person phrasing moves probability mass off the socially desirable option toward the human distribution. \textit{Gender Role} is the exception for Llama-8B (Replicate $D=+0.240, \ \Delta|D|=+0.060$). 
Here the human distribution concentrates on the neutral ``makes no difference'' option (yellow), while reformulation shifts the model further onto the socially desirable option and away from the human distribution (\autoref{fig:reformulated-bars-llama8b}), widening the absolute desirability gap.
This may imply a general limitation of bias mitigation: 
When a model’s internal representations diverge from empirical reality, reducing SDB can push the simulated distribution further from human data.

Llama-70B behaves differently. Its mean gap \emph{widens} under reformulation ($\overline{\Delta|D|}=+0.043$) with almost no change in JS-divergence ($\overline{\Delta\mathrm{JSD}}=+0.005$), and its items disperse across all four quadrants of~\autoref{fig:reformulated-mechanism}, with the largest movements in the upper right. The distributions in~\autoref{fig:reformulated-70b-bars-sel} illustrate this in more detail. The two largest increases, \textit{Gay Marriage} ($\Delta|D|=+0.187$) and \textit{Gender Role} ($\Delta|D|=+0.184$), widen the gap in opposite directions. On \textit{Gay Marriage} the model already selects the socially desirable ``Favor'' (blue) option less often than ANES respondents at baseline (negative Replicate $D=-0.169$), so there is no SDB to mitigate, and shifting further mass toward the socially undesirable ``Oppose'' (green) moves the distribution further from the human one. On \textit{Gender Role} the baseline gap is small and positive (Replicate $D=+0.052$), and reformulation shifts mass from the neutral ``Makes no difference'' (yellow) toward the socially desirable ``Worse'' (green), amplifying the over-selection rather than reducing it toward the human distribution. In both cases mass moves toward the extremes rather than toward the moderate options humans more often choose, reflecting the polarized baseline documented in the Replicate condition. 
\textit{Race Diversity} shows a similar dispersion in its option distribution, and although its gap shrinks, its JS-divergence still worsens ($\Delta\mathrm{JSD}=+0.015$), as reducing the desirability gap does not move it toward the human distribution. \textit{Refugee Allowing} is one of the items where both JS-divergence and the gap improve.

Overall, reformulation reduces both divergence and desirability bias on the two models where it improves alignment. On items with a positive baseline gap the two reductions co-occur at the item level, fulfilling the joint criterion, which is consistent with reduced over-selection of the socially desirable option rather than arbitrary redistribution. 
On the polarized Llama-70B the same intervention sometimes increases diversity without reducing the gap and alignment (Appendix \autoref{fig:reformulated-bars-llama70b}). 

\paragraph{Reverse-coded Condition (2)}
The Reverse-coded condition applies to only the six items for which reverse-coding was judged feasible, and is evaluated against its own six-item Replicate baseline. 
Its results are unstable, and reverse-coding provides limited value as an SDB mitigation strategy.
On average it increases both the JS-divergence and the absolute desirability gap for all three models (\autoref{tab:all-conditions}), and the item-level improvements it does produce do not generalize across models. 
\autoref{fig:combined-reversed-bars} shows that both Llama models exhibit improvements on \textit{Climate Change}, with response distributions more closely resembling the human data. The Replicate signed $D$ is negative for this item in both models (\autoref{tab:peritem-absD-8b} and \autoref{tab:peritem-absD-70b}), so we do not interpret these improvements as SDB mitigation.

A consistent pattern of degradation appears for \textit{Gender Role}, where all models worsen substantially.
This likely reflects the difficulty of preserving semantic equivalence when implementing reverse-coding. The original question evaluates attitudes toward a situation, ``\textit{the man works outside the home and the woman takes care of the home and family.}'' In this condition, it was modified to ``\textit{both the man and the woman share work outside the home and take care of the home and family.}'' We consider this the closest plausible reverse-coded variant, yet it arguably alters the underlying construct rather than merely reversing its polarity. Such semantic shifts likely account for the pronounced misalignment observed for this question.

\paragraph{Priming (3) and Preamble (4) Conditions}
Priming and Preamble produce mixed mean effects across models (\autoref{tab:all-conditions}). Priming increases mean JS-divergence for all three models, while reducing the mean absolute desirability gap for both Llama models and increasing it for GPT-4.1-mini. Preamble slightly reduces both metrics for Llama-8B, but increases them for Llama-70B and GPT-4.1-mini. 

For GPT-4.1-mini, both Priming and Preamble significantly increase the mean absolute gap across items ($\overline{\Delta|D|}=+0.042$ and $+0.065$, both intervals excluding zero), the only conditions in \autoref{tab:all-conditions} for which both metrics generalize across items; both also raise JS-divergence.
At the item level, Llama-8B shows JS-divergence reductions on a few individual items, most visibly on \textit{Race Diversity}, where its responses align more closely with the human data and the absolute gap drops substantially ($\Delta|D|=-0.336$). However, these gains are inconsistent and often reversed on other items.
For Llama-70B and GPT-4.1-mini, both Priming and Preamble increase JS-divergence, and lower response diversity in many cases compared to the Replicate condition. Per-item response distributions are shown in \autoref{fig:combined-priming-bars} and \autoref{fig:combined-preamble-bars}, and complete per-item results are reported in Appendix \ref{sec:full-jsd-gap-tables}.

The degradation in Priming and Preamble is often driven by increased response homogeneity, as model outputs concentrate more heavily on socially desirable options.
One possibility, which our design does not test, is that explicitly instructing models to be ``truthful'' or ``sincere'' makes the evaluative frame more salient, reinforcing socially safe patterns.
This would be consistent with recent evidence on LLMs' agentic behavior where models made more conforming choices under evaluation compared to when they perceived the task as real deployment \citep{lynch2025agentic}. Rather than reducing SDB, such conditions appear to amplify it, resulting in less representative distributions.
Full distribution visualizations for models and conditions are in Appendix~\ref{sec:full-distr-4cond}.

\subsection{Demographic conditioning}
\label{sec:conditioning}
Silicon sampling relies on conditioning the model on a demographic profile, so the procedure is only meaningful if the profile changes the response distribution. A model that ignored it would return the same distribution for every profile, and the aggregate agreement would carry no information about subgroup structure. We therefore re-run the Replicate condition with parts of the profile removed and measure the aggregate JS-divergence from the human distribution.
\begin{table}[h]
\centering
\caption{Mean JS-divergence from the human distribution for GPT-4.1-mini under the Replicate condition, averaged over the ten ANES 2020 items, with the full demographic profile, with the political variables removed, and with no profile at all.}
\label{tab:ablation}
\begin{tabular}{lc}
\toprule
Profile & $\overline{\mathrm{JSD}}$ \\
\midrule
Full demographic profile & 0.090 \\
No Ideology & 0.098 \\
No Ideology or Party & 0.186 \\
No demographics & 0.265 \\
\bottomrule
\end{tabular}
\end{table}

\autoref{tab:ablation} reports the result. Removing the full profile raises the divergence from 0.090 to 0.265, so the profile substantially shapes the response distribution. Removing Ideology alone worsens alignment only slightly, from 0.090 to 0.098, whereas removing Party as well raises it to 0.186, so Party carries signal that is not simply a proxy for Ideology.

Appendix~\ref{app:conditioning} tests each of the eight demographic variables separately, comparing every silicon subgroup against its own human counterpart and against the non-matching ones. Only Ideology and Party produce substantial signal, and every other gain is an order of magnitude smaller. Demographic conditioning is therefore real but narrow. 
The profile as a whole shapes the response distribution, and removing Ideology and Party accounts for roughly half of that (0.090 to 0.186, against 0.265 with no profile), so the remaining attributes affect the distribution as well. However, Ideology and Party are the only attributes that substantially reproduce the structure of the human subgroups.

This narrowness constrains the subgroup analysis that follows. Profiles are sampled independently from the empirical marginals, so every silicon subgroup shares the population-average composition on the remaining attributes while human subgroups differ on those attributes as well. Sampling from the joint distribution would separate these effects, and we leave this to future work.

 \subsection{Demographic-stratified results}

\autoref{tab:stratified-group-joint} summarizes how the four mitigation strategies affect simulation alignment across eight \textit{aggregated} demographic groups using GPT-4.1-mini.
\footnote{We use GPT-4.1-mini for this and the following analyses because it aligns most closely with ANES 2020 at baseline ($\mathrm{JSD}=0.078$) and because its baseline bias is the most directionally consistent of the three models, with signed and absolute mean gaps close ($D=0.187$ against $|D|=0.198$ in \autoref{tab:baseline}).}  For each demographic group (e.g., Race) we compute, per subgroup (e.g., White, Black, Asian), both the JS-divergence and the absolute desirability gap $|D|$, and report the average change in each relative to the Replicate setting. We treat this analysis as \textit{descriptive} and therefore do not test entries individually.
We focus on three representative items with a relatively high baseline signed desirability gap across the three models (\autoref{fig:baseline-gap}): \textit{Race Diversity}, \textit{Refugee Allowing}, and \textit{Income Inequality}.
All three items have a positive baseline gap in every subgroup, so the joint criterion applies throughout.

The Reformulated condition reduces JS-divergence for all groups on \textit{Race Diversity} and \textit{Refugee Allowing}, with the absolute desirability gap narrowing in every case.
The economic item \textit{Income Inequality} behaves differently and fails the joint criterion: the desirability gap narrows for every group, but JS-divergence rises, so the distribution moves away from the human one for a reason unrelated to desirability. 
The other conditions move all eight groups in the same direction on each item. Reverse-coded improves \textit{Race Diversity} for every group and is the only condition to improve \textit{Income Inequality} on both metrics, though this reverses for Ideology and Party, the two attributes the model conditions on most strongly (Appendix \ref{app:conditioning}). Priming degrades every item for every group on both metrics, and Preamble sharply worsens \textit{Refugee Allowing} while leaving \textit{Race Diversity} near baseline.

\begin{table*}[!t]
\centering
\caption{Change relative to the Replicate baseline in JS-divergence $\Delta\mathrm{JSD}$ (top, shaded) and in the absolute desirability gap $\Delta|D|$ (bottom, gray) for GPT-4.1-mini (ANES~2020), across stratified groups and items. Each value is the per-subgroup change averaged within the group; negative denotes improvement for both metrics. Cells are shaded by $\Delta\mathrm{JSD}$ (blue $<0$, red $>0$, intensity scaling with magnitude). The table is a descriptive check of whether the aggregate effects hold in direction across demographic strata; entries are not individually tested for significance.}
\label{tab:stratified-group-joint}\vspace{-5pt}
\resizebox{\linewidth}{!}{\begin{tabular}{lC{1.4cm} C{1.5cm} C{1.6cm}||C{1.4cm} C{1.5cm} C{1.6cm}||C{1.4cm} C{1.5cm} C{1.6cm}||C{1.4cm} C{1.5cm} C{1.6cm}}
\toprule
                   & \multicolumn{3}{c||}{\textbf{Reformulated}} & \multicolumn{3}{c||}{\textbf{Reverse-coded}} & \multicolumn{3}{c||}{\textbf{Priming}} & \multicolumn{3}{c}{\textbf{Preamble}} \\
                   \cmidrule(lr){2-13}
Group              & Race Diversity & Refugee Allowing & Income Inequality & Race Diversity & Refugee Allowing & Income Inequality & Race Diversity & Refugee Allowing & Income Inequality & Race Diversity & Refugee Allowing & Income Inequality \\
\midrule
\multirow{2}{*}{Ideology} & \colorcell{-0.06} & \colorcell{-0.046} & \colorcell{0.041} & \colorcell{-0.139} & \colorcell{-0.042} & \colorcell{0.106} & \colorcell{0.046} & \colorcell{0.025} & \colorcell{0.065} & \colorcell{-0.017} & \colorcell{0.133} & \colorcell{0.065} \\[-11pt]
                         & {\scriptsize\color{black!55}-0.172} & {\scriptsize\color{black!55}-0.122} & {\scriptsize\color{black!55}-0.04} & {\scriptsize\color{black!55}-0.146} & {\scriptsize\color{black!55}-0.086} & {\scriptsize\color{black!55}0.145} & {\scriptsize\color{black!55}0.023} & {\scriptsize\color{black!55}0.027} & {\scriptsize\color{black!55}0.092} & {\scriptsize\color{black!55}-0.011} & {\scriptsize\color{black!55}0.145} & {\scriptsize\color{black!55}0.085} \\
\addlinespace[1pt]
\multirow{2}{*}{Party} & \colorcell{-0.074} & \colorcell{-0.073} & \colorcell{0.035} & \colorcell{-0.146} & \colorcell{-0.059} & \colorcell{0.064} & \colorcell{0.036} & \colorcell{0.009} & \colorcell{0.05} & \colorcell{-0.009} & \colorcell{0.118} & \colorcell{0.057} \\[-11pt]
                      & {\scriptsize\color{black!55}-0.147} & {\scriptsize\color{black!55}-0.117} & {\scriptsize\color{black!55}-0.016} & {\scriptsize\color{black!55}-0.132} & {\scriptsize\color{black!55}-0.109} & {\scriptsize\color{black!55}0.078} & {\scriptsize\color{black!55}0.013} & {\scriptsize\color{black!55}0.026} & {\scriptsize\color{black!55}0.078} & {\scriptsize\color{black!55}-0.005} & {\scriptsize\color{black!55}0.125} & {\scriptsize\color{black!55}0.08} \\
\midrule
\multirow{2}{*}{Political interest} & \colorcell{-0.085} & \colorcell{-0.052} & \colorcell{0.034} & \colorcell{-0.151} & \colorcell{-0.053} & \colorcell{-0.04} & \colorcell{0.057} & \colorcell{0.015} & \colorcell{0.034} & \colorcell{0.005} & \colorcell{0.157} & \colorcell{0.048} \\[-11pt]
                                   & {\scriptsize\color{black!55}-0.153} & {\scriptsize\color{black!55}-0.169} & {\scriptsize\color{black!55}-0.07} & {\scriptsize\color{black!55}-0.145} & {\scriptsize\color{black!55}-0.124} & {\scriptsize\color{black!55}-0.2} & {\scriptsize\color{black!55}0.024} & {\scriptsize\color{black!55}0.032} & {\scriptsize\color{black!55}0.068} & {\scriptsize\color{black!55}0.005} & {\scriptsize\color{black!55}0.156} & {\scriptsize\color{black!55}0.08} \\
\addlinespace[1pt]
\multirow{2}{*}{Church} & \colorcell{-0.077} & \colorcell{-0.064} & \colorcell{0.045} & \colorcell{-0.15} & \colorcell{-0.045} & \colorcell{-0.035} & \colorcell{0.035} & \colorcell{0.013} & \colorcell{0.031} & \colorcell{-0.013} & \colorcell{0.133} & \colorcell{0.035} \\[-11pt]
                       & {\scriptsize\color{black!55}-0.159} & {\scriptsize\color{black!55}-0.152} & {\scriptsize\color{black!55}-0.044} & {\scriptsize\color{black!55}-0.137} & {\scriptsize\color{black!55}-0.085} & {\scriptsize\color{black!55}-0.146} & {\scriptsize\color{black!55}0.013} & {\scriptsize\color{black!55}0.033} & {\scriptsize\color{black!55}0.072} & {\scriptsize\color{black!55}-0.005} & {\scriptsize\color{black!55}0.142} & {\scriptsize\color{black!55}0.069} \\
\addlinespace[1pt]
\multirow{2}{*}{Gender} & \colorcell{-0.075} & \colorcell{-0.061} & \colorcell{0.047} & \colorcell{-0.15} & \colorcell{-0.045} & \colorcell{-0.031} & \colorcell{0.037} & \colorcell{0.013} & \colorcell{0.035} & \colorcell{-0.013} & \colorcell{0.134} & \colorcell{0.038} \\[-11pt]
                       & {\scriptsize\color{black!55}-0.159} & {\scriptsize\color{black!55}-0.151} & {\scriptsize\color{black!55}-0.041} & {\scriptsize\color{black!55}-0.14} & {\scriptsize\color{black!55}-0.085} & {\scriptsize\color{black!55}-0.165} & {\scriptsize\color{black!55}0.013} & {\scriptsize\color{black!55}0.033} & {\scriptsize\color{black!55}0.077} & {\scriptsize\color{black!55}-0.006} & {\scriptsize\color{black!55}0.143} & {\scriptsize\color{black!55}0.072} \\
\addlinespace[1pt]
\multirow{2}{*}{Age} & \colorcell{-0.072} & \colorcell{-0.059} & \colorcell{0.05} & \colorcell{-0.149} & \colorcell{-0.044} & \colorcell{-0.03} & \colorcell{0.037} & \colorcell{0.014} & \colorcell{0.033} & \colorcell{-0.012} & \colorcell{0.134} & \colorcell{0.036} \\[-11pt]
                    & {\scriptsize\color{black!55}-0.158} & {\scriptsize\color{black!55}-0.153} & {\scriptsize\color{black!55}-0.042} & {\scriptsize\color{black!55}-0.139} & {\scriptsize\color{black!55}-0.086} & {\scriptsize\color{black!55}-0.15} & {\scriptsize\color{black!55}0.013} & {\scriptsize\color{black!55}0.035} & {\scriptsize\color{black!55}0.075} & {\scriptsize\color{black!55}-0.006} & {\scriptsize\color{black!55}0.144} & {\scriptsize\color{black!55}0.073} \\
\addlinespace[1pt]
\multirow{2}{*}{Race} & \colorcell{-0.036} & \colorcell{-0.069} & \colorcell{0.069} & \colorcell{-0.151} & \colorcell{-0.057} & \colorcell{-0.027} & \colorcell{0.037} & \colorcell{0.028} & \colorcell{0.028} & \colorcell{-0.001} & \colorcell{0.145} & \colorcell{0.021} \\[-11pt]
                     & {\scriptsize\color{black!55}-0.124} & {\scriptsize\color{black!55}-0.151} & {\scriptsize\color{black!55}-0.043} & {\scriptsize\color{black!55}-0.142} & {\scriptsize\color{black!55}-0.108} & {\scriptsize\color{black!55}-0.19} & {\scriptsize\color{black!55}0.014} & {\scriptsize\color{black!55}0.041} & {\scriptsize\color{black!55}0.067} & {\scriptsize\color{black!55}0} & {\scriptsize\color{black!55}0.133} & {\scriptsize\color{black!55}0.043} \\
\addlinespace[1pt]
\multirow{2}{*}{Discuss politics} & \colorcell{-0.13} & \colorcell{-0.122} & \colorcell{0.036} & \colorcell{-0.193} & \colorcell{-0.081} & \colorcell{-0.061} & \colorcell{0.042} & \colorcell{0.005} & \colorcell{0.032} & \colorcell{-0.033} & \colorcell{0.147} & \colorcell{0.013} \\[-11pt]
                                 & {\scriptsize\color{black!55}-0.166} & {\scriptsize\color{black!55}-0.212} & {\scriptsize\color{black!55}-0.078} & {\scriptsize\color{black!55}-0.185} & {\scriptsize\color{black!55}-0.111} & {\scriptsize\color{black!55}-0.276} & {\scriptsize\color{black!55}0.014} & {\scriptsize\color{black!55}0.018} & {\scriptsize\color{black!55}0.057} & {\scriptsize\color{black!55}-0.018} & {\scriptsize\color{black!55}0.124} & {\scriptsize\color{black!55}0.024} \\
\bottomrule
\end{tabular}}
\end{table*}

We find no evidence that any single demographic axis drives the mitigation effects.
At the subgroup level the pattern is broadly the same, with exceptions concentrated at the poles of the political variables. Per-subgroup breakdowns for both metrics are in Appendix~\ref{sec:full-stratified-results}.

\subsection{Consistency across temporal variation}\label{sec:temporal-analysis}
A key question for silicon sampling is whether prompt-based mitigation strategies remain effective under temporal shifts in survey populations and response patterns. To evaluate this, we compare the Reformulated condition on eight overlapping items between ANES 2020 and the more recent ANES 2024 survey.\footnote{\url{https://electionstudies.org/data-center/2024-time-series-study/}}

As shown in \autoref{tab:reformulated-anes2024}, reformulation significantly lowers JS-divergence for four of the eight overlapping items in ANES 2024: \textit{Climate Change}, \textit{Race Diversity}, \textit{Drug Addiction}, and \textit{Gay Marriage}. For these items the reduction is accompanied by a narrower desirability gap, and the same holds in ANES 2020. The exceptions are also stable across waves: reformulation does not lower divergence for \textit{Income Inequality}, \textit{Gun Regulation}, or \textit{Health Insurance}. 
\textit{Gender Role} falls outside both groups, since divergence falls in 2020 and is essentially unchanged in 2024, but its baseline gap is negative in both years and the mitigation framing therefore does not apply.
\textit{Gun Regulation} crosses from a positive to a negative baseline gap between waves. 

Reformulation helps and fails on the same items in both waves, which suggests the effect is not tied to a single survey year, though the evidence comes from only eight items and one model.
The demographic composition of the two waves differs only modestly (Appendix~\ref{sec:demographic-shifts}). This means the comparison only reflects a small shift in the target population, so it primarily tests whether the effect is stable over time rather than robust to large demographic differences.

\begin{table}[]
\centering
\caption{Reformulated versus Replicate across the eight overlapping items in ANES 2020 and 2024, using GPT-4.1-mini. Replicate $D$ is the signed baseline desirability gap; $\Delta\mathrm{JSD}$ and $\Delta|D|$ give the change relative to Replicate, where negative values indicate improvement. Bold marks items satisfying the joint criterion. 
\textit{Gender Role} in both waves and \textit{Gun Regulation} in 2024 have a negative baseline gap, so the criterion does not apply to them. 
Items marked \textsuperscript{\textdagger} retain their answer options verbatim under reformulation, so only the question framing changes. 
All $\Delta$ cells are significant under the two-sided paired bootstrap test with Benjamini-Hochberg correction at $q=0.05$ except those marked \textsuperscript{ns}.}
\label{tab:reformulated-anes2024}\vspace{-5pt}
\resizebox{\linewidth}{!}{\begin{tabular}{llC{1.5cm}C{1.5cm}C{1.6cm}C{1.5cm}C{1.5cm}C{1.5cm}C{1.5cm}C{1.5cm}}
\toprule
 &  & Climate Change\textsuperscript{\textdagger} & Health Insurance\textsuperscript{\textdagger} & Income Inequality & Race Diversity & Drug Addiction\textsuperscript{\textdagger} & Gay Marriage\textsuperscript{\textdagger} & Gun Regulation\textsuperscript{\textdagger} & Gender Role\textsuperscript{\textdagger} \\
\midrule
\multirow{2}{*}{2020} & $\Delta\mathrm{JSD}$ & $\mathbf{-0.032}$ & $+0.024$ & $+0.047$ & $\mathbf{-0.073}$ & $\mathbf{-0.024}$ & $\mathbf{-0.007}$ & $+0.014$ & $-0.059$ \\
 & Replicate $D$ & $+0.094$ & $+0.198$ & $+0.223$& $+0.429$ & $+0.301$ &$+0.150$ & $+0.049$ & $-0.048$ \\
 & $\Delta|D|$ & $\mathbf{-0.044}$ & $+0.035$ & $-0.043$ & $\mathbf{-0.158}$ & $\mathbf{-0.062}$ & $\mathbf{-0.015}$ & $-0.024^{\text{ns}}$ & $+0.009^{\text{ns}}$ \\
\midrule
\multirow{2}{*}{2024} & $\Delta\mathrm{JSD}$ & $\mathbf{-0.051}$ & $+0.005$ & $+0.074$ & $\mathbf{-0.102}$ & $\mathbf{-0.010}$ & $\mathbf{-0.016}$ & $+0.022$ & $-0.007^{\text{ns}}$ \\
 & Replicate $D$ & $+0.145$ &  $+0.170$ & $+0.173$  & $+0.507$  &  $+0.242$ & $+0.156$ & $-0.034$ & $-0.057$\\
 & $\Delta|D|$ & $\mathbf{-0.142}$ & $+0.017$ & $-0.010^{\text{ns}}$ & $\mathbf{-0.186}$ & $\mathbf{-0.092}$ & $\mathbf{-0.082}$ & $+0.056$ & $+0.030$ \\
\bottomrule
\end{tabular}}
\end{table}

\subsection{Transfer to a new survey and population}\label{sec:generalization-analysis}
We further test the effectiveness of the Reformulated condition under a different survey and population.
We apply it to ten World Values Survey (WVS) Wave~7 items in three Western European countries (Netherlands, Germany, and Great Britain)\footnote{\url{https://www.worldvaluessurvey.org/WVSDocumentationWV7.jsp}} using GPT-4.1-mini. The items were selected to span comparable social, economic, and governance topics with differing levels of social-desirability pressure, and the demographic conditioning variables parallel those used for ANES, with the addition of \textit{education} and \textit{household income} (\autoref{tab:wvs-demo} in Appendix).
We report the change in JS-divergence for each item and the change in the absolute desirability gap $|D|$ for items that have a directional social-desirability pole.

As shown in Table~\ref{tab:reformulated-wvs7}, two items satisfy the joint criterion in all three countries, \textit{Environment vs Economy} and \textit{Strong Leader}, where reformulation significantly lowers JS-divergence and narrows the absolute desirability gap in every country.
In the ideological-mapping analysis (Appendix~\ref{sec:politics-analysis}), \textit{Environment vs Economy} has the strongest left-right separation of any WVS item, while \textit{Strong Leader} is near the middle. For \textit{Environment vs Economy} the strong ideological structure means we cannot fully separate the improvement from a shift along the political axis, so we do not attribute it specifically to reduced social-desirability pressure.

\textit{Government Surveillance} shows one of the largest divergence reductions of any item in all three countries, with the absolute gap narrowing in all three, but its baseline gap is negative throughout.
The narrowing gap therefore reflects improved calibration rather than reduced SDB.
\textit{Maternal Employment} and \textit{Gay Parenting} show conflicting metric directions or worsen on both.
\textit{Immigration Impact} worsens both metrics in the Netherlands and Germany but improves divergence in Great Britain, where the gap change is not significant. 
\textit{Refugee Asylum} and \textit{Immigration Policy} each satisfy the joint criterion in two of the three countries, failing in Germany and Great Britain respectively, where the decrease in divergence or in the gap is not significant.
We report these patterns as observations rather than explaining them.

The cross-country comparison is partly confounded. \textit{Race} (Q290) is not used for Germany in WVS Wave~7, so the German personas condition on one fewer attribute. 
Full response distributions and per-item divergences for the three countries are in Appendix \autoref{fig:wvs-jsd-country}. 
These results indicate that reformulation transfers to a new survey and to Western populations outside the U.S. but only on a subset of items, while reproducing the failure mode the joint criterion is designed to detect. Because the three countries are culturally proximate and the transfer holds for only some items and a single model, broader cross-cultural generalization, in particular to populations whose majority views differ in direction, remains future work.

\begin{table}[]
\centering
\caption{
Reformulated versus Replicate across the selected items, per country (Netherlands, Germany, and Great Britain) in \textbf{WVS Wave 7}, using GPT-4.1-mini. Replicate $D$ is the signed baseline desirability gap; $\Delta\mathrm{JSD}$ and $\Delta|D|$ give the change relative to Replicate, where negative values indicate improvement. Bold marks items satisfying the joint criterion. 
Items marked \textsuperscript{\textdagger} retain their answer options verbatim under reformulation, so only the question framing changes.
All $\Delta$ cells are significant under the two-sided paired bootstrap test with Benjamini-Hochberg correction at $q=0.05$ except those marked \textsuperscript{ns}.
\textit{Government Responsibility} and \textit{Income Equality} are non-directional items for which no desirability gap is defined, so only $\Delta\mathrm{JSD}$ is reported. $N$ is the human sample size.}\label{tab:reformulated-wvs7}\vspace{-5pt}
\resizebox{\linewidth}{!}{\begin{tabular}{lC{2cm}C{1.3cm}C{2.2cm}C{1.4cm}C{1.5cm}C{1.5cm}C{2.0cm}C{1.5cm}C{1.9cm}C{1.3cm}}
\toprule
  & Environment vs Economy\textsuperscript{\textdagger} & Refugee Asylum\textsuperscript{\textdagger} & Government Responsibility\textsuperscript{\textdagger} & Income Equality\textsuperscript{\textdagger} & Immigration Impact & Gay Parenting & Maternal Employment & Immigration Policy\textsuperscript{\textdagger} & Government Surveillance\textsuperscript{\textdagger} & Strong Leader \\
\midrule
\multicolumn{11}{l}{\textbf{Netherlands ($N=2145$)}} \\
$\Delta\mathrm{JSD}$ & $\mathbf{-0.008}$ & $\mathbf{-0.010}$ & $-0.017$ & $-0.028$ &$+0.048$ & $-0.002$\textsuperscript{ns} & $-0.032$ & $\mathbf{-0.077}$ & $-0.081$ & $\mathbf{-0.156}$ \\
 Replicate $D$ & $+0.102$ & $+0.171$ & N/A & N/A & $+0.177$ & $+0.071$ & $+0.029$ & $+0.165$ & $-0.262$ & $+0.336$\\
$\Delta|D|$ & $\mathbf{-0.056}$ & $\mathbf{-0.005}$ & N/A & N/A & $+0.023$ & $+0.027$ & $+0.054$ & $\mathbf{-0.030}$ & $-0.006$\textsuperscript{ns} & $\mathbf{-0.046}$ \\
\midrule
\multicolumn{11}{l}{\textbf{Germany ($N=1528$)}} \\
$\Delta\mathrm{JSD}$ & $\mathbf{-0.017}$ & $-0.001$\textsuperscript{ns} & $-0.019$\textsuperscript{ns} & $-0.031$ & $+0.073$ & $+0.004$\textsuperscript{ns} & $-0.001$\textsuperscript{ns} & $\mathbf{-0.021}$ & $-0.137$ & $\mathbf{-0.091}$ \\
 Replicate $D$ &  $+0.267$ & $+0.111$ & N/A & N/A & $+0.172$ & $+0.141$ & $+0.140$ & $+0.062$ & $-0.184$ & $+0.245$\\
$\Delta|D|$ & $\mathbf{-0.026}$ & $-0.001$\textsuperscript{ns} & N/A & N/A & $+0.054$ & $+0.003$\textsuperscript{ns} & $+0.061$ & $\mathbf{-0.010}$ & $-0.044$ & $\mathbf{-0.022}$ \\
\midrule
\multicolumn{11}{l}{\textbf{Great Britain ($N=2609$)}} \\
$\Delta\mathrm{JSD}$ & $\mathbf{-0.020}$ & $\mathbf{-0.009}$ & $-0.026$ & $-0.039$ & ${-0.032}$ & $+0.006$\textsuperscript{ns} & $-0.011$ & ${-0.010}$ & $-0.102$ & $\mathbf{-0.141}$ \\
 Replicate $D$ & $+0.201$ & $+0.394$ & N/A & N/A & $+0.034$ & $+0.066$ & $+0.092$ & $+0.014$ & $-0.309$ & $+0.257$ \\
$\Delta|D|$ & $\mathbf{-0.050}$ & $\mathbf{-0.004}$ & N/A & N/A & ${-0.006}\textsuperscript{ns}$ & $+0.021$ & $+0.055$ & ${-0.013}$\textsuperscript{ns} & $-0.037$ & $\mathbf{-0.051}$ \\
\bottomrule
\end{tabular}}
\end{table}

\begin{table}[t]
\centering
\small
\caption{Effect of third-person framing alone relative to Replicate, per model (ANES 2020). Cells report the mean change in JS-divergence $\overline{\Delta \mathrm{JSD}}$ and absolute desirability gap $\overline{\Delta|D|}$ (condition $-$ baseline) across the nine directional items, with 95\% intervals over the set of per-item changes; negative denotes improvement, and bold marks intervals excluding zero. On six items the reformulation already applies third-person framing alone, so this condition combines those runs with dedicated framing-only runs on the four items that received additional modifications beyond third-person framing. The item set includes items with a negative baseline gap, so $\Delta|D|$ describes distance from the human distribution rather than SDB reduction alone.}
\label{tab:framing-vs-replicate}
\begin{tabular}{l cc cc}
\toprule
 & \multicolumn{2}{c}{Replicate} & \multicolumn{2}{c}{Third-person only} \\
\cmidrule(lr){2-3}\cmidrule(lr){4-5}
Model & JSD & $|D|$ & $\overline{\Delta \mathrm{JSD}}$ & $\overline{\Delta |D|}$ \\
\midrule
Llama-3.1-8B & 0.172 & 0.314 & $\mathbf{-0.039}$ {\scriptsize $\mathbf{[-0.070,\,-0.007]}$} & $\mathbf{-0.064}$ {\scriptsize $\mathbf{[-0.122,\,-0.006]}$} \\
Llama-3.1-70B & 0.141 & 0.178 & $+0.028$ {\scriptsize $[-0.017,\,+0.073]$} & $+0.029$ {\scriptsize $[-0.089,\,+0.148]$} \\
GPT-4.1-mini & 0.078 & 0.198 & $-0.017$ {\scriptsize $[-0.043,\,+0.008]$} & $\mathbf{-0.043}$ {\scriptsize $\mathbf{[-0.082,\,-0.004]}$} \\
\bottomrule
\end{tabular}
\end{table}

\subsection{Decomposing the Reformulated condition}
\label{sec:decomposition}

The Reformulated condition combines third-person framing with, on some items, further edits to the wording and to the response options. On six of the ten ANES 2020 items the reformulation applies third-person framing alone.
The other \textit{four} also carry additional edits: neutralized wording, relabeled response options, and, for \textit{Refugee Allowing}, a shift to a policy formulation. 
For these four items we ran an additional condition that applies \textit{third-person framing alone} and leaves the question stem and the response options as in the Replicate baseline.
Combined with the six items on which reformulation is already framing alone, this gives a framing-only condition covering the full item set.

\textbf{Effect of third-person framing.} \autoref{tab:framing-vs-replicate} compares framing alone with Replicate across the nine directional items. Framing alone lowers both divergence and the desirability gap for Llama-8B and GPT-4.1-mini, and raises both slightly for Llama-70B. Two of these effects generalize across items. 
%
For Llama-8B the across-item intervals exclude zero on both metrics, the only case in our experiments where both metrics improve reliably across items. Because this item set contains two items (\textit{Climate Change} and \textit{Gun Regulation}) with a negative baseline gap for this model, the across-item result is not itself a test of the joint criterion, which is defined per item and requires a positive baseline gap.
For GPT-4.1-mini the interval on $\Delta|D|$ excludes zero, whereas the corresponding interval under the full reformulation in \autoref{tab:all-conditions} does not. Framing alone therefore reduces the desirability gap more reliably across items than the full reformulation does.

\textbf{Effect of the additional edits.} \autoref{tab:reform-decomposition} reports both conditions against Replicate on the four items that carry additional edits. For the desirability gap, framing alone narrows the gap in eight of the nine item and model combinations, the exception being \textit{Income Inequality} for Llama-70B. Averaged over the three directional items, framing alone is close to the full reformulation for Llama-8B, weaker for GPT-4.1-mini, and stronger for Llama-70B. On \textit{Refugee Allowing} and \textit{Race Diversity} for both Llama models it narrows the gap more than the full reformulation does, so the additional edits add nothing to framing on these items and appear to cancel part of it.

For JS-divergence the pattern is less consistent. Averaged over the four items, framing alone reduces divergence more than the full reformulation for Llama-8B and GPT-4.1-mini, while for Llama-70B the two conditions differ in sign. At the item level they agree in direction on eight of the twelve item and model combinations. Two disagreements fall on \textit{Refugee Allowing}, where the full reformulation gives the largest reduction for Llama-70B ($-0.138$) and a substantial one for GPT-4.1-mini ($-0.062$) but framing alone does not. This is the only item that receives a policy formulation, which plausibly accounts for the difference. On \textit{Income Inequality} for GPT-4.1-mini the ordering reverses and on Llama-70B the effect is not significant. The additional edits therefore help on some combinations and hurt on others, and with four items we cannot say which part of the reformulation drives the change in divergence.

This four-item comparison is descriptive: there are too few items for inference across items, and the remaining edits vary together rather than one at a time. A factorial design over the full item set is left to future work. Overall, however, the third-person framing appears to carry most of what the Reformulated condition achieves. The additional edits are item-specific and contribute inconsistently.

\begin{table}[t]
\centering
\setlength{\tabcolsep}{4.5pt}
\small
\caption{Decomposition of the Reformulated condition on the four ANES 2020 items with a third-person-only variant. For each item and model we report the change relative to Replicate under third-person framing alone and under the full reformulation, for both $\Delta$JSD (four items) and $\Delta|D|$ (three directional items; \textit{Current Economy} has no desirability pole). Negative values indicate improvement; \textsuperscript{ns} marks a change that is not significant after Benjamini-Hochberg correction, applied within this decomposition (eight comparisons for $\Delta\mathrm{JSD}$, six for $\Delta|D|$) separately per metric and model.
}
\label{tab:reform-decomposition}
\begin{tabular}{ll ccc ccc}
\toprule
 & & \multicolumn{3}{c}{Third-person only} & \multicolumn{3}{c}{Full reformulation} \\
\cmidrule(lr){3-5}\cmidrule(lr){6-8}
 & Item & 8B & 70B & GPT & 8B & 70B & GPT \\
\midrule
\multirow{5}{*}{$\Delta$JSD} & Current Economy & $-0.105$ & $+0.003$\textsuperscript{ns} & $-0.106$ & $-0.134$ & $+0.070$ & $-0.049$ \\
 & Refugee Allowing & $-0.044$ & $+0.002$\textsuperscript{ns} & $+0.013$ & $-0.027$ & $-0.138$ & $-0.062$ \\
 & Income Inequality & $-0.035$ & $+0.003$\textsuperscript{ns} & $-0.012$ & $-0.001$\textsuperscript{ns} & $-0.002$\textsuperscript{ns} & $+0.047$ \\
 & Race Diversity & $-0.074$ & $+0.074$ & $-0.074$ & $-0.014$ & $+0.015$ & $-0.073$ \\
\addlinespace[2pt]
 & \textit{Mean} & $-0.065$ & $+0.020$ & $-0.045$ & $-0.044$ & $-0.014$ & $-0.034$ \\
\midrule
\multirow{4}{*}{$\Delta|D|$} & Refugee Allowing & $-0.084$ & $-0.075$ & $-0.094$ & $-0.055$ & $-0.042$ & $-0.152$ \\
 & Income Inequality & $-0.046$ & $+0.018$ & $-0.065$ & $-0.118$ & $-0.000$ & $-0.043$ \\
 & Race Diversity & $-0.102$ & $-0.297$ & $-0.129$ & $-0.067$ & $-0.184$ & $-0.158$ \\
\addlinespace[2pt]
 & \textit{Mean} & $-0.077$ & $-0.118$ & $-0.096$ & $-0.080$ & $-0.075$ & $-0.118$ \\
\bottomrule
\end{tabular}
\end{table}
 
\subsection{Possible mechanisms behind the reformulation effect}\label{sec:discussion-reform}
There are at least three possible explanations for why reformulation works. First, it may reduce the model's perception of being evaluated, the motivation behind our design~\citep{salecha2024large}.
This is consistent with the third-person framing reducing the perception of evaluation. 
Second, the third-person format could change the model's position relative to the user. Instead of stating its own opinion to a user whose stance it may infer and attempt to agree with, it predicts another person's answer, leading to reduced sycophancy~\citep{ranaldi2023large,sharma2024towards}. 
Third, describing a respondent may shift the model from role-playing a persona~\citep{Argyle2023,shanahan2023role} to predicting how a population would answer. 

The decomposition in Section~\ref{sec:decomposition} shows which edit carries the effect but not which mechanism. Third-person framing narrows the gap on most items that receive no other edits, and lowers divergence on most of them for Llama-8B and GPT-4.1-mini, so the additional edits are not necessary for the effect. All three explanations can operate through the third-person format, so this does not favor any of them. Separating them would require varying the stated user stance or the persona instruction directly, which we leave to future work.

\section{Conclusion}

This work presents a systematic study of prompt-level mitigation strategies for SDB in LLM-based survey simulation. Building on random silicon sampling, we evaluate four strategies, Reformulated, Reverse-coded, Priming, and Preamble, against a Replicate baseline on three instruction-tuned LLMs. 
We pair JS-divergence with a signed desirability gap, counting as SDB mitigation only a joint decrease in divergence and in the absolute gap for items with a positive baseline signed gap. We note that on items where the desirable pole aligns with the politically progressive response, the gap may reflect desirability and ideological priors jointly (Appendix~\ref{sec:politics-analysis}).
At baseline, all models on average assign more probability mass to socially desirable responses than human respondents, which illustrates SDB in instruction-tuned models and motivates our mitigation study. The form of this bias differs across models: concentration on the desirable option in Llama-8B and GPT-4.1-mini, and a more polarized response pattern in Llama-70B.

Reformulation is the most effective of the strategies we test. On average, it yields both lower divergence from human data and less concentration on socially desirable options for Llama-8B and GPT-4.1-mini. However, when the model's baseline is already polarized, it can shift the distribution further away from the more moderate human distribution. These effects hold across survey waves on overlapping items, but transfer to a different survey and three non-U.S.\ Western populations on a subset of items only. Results are stable across decoding strategies for GPT-4.1-mini (Appendix~\ref{app:decoding}), indicating that they follow from the prompt design rather than from decoding.

Treating items as the unit of analysis narrows what can be claimed. Most aggregate effects do not generalize beyond the items we study. Third-person framing is the exception: isolated from the other edits, it reduces both divergence and the desirability gap reliably across items for Llama-8B, and the desirability gap for GPT-4.1-mini, where the full reformulation does not (Section~\ref{sec:decomposition}). The additional edits are item-specific and contribute inconsistently. Conversely, Priming and Preamble reliably increase both divergence and the gap for GPT-4.1-mini.

Demographic-stratified analyses are descriptive and we do not test their entries individually. Mitigation effects are broadly similar across demographic subgroups, with exceptions concentrated at the poles of the political variables. We find no evidence that any single demographic axis drives the mitigation effects. 
Removing the profile substantially raises the aggregate divergence, which suggests that demographic conditioning meaningfully shapes the response distribution. However, only Ideology and Party reproduce the structure of the human subgroups (Appendix~\ref{app:conditioning}).

Reformulation does not fully eliminate SDB. Its effectiveness varies with item sensitivity and is bounded by the model's underlying knowledge of subgroup opinion. It also remains an open question which mechanism produces the effect, since reduced evaluation awareness, reduced sycophancy, and a shift away from persona role-play can all operate through this intervention.
The remaining strategies produce limited or inconsistent benefits. Reverse-coding is highly sensitive to semantic fidelity. Priming and preamble tend to increase response uniformity, suggesting that simple instructions are insufficient to mitigate SDB and may even exacerbate it.

These findings underscore the importance of careful prompt design for LLM-based population simulation and of evaluating mitigation with a direction-aware measure alongside distributional distance. 
There are two future directions. Separating the candidate mechanisms requires varying the stated user stance or the persona instruction directly, and separating the individual edits requires a factorial design over the full item set. Beyond this, the evidence base itself needs widening: larger item pools that can support inference across items, desirability labels validated by expert annotators, comparison across further model families, demographic profiles that preserve attribute correlations, and populations whose majority views differ in direction from those studied here. Combining prompt-level strategies with post-hoc calibration is a further direction where human benchmarks are available.

\section*{Limitations}
\label{sec:limitations}
\textbf{Model dependence.}
We observe variation in baseline silicon-sampling performance across LLMs. These differences likely reflect model-specific training data, architectures, and bias mechanisms. Although reformulation reduces SDB on most items for two of the three models, its effectiveness may vary in others. Thus, our conclusions should not be assumed to generalize to all LLMs, and broader evaluation across model families is needed.

\textbf{Alignment scope.}
Our analysis is restricted to aligned, instruction-tuned models that the systems practitioners usually deploy for silicon sampling. Because alignment and preference tuning likely shape how models respond to socially sensitive items, we cannot rule out that the divergence we document is driven substantially by these post-training stages rather than by pretraining alone. We do not compare against pre-alignment base checkpoints, which would help isolate this effect but that are often unavailable for closed models and exhibit weaker instruction following (e.g., less stable output formatting). 
We thus view SDB as a property of the deployed systems silicon sampling emulates rather than an artifact to be avoided by sampling from base models, and leave a systematic study across the pretraining–alignment pipeline to future work.

\textbf{Population coverage.}
Our analysis focuses on the U.S. population represented in the ANES data, with an extended study on three Western European populations (WVS Wave~7). 
These are culturally proximate to the U.S., so the effectiveness of mitigation strategies like item reformulation may not extend to more distant cultural contexts. 
Because we sample attributes independently from their empirical marginals (Equation~\ref{eq:sampling}), some profiles may be implausible. Since all comparisons are aggregate-level, this is unlikely to invalidate our conclusions, though we cannot rule out residual artifacts from profiles with conflicting attributes.
Future research should validate these findings across broader demographic dimensions and global survey datasets.

\textbf{Item coverage.} Our main benchmark comprises ten items, nine of which carry a desirability direction. To ask whether an effect extends beyond these items, we treat the item as the unit of analysis and report an interval over the per-item changes. With nine items these intervals are wide and mostly include zero, so we cannot rule out a null average effect over a broader item set.  

\textbf{Bias, knowledge gaps, and stereotypes.}
Even when SDB is present, attributing a given response to it alone remains
challenging.
A model’s tendency to default to a ``safe'' response may reflect normative pressure, lack of group-specific knowledge, stereotyped associations learned about demographic groups~\citep{cheng2023marked}, or some combination of these. 
Because these sources are not separable from the prompt alone, our aggregate distributions may conflate genuine population-level regularities with stereotypical priors. Moreover, where the model lacks accurate internal representations of subgroup preferences, bias mitigation alone may be ineffective or even misleading. This limits our ability to precisely quantify the magnitude of SDB and the upper bound of mitigation performance.

\textbf{Other sources of response bias.}
SDB is one of several biases that may shape silicon-sampling outputs. 
LLMs exhibit a mix of human- and non-human-like biases, reproducing some human tendencies but omitting or reversing others~\citep{chen2025manager}. They can also favor answer options by placement or ID rather than content (position bias)~\citep{zheng2024large}; we use a fixed option ordering across all conditions, holding this constant for matched comparisons, though absolute distributions may retain residual effects.
Models may also homogenize, collapsing toward similar answers even under stochastic decoding~\citep{jiang2026artificial}; since faithful sampling must capture the variance of human attitudes, this flattens within-group diversity regardless of SDB.  
Sycophancy may further interact with SDB: models defer to perceived user views on subjective items but resist on objective ones~\citep{ranaldi2023large}. 
We focus on SDB because it is measurable and theoretically grounded, and leave the interaction of these mechanisms to future work.

\textbf{Contextual independence.} Our study uses single-item prompting to isolate the effects of specific mitigation strategies and items, maintaining a controlled environment. However, real-world respondents do not provide answers in isolation; they navigate a sequence of questions that are susceptible to Question Order Bias~\citep{mcfarland1981effects}. Thus, while our study establishes the efficacy of reformulation for individual items, it does not account for the cumulative contextual biases that may emerge in full-scale sequential survey simulations.

\textbf{Reformulation as a combined intervention.} Section~\ref{sec:decomposition} isolates third-person framing across the full item set, but the further edits applied to four items are varied together, so the contribution of each one separately is unknown.
The reformulated items are also a different instrument. We have not put them to human respondents, so we cannot say whether humans would answer them as they answer the originals, though the human distribution on the original items remains our target throughout. Each condition also uses a single hand-written wording, so framing and our particular phrasing are not separable.

\textbf{Prompt placement.} When prompting LLMs, the demographic profile is always in the system message and the item in the user message (Appendix \ref{sec:prompt-eg}). We do not vary this placement, and results may differ under other placements.

\section*{Ethical Considerations}
\label{sec:ethical}

Silicon sampling raises ethical complications that deserve careful attention, and should support human perspectives rather than replace them. 
In the following, we discuss the key ethical considerations involved in mitigating SDB in silicon sampling.

\noindent\textbf{Risks of mitigation.} Mitigating SDB can mask rather than resolve a deeper limitation. A closer match to a population benchmark may reflect either grounded subgroup knowledge or merely outputs shifted toward a plausible target. Where the model lacks genuine subgroup knowledge, removing surface bias is not merely ineffective but misleading.
Reframing should therefore be reported as the reduction of a specific, measurable bias, not as evidence that the resulting sample is representative.
Relatedly, the defined socially desirable option aligns with the more left-leaning response on all directional items we study, strongly so on most ANES items (Appendix~\ref{sec:politics-analysis}), so reducing $D$ on these items also shifts the simulated distribution along a political axis. Mitigation results on such items should not be reported as bias removal alone, but alongside a measure of how far the item's desirability direction is entangled with \textit{political direction} in the target population.

\noindent\textbf{Synthetic data.}
While LLMs efficiently simulate population-level responses, their outputs are model-based, not human-based, and should not be treated as empirical facts. Researchers should clearly disclose the synthetic nature of these data and \emph{label synthetic outputs wherever they appear}.

\noindent\textbf{Bias.} Even carefully framed prompts cannot eliminate the effects of societal inequalities, blind spots, or misinformation embedded in training data. This is critical when simulating perspectives from marginalized groups, where biased outputs can reinforce stigma. Transparency in prompt design, model parameters, and demographic conditioning, as well as a brief \emph{harm review} for sensitive items, is essential.

\noindent\textbf{Privacy.} Participant privacy must also be protected. Research should rely on aggregate data, avoid reconstructing individual responses or using outputs to characterize or make decisions about individuals, comply with dataset licenses, and share only what is needed for replication.

\subsubsection*{Acknowledgments}
We thank our reviewers for their insightful comments, which helped improve the paper.

We used a large language model as an assistive tool to refine the writing and to make minor additions to our analysis code. All experimental results, code, and claims were checked and verified by the authors, who take full responsibility for the content of the paper.

\bibliography{main}

@article{kojima2022large,
  title={Large language models are zero-shot reasoners},
  author={Kojima, Takeshi and Gu, Shixiang Shane and Reid, Machel and Matsuo, Yutaka and Iwasawa, Yusuke},
  journal={Advances in neural information processing systems},
  volume={35},
  pages={22199--22213},
  year={2022}
}

@article{radford2018improving,
  title={Improving language understanding by generative pre-training},
  author={Radford, Alec and Narasimhan, Karthik and Salimans, Tim and Sutskever, Ilya and others},
  year={2018},
  publisher={San Francisco, CA, USA}
}

@article{yang2024oasis,
  title={Oasis: Open agent social interaction simulations with one million agents},
  author={Yang, Ziyi and Zhang, Zaibin and Zheng, Zirui and Jiang, Yuxian and Gan, Ziyue and Wang, Zhiyu and Ling, Zijian and Chen, Jinsong and Ma, Martz and Dong, Bowen and others},
  journal={arXiv preprint arXiv:2411.11581},
  year={2024}
}

@misc{aher2023,
  title={Using Large Language Models to Simulate Multiple Humans and Replicate Human Subject Studies},
  author={Gati Aher and Rosa I. Arriaga and Adam Tauman Kalai},
  year={2023},
  eprint={2208.10264},
  archivePrefix={arXiv},
  primaryClass={cs.CL},
  url={https://arxiv.org/abs/2208.10264}
}

@misc{törnberg2023,
  title={ChatGPT-4 Outperforms Experts and Crowd Workers in Annotating Political Twitter Messages with Zero-Shot Learning},
  author={Petter Törnberg},
  year={2023},
  eprint={2304.06588},
  archivePrefix={arXiv},
  primaryClass={cs.CL},
  url={https://arxiv.org/abs/2304.06588}
}

@article{yu2024towards,
  title={Towards more accurate US presidential election via multi-step reasoning with large language models},
  author={Yu, Chenxiao and Weng, Zhaotian and Li, Yuangang and Li, Zheng and Hu, Xiyang and Zhao, Yue},
  journal={arXiv preprint arXiv:2411.03321},
  year={2024}
}

@article{zhang2024electionsim,
  title={Electionsim: Massive population election simulation powered by large language model driven agents},
  author={Zhang, Xinnong and Lin, Jiayu and Sun, Libo and Qi, Weihong and Yang, Yihang and Chen, Yue and Lyu, Hanjia and Mou, Xinyi and Chen, Siming and Luo, Jiebo and others},
  journal={arXiv preprint arXiv:2410.20746},
  year={2024}
}

@article{arora2025ai,
  title={AI--human hybrids for marketing research: Leveraging large language models (LLMs) as collaborators},
  author={Arora, Neeraj and Chakraborty, Ishita and Nishimura, Yohei},
  journal={Journal of Marketing},
  volume={89},
  number={2},
  pages={43--70},
  year={2025},
  publisher={SAGE Publications Sage CA: Los Angeles, CA}
}

@article{serapio2023personality,
  title={Personality traits in large language models},
  author={Serapio-Garc{\'\i}a, Gregory and Safdari, Mustafa and Crepy, Cl{\'e}ment and Sun, Luning and Fitz, Stephen and Abdulhai, Marwa and Faust, Aleksandra and Matari{\'c}, Maja},
  year={2023}
}

@misc{besta2025psychologicallyenhancedaiagents,
  title={Psychologically Enhanced AI Agents},
  author={Maciej Besta and Shriram Chandran and Robert Gerstenberger and Mathis Lindner and Marcin Chrapek and Sebastian Hermann Martschat and Taraneh Ghandi and Patrick Iff and Hubert Niewiadomski and Piotr Nyczyk and Jürgen Müller and Torsten Hoefler},
  year={2025},
  eprint={2509.04343},
  archivePrefix={arXiv},
  primaryClass={cs.AI},
  url={https://arxiv.org/abs/2509.04343}
}

@article{gao2023s3,
  title={S3: Social-network simulation system with large language model-empowered agents},
  author={Gao, Chen and Lan, Xiaochong and Lu, Zhihong and Mao, Jinzhu and Piao, Jinghua and Wang, Huandong and Jin, Depeng and Li, Yong},
  journal={arXiv preprint arXiv:2307.14984},
  year={2023}
}

@article{qiu2024interactive,
  title={Interactive agents: Simulating counselor-client psychological counseling via role-playing llm-to-llm interactions},
  author={Qiu, Huachuan and Lan, Zhenzhong},
  journal={arXiv preprint arXiv:2408.15787},
  year={2024}
}

@article{schramowski2022large,
  title={Large pre-trained language models contain human-like biases of what is right and wrong to do},
  author={Schramowski, Patrick and Turan, Cigdem and Andersen, Nico and Rothkopf, Constantin A and Kersting, Kristian},
  journal={Nature Machine Intelligence},
  volume={4},
  number={3},
  pages={258--268},
  year={2022},
  publisher={Nature Publishing Group UK London}
}

@article{lee2024can,
  title={Can large language models estimate public opinion about global warming? An empirical assessment of algorithmic fidelity and bias},
  author={Lee, Sanguk and Peng, Tai-Quan and Goldberg, Matthew H and Rosenthal, Seth A and Kotcher, John E and Maibach, Edward W and Leiserowitz, Anthony},
  journal={PLoS Climate},
  volume={3},
  number={8},
  pages={e0000429},
  year={2024},
  publisher={Public Library of Science}
}

@inproceedings{park2023generative,
  title={Generative agents: Interactive simulacra of human behavior},
  author={Park, Joon Sung and O'Brien, Joseph and Cai, Carrie Jun and Morris, Meredith Ringel and Liang, Percy and Bernstein, Michael S},
  booktitle={Proceedings of the 36th annual acm symposium on user interface software and technology},
  pages={1--22},
  year={2023}
}

@article{gallegos2024bias,
  title={Bias and fairness in large language models: A survey},
  author={Gallegos, Isabel O and Rossi, Ryan A and Barrow, Joe and Tanjim, Md Mehrab and Kim, Sungchul and Dernoncourt, Franck and Yu, Tong and Zhang, Ruiyi and Ahmed, Nesreen K},
  journal={Computational Linguistics},
  volume={50},
  number={3},
  pages={1097--1179},
  year={2024},
  publisher={MIT Press 255 Main Street, 9th Floor, Cambridge, Massachusetts 02142, USA~…}
}

@article{Argyle2023,
  title={Out of One, Many: Using Language Models to Simulate Human Samples},
  volume={31},
  DOI={10.1017/pan.2023.2},
  number={3},
  journal={Political Analysis},
  author={Argyle, Lisa P. and Busby, Ethan C. and Fulda, Nancy and Gubler, Joshua R. and Rytting, Christopher and Wingate, David},
  year={2023},
  pages={337--351}
}

@article{sarstedt2024using,
  title={Using large language models to generate silicon samples in consumer and marketing research: Challenges, opportunities, and guidelines},
  author={Sarstedt, Marko and Adler, Susanne J and Rau, Lea and Schmitt, Bernd},
  journal={Psychology \& Marketing},
  volume={41},
  number={6},
  pages={1254--1270},
  year={2024},
  publisher={Wiley Online Library}
}

@misc{sun2024,
  title={Random Silicon Sampling: Simulating Human Sub-Population Opinion Using a Large Language Model Based on Group-Level Demographic Information},
  author={Seungjong Sun and Eungu Lee and Dongyan Nan and Xiangying Zhao and Wonbyung Lee and Bernard J. Jansen and Jang Hyun Kim},
  year={2024},
  eprint={2402.18144},
  archivePrefix={arXiv},
  primaryClass={cs.AI},
  url={https://arxiv.org/abs/2402.18144}
}

@article{salecha2024large,
  title={Large language models display human-like social desirability biases in Big Five personality surveys},
  author={Salecha, Aadesh and Ireland, Molly E and Subrahmanya, Shashanka and Sedoc, Jo{\~a}o and Ungar, Lyle H and Eichstaedt, Johannes C},
  journal={PNAS Nexus},
  volume={3},
  number={12},
  pages={pgae533},
  year={2024},
  publisher={Oxford University Press US}
}

@article{bai2025explicitly,
  title={Explicitly unbiased large language models still form biased associations},
  author={Bai, Xuechunzi and Wang, Angelina and Sucholutsky, Ilia and Griffiths, Thomas L},
  journal={Proceedings of the National Academy of Sciences},
  volume={122},
  number={8},
  pages={e2416228122},
  year={2025},
  publisher={National Academy of Sciences}
}

@inproceedings{li2025actions,
  title={Actions speak louder than words: Agent decisions reveal implicit biases in language models},
  author={Li, Yuxuan and Shirado, Hirokazu and Das, Sauvik},
  booktitle={Proceedings of the 2025 ACM Conference on Fairness, Accountability, and Transparency},
  pages={3303--3325},
  year={2025}
}

@inproceedings{liang2021towards,
  title={Towards understanding and mitigating social biases in language models},
  author={Liang, Paul Pu and Wu, Chiyu and Morency, Louis-Philippe and Salakhutdinov, Ruslan},
  booktitle={International conference on machine learning},
  pages={6565--6576},
  year={2021},
  organization={PMLR}
}

@article{lin2024towards,
  title={Towards trustworthy LLMs: a review on debiasing and dehallucinating in large language models},
  author={Lin, Zichao and Guan, Shuyan and Zhang, Wending and Zhang, Huiyan and Li, Yugang and Zhang, Huaping},
  journal={Artificial Intelligence Review},
  volume={57},
  number={9},
  pages={243},
  year={2024},
  publisher={Springer}
}

@article{zhao2025explicit,
  title={Explicit vs. implicit: Investigating social bias in large language models through self-reflection},
  author={Zhao, Yachao and Wang, Bo and Wang, Yan and Zhao, Dongming and He, Ruifang and Hou, Yuexian},
  journal={arXiv preprint arXiv:2501.02295},
  year={2025}
}

@article{white2023prompt,
  title={A prompt pattern catalog to enhance prompt engineering with ChatGPT},
  author={White, J},
  journal={arXiv preprint arXiv:2302.11382},
  year={2023}
}

@article{crowne1960new,
  title={A new scale of social desirability independent of psychopathology.},
  author={Crowne, Douglas P and Marlowe, David},
  journal={Journal of consulting psychology},
  volume={24},
  number={4},
  pages={349},
  year={1960},
  publisher={American Psychological Association}
}

@article{paulhus1991enhancement,
  title={Enhancement and denial in socially desirable responding.},
  author={Paulhus, Delroy L and Reid, Douglas B},
  journal={Journal of personality and social psychology},
  volume={60},
  number={2},
  pages={307},
  year={1991},
  publisher={American Psychological Association}
}

@article{fisher1993social,
  title={Social desirability bias and the validity of indirect questioning},
  author={Fisher, Robert J},
  journal={Journal of consumer research},
  volume={20},
  number={2},
  pages={303--315},
  year={1993},
  publisher={The University of Chicago Press}
}

@article{lynch2025agentic,
  title={Agentic misalignment: How llms could be insider threats},
  author={Lynch, Aengus and Wright, Benjamin and Larson, Caleb and Ritchie, Stuart J and Mindermann, Soren and Hubinger, Evan and Perez, Ethan and Troy, Kevin},
  journal={arXiv preprint arXiv:2510.05179},
  year={2025}
}

@article{mcfarland1981effects,
  title={Effects of question order on survey responses},
  author={McFarland, Sam G},
  journal={Public Opinion Quarterly},
  volume={45},
  number={2},
  pages={208--215},
  year={1981},
  publisher={Oxford University Press}
}

@inproceedings{cheng2023marked,
  title={Marked personas: Using natural language prompts to measure stereotypes in language models},
  author={Cheng, Myra and Durmus, Esin and Jurafsky, Dan},
  booktitle={Proceedings of the 61st Annual Meeting of the Association for Computational Linguistics (Volume 1: Long Papers)},
  pages={1504--1532},
  year={2023}
}

@article{chen2025manager,
  title={A manager and an AI walk into a bar: does ChatGPT make biased decisions like we do?},
  author={Chen, Yang and Kirshner, Samuel N and Ovchinnikov, Anton and Andiappan, Meena and Jenkin, Tracy},
  journal={Manufacturing \& Service Operations Management},
  volume={27},
  number={2},
  pages={354--368},
  year={2025},
  publisher={INFORMS}
}

@article{jiang2026artificial,
  title={Artificial hivemind: The open-ended homogeneity of language models (and beyond)},
  author={Jiang, Liwei and Chai, Yuanjun and Li, Margaret and Liu, Mickel and Fok, Raymond and Dziri, Nouha and Tsvetkov, Yulia and Sap, Maarten and Choi, Yejin},
  journal={Advances in Neural Information Processing Systems},
  volume={38},
  year={2026}
}

@article{jones1971bogus,
  title={The bogus pipeline: A new paradigm for measuring affect and attitude.},
  author={Jones, Edward E and Sigall, Harold},
  journal={Psychological bulletin},
  volume={76},
  number={5},
  pages={349},
  year={1971},
  publisher={American Psychological Association}
}

@article{warner1965randomized,
  title={Randomized response: A survey technique for eliminating evasive answer bias},
  author={Warner, Stanley L},
  journal={Journal of the American statistical association},
  volume={60},
  number={309},
  pages={63--69},
  year={1965},
  publisher={Taylor \& Francis}
}

@article{deville1992calibration,
  title={Calibration estimators in survey sampling},
  author={Deville, Jean-Claude and S{\"a}rndal, Carl-Erik},
  journal={Journal of the American statistical Association},
  volume={87},
  number={418},
  pages={376--382},
  year={1992},
  publisher={Taylor \& Francis}
}

@article{royal2019survey,
  title={Survey research methods: A guide for creating post-stratification weights to correct for sample bias},
  author={Royal, Kenneth D},
  journal={Education in the Health Professions},
  volume={2},
  number={1},
  pages={48--50},
  year={2019},
  publisher={Medknow}
}

@article{angelopoulos2023prediction,
  title={Prediction-powered inference},
  author={Angelopoulos, Anastasios N and Bates, Stephen and Fannjiang, Clara and Jordan, Michael I and Zrnic, Tijana},
  journal={Science},
  volume={382},
  number={6671},
  pages={669--674},
  year={2023},
  publisher={American Association for the Advancement of Science}
}

@inproceedings{zheng2024large,
  title={Large language models are not robust multiple choice selectors},
  author={Zheng, Chujie and Zhou, Hao and Meng, Fandong and Zhou, Jie and Huang, Minlie},
  booktitle={International Conference on Learning Representations},
  volume={2024},
  pages={19426--19454},
  year={2024}
}

@article{stocke2004interdependence,
  title={The interdependence of determinants for the strength and direction of social desirability bias in racial attitude surveys},
  author={Stock{\'e}, Volker},
  journal={None},
  year={2004}
}

@article{tourangeau2007sensitive,
  title={Sensitive questions in surveys.},
  author={Tourangeau, Roger and Yan, Ting},
  journal={Psychological bulletin},
  volume={133},
  number={5},
  pages={859},
  year={2007},
  publisher={American Psychological Association}
}

@inproceedings{ranaldi2024trip,
  title={A trip towards fairness: Bias and de-biasing in large language models},
  author={Ranaldi, Leonardo and Ruzzetti, Elena Sofia and Venditti, Davide and Onorati, Dario and Zanzotto, Fabio Massimo},
  booktitle={Proceedings of the 13th Joint Conference on Lexical and Computational Semantics (* SEM 2024)},
  pages={372--384},
  year={2024}
}

@article{schick2021self,
  title={Self-diagnosis and self-debiasing: A proposal for reducing corpus-based bias in nlp},
  author={Schick, Timo and Udupa, Sahana and Sch{\"u}tze, Hinrich},
  journal={Transactions of the Association for Computational Linguistics},
  volume={9},
  pages={1408--1424},
  year={2021},
  publisher={MIT Press One Rogers Street, Cambridge, MA 02142-1209, USA journals-info~…}
}

@incollection{lu2020gender,
  title={Gender bias in neural natural language processing},
  author={Lu, Kaiji and Mardziel, Piotr and Wu, Fangjing and Amancharla, Preetam and Datta, Anupam},
  booktitle={Logic, language, and security: essays dedicated to Andre Scedrov on the occasion of his 65th birthday},
  pages={189--202},
  year={2020},
  publisher={Springer}
}

@inproceedings{ravfogel2020null,
  title={Null it out: Guarding protected attributes by iterative nullspace projection},
  author={Ravfogel, Shauli and Elazar, Yanai and Gonen, Hila and Twiton, Michael and Goldberg, Yoav},
  booktitle={Proceedings of the 58th annual meeting of the association for computational linguistics},
  pages={7237--7256},
  year={2020}
}

@inproceedings{tokpo2022text,
  title={Text style transfer for bias mitigation using masked language modeling},
  author={Tokpo, Ewoenam Kwaku and Calders, Toon},
  booktitle={Proceedings of the 2022 Conference of the North American Chapter of the Association for Computational Linguistics: Human Language Technologies: Student Research Workshop},
  pages={163--171},
  year={2022}
}

@article{ranaldi2023large,
  title={When large language models contradict humans? large language models' sycophantic behaviour},
  author={Ranaldi, Leonardo and Pucci, Giulia},
  journal={arXiv preprint arXiv:2311.09410},
  year={2023}
}

@article{benjamini1995controlling,
  title={Controlling the false discovery rate: a practical and powerful approach to multiple testing},
  author={Benjamini, Yoav and Hochberg, Yosef},
  journal={Journal of the Royal statistical society: series B (Methodological)},
  volume={57},
  number={1},
  pages={289--300},
  year={1995},
  publisher={Wiley Online Library}
}

@article{grattafiori2024llama,
  title={The llama 3 herd of models},
  author={Grattafiori, Aaron and Dubey, Abhimanyu and Jauhri, Abhinav and Pandey, Abhinav and Kadian, Abhishek and Al-Dahle, Ahmad and Letman, Aiesha and Mathur, Akhil and Schelten, Alan and Vaughan, Alex and others},
  journal={arXiv preprint arXiv:2407.21783},
  year={2024}
}

@article{shanahan2023role,
  title={Role play with large language models},
  author={Shanahan, Murray and McDonell, Kyle and Reynolds, Laria},
  journal={Nature},
  volume={623},
  number={7987},
  pages={493--498},
  year={2023},
  publisher={Nature Publishing Group UK London}
}

@inproceedings{sharma2024towards,
  title={Towards understanding sycophancy in language models},
  author={Sharma, Mrinank and Tong, Meg and Korbak, Tomek and Duvenaud, David and Askell, Amanda and Bowman, Sam and Durmus, Esin and Hatfield-Dodds, Zac and Johnston, Scott and Kravec, Shauna and others},
  booktitle={International Conference on Learning Representations},
  volume={2024},
  pages={110--144},
  year={2024}
}
\bibliographystyle{tmlr}

\clearpage
\appendix

\section{Additional Related Work}
\label{sec:additional-related-work}
\textbf{Survey methodology corrections beyond question design.}
Beyond framing, human survey methodology offers corrections that do not transfer cleanly to LLMs. 
\emph{Procedural} techniques such as anonymity assurances, the bogus-pipeline paradigm~\citep{jones1971bogus}, and randomized response~\citep{warner1965randomized}, reduce desirable responding by altering perceived detectability; for an LLM these are delivered through the prompt and so reduce to the framing interventions.
Two distinct approaches remain. 
The first \emph{measures and controls} for desirable responding via validated scales such as Marlowe--Crowne~\citep{crowne1960new} or the BIDR~\citep{paulhus1991enhancement}, using a respondent's desirability score to adjust their substantive answers. These scales were developed and validated for human respondents, and it remains unclear whether they apply to LLMs, so this is a non-trivial future direction rather than a readily available correction.
The second, \emph{post-hoc statistical adjustment} through calibration weighting~\citep{deville1992calibration} and post-stratification~\citep{royal2019survey}, and human-LLM rectification such as prediction-powered inference~\citep{angelopoulos2023prediction}, corrects responses against population benchmarks or held-out human labels. 
As we treat the LLM as an unsupervised respondent without such calibration signals, we focus on prompt-level interventions and regard these corrections as complementary directions where validation data are available.

\textbf{Bias measurement and mitigation in LLMs.}
A growing body of work measures and mitigates bias in LLMs~\citep{gallegos2024bias}, and the techniques are commonly grouped by the stage at which they intervene. Pre-processing methods modify inputs or training data, as in counterfactual data augmentation~\citep{lu2020gender}. In-training methods edit parameters or the optimization objective, ranging from alignment fine-tuning~\citep{bai2025explicitly} to parameter-efficient~\citep{ranaldi2024trip} and projection-based~\citep{ravfogel2020null} approaches.
Intra-processing methods alter inference behavior without retraining, such as decoding-time self-debiasing~\citep{schick2021self}.
Post-processing methods rewrite or filter generated outputs~\citep{tokpo2022text}. Parameter-level methods reach a model's internal representations but require model access and retraining, while intra-processing methods (which act at inference time, including prompt-level conditioning) are cheap and model-agnostic yet reach only elicited behavior, not the underlying representations. 
Our approach is an intra-processing method, but departs from prior work in both target and goal: we target \emph{social desirability bias} rather than stereotyping, and our goal is population representativeness in silicon sampling, not harm reduction.

\section{Implementation Details}\label{sec:implementation-details}

\subsection{Model configuration}
We use Llama-3.1-8B-Instruct (Llama-8B), Llama-3.1-70B-Instruct (Llama-70B), and GPT-4.1-mini in our study. 
All three models use stochastic sampling. For each demographic profile, we sample from the response distribution rather than selecting the model's most likely answer.
For Llama-8B and Llama-70B, sampling was enabled ($\texttt{do\_sample=True}$) with temperature $=0.6$, top-$p$ $=0.9$, and top-$k=50$; the maximum number of new tokens was set to two, sufficient for the single-word numerical answer options. GPT-4.1-mini followed the same elicitation procedure via the OpenAI API but with temperature $=1.0$ and top-$p$ $=1.0$. Llama models were run on a
high-performance computing cluster 
with nodes equipped with NVIDIA H100 GPUs.

\subsection{Non-response handling}\label{sec:non-response}
Across all items and in both the ANES and WVS data, we compute human response distributions over the substantive answer options only, i.e., the same options presented to the model. Non-substantive responses (e.g., ``Don't know'', ``Refused'', ``Inapplicable'', and equivalent negative survey codes) are excluded, and the remaining option probabilities are renormalized to sum to one before computing JS-divergence. 
Exclusion is applied per item rather than per respondent: an answer is dropped only for the item on which it is non-substantive, and the respondent is kept for every item they did answer. We use this available-case treatment rather than listwise deletion, following prior work \citep{sun2024}.
This keeps the human and silicon distributions defined over an identical support for each item and applies uniformly across datasets. It also ensures that comparisons between ANES and WVS rest on the same treatment of missing responses. Non-substantive codes on the demographic conditioning variables are likewise excluded, and the sampling marginals are computed over the substantive categories only.
We consider silicon sampling with an explicit abstention option as future work.

\subsection{Annotation procedure for desirability labels}
\label{app:annotation}

\noindent\textbf{Criteria.} For each item, annotators identified the single most socially desirable response option, applying the typology of Section~\ref{sec:data-sample} under the following rules:
\begin{itemize}[noitemsep, topsep=0pt]
\item The \textit{socially desirable} option (or pole, for ordinal items) is the answer a respondent would choose to present themselves favorably, e.g., as tolerant, compassionate, fair-minded, or supportive of others.
\item The \textit{socially undesirable} option is the answer that risks social disapproval, e.g., appearing prejudiced, selfish, or unsympathetic.
\item The \textit{safe (neutral or moderate)} option is the answer that avoids strong judgment or policy change, typically a status-quo or middle response.
\item Judge by the norms of the surveyed population (the United States for ANES; the respective countries, the Netherlands, Germany, and Great Britain for WVS), not by one's own views.
\item Judge the position an answer endorses, not its wording: disagreeing with a negatively worded statement can be the socially desirable option.
\item When multiple options avoid the disapproved position, label as desirable only the option that \textit{affirmatively} expresses fair treatment of the group concerned.
\item When no option would make the respondent appear clearly better or worse to others, mark the item as having no clear approval direction.
\end{itemize}

\noindent\textbf{Annotators and blinding.} Labels were assigned independently by five annotators: three authors (all in Computer Science) and two annotators external to the project with no involvement in the study (a graduate student in Mathematics and a researcher in Software Engineering). Two of the five completed several years of education in the United States; the remainder are based in Europe or Asia. Desirability judgments for the ANES items therefore draw on a mix of direct familiarity with U.S. political discourse and familiarity acquired through the survey literature.  The two external annotators were given only the item text and the labeling criteria, with no description of the hypotheses, the models, or the intended analysis. The three author annotators were aware of the study's aims, but all labels were fixed before any model outputs were examined.

\noindent\textbf{Task.} 
Each annotator independently labeled the 20 items (ten from ANES 2020 and ten from WVS Wave 7). Each WVS item was labeled once and the resulting label was applied to all three countries. Annotators received the items in a fixed order together with the criteria above and saw only the original survey question wording and response options, in English. 

\noindent\textbf{Adjudication.} 
Labels were first assigned independently, yielding exact initial agreement on $7/10$ ANES and $4/10$ WVS items (inter-rater reliability: Krippendorff's $\alpha = 0.68$, with non-directional treated as a separate category). For three items (\textit{Climate Change} in ANES and \textit{Immigration Policy}, \textit{Government Surveillance} in WVS), all annotators identified the same directional pole, with two annotators selecting a more moderate option on the scale. Following a brief calibration discussion, all annotators uniformly adopted the option nearest the scale endpoint ($8/10$ and $6/10$, respectively).
For \textit{Environment vs Economy} (WVS), one annotator selected ``other answers'' and adopted the majority reading after discussion ($7/10$ WVS).
For \textit{Gay Marriage} (ANES) and \textit{Gay Parenting} (WVS), initial labels differed because one annotator applied a different normative framework; after clarifying that the coding baseline is the norm prevailing in the survey country, all converged on the majority directional reading ($9/10$ and $8/10$).
The remaining items (\textit{Current Economy} in ANES; \textit{Income Equality} and \textit{Government Responsibility} in WVS) were classified as non-directional: annotators either disagreed on which pole was desirable or judged the item to have no desirable direction, and the disagreement persisted after discussion.  All items classified as directional carry a unanimous final label.

\subsubsection{Political direction analysis on the desirability labels}
\label{sec:politics-analysis}
Because socially desirable and politically progressive positions may coincide on sensitive items, the defined signed desirability gap $D$ could partly reflect an ideological prior induced by preference tuning rather than social desirability pressure alone. To quantify this, we measure each item's left-right difference directly from the human data. We split respondents into \textit{left} and \textit{right} groups by ideology (ANES V201200: codes 1-3 vs.\ 5-7, moderates excluded; WVS Q240: codes 1-4 vs.\ 7-10, center excluded).
We restrict to substantive answers and compute the desirability-weighted score $S(\cdot)$ of Eq.~\ref{equ:signed_gap} within each group. 
We define the left-right difference $\ell = S_{\text{left}} - S_{\text{right}}$ to measure how much more the \textit{left} leans toward the desirable pole than the \textit{right}. We obtain 95\% confidence intervals by resampling respondents within each group with replacement (2,000 replicates) and taking the percentile interval of the recomputed $\ell$. For ANES, a party-based split (V201231x) serves as a robustness check.

Results are in \autoref{tab:anes-ideology-loading} and \autoref{tab:wvs-ideology-loading} for ANES 2020 and WVS Wave 7, respectively. The desirable direction aligns with the more left-leaning response on all directional items, but the strength varies widely. Seven ANES items divide liberal and conservative respondents by $0.44$-$0.60$, whereas \textit{Drug Addiction} ($\ell = 0.18$) and \textit{Gender Role} ($\ell = 0.05$) divide them only weakly, and the WVS differences are consistently smaller ($0.05$-$0.33$). On strongly divided items we interpret $D$ as capturing social desirability pressure and ideological priors jointly; weakly divided items provide comparatively cleaner tests of social desirability. 
On ANES, \textit{Drug Addiction} and \textit{Gender Role} are comparatively cleaner tests, standing apart from the remaining items ($\ell \le 0.18$ against $0.44$-$0.60$), while the WVS differences are more uniform, with \textit{Maternal Employment}, \textit{Refugee Asylum}, \textit{Government Surveillance}, and \textit{Strong Leader} the least politically divided (cross-country means $\ell \le 0.14$).

\begin{table}[t]
\centering
\caption{Left-right difference on the desirability labels in the ANES 2020 human data: $\ell = S_{\mathrm{left}} - S_{\mathrm{right}}$ under the liberal-conservative split of V201200 (top) and, as robustness, the Democrat-Republican split of V201231x (bottom), with 95\% bootstrap CIs in brackets. Positive values indicate that liberal respondents sit closer to the socially desirable pole. Items are ordered by $\ell$ within each block.}
\label{tab:anes-ideology-loading}
\begin{tabular}{l rr cc c}
\toprule
Item & $n_{\mathrm{left}}$ & $n_{\mathrm{right}}$ & $S_{\mathrm{left}}$ & $S_{\mathrm{right}}$ & Left-right gap $\ell$ \\
\midrule
\multicolumn{6}{l}{\textbf{Ideology V201200}}\\
Income Inequality & 1468 & 1509 & 0.779 & 0.178 & 0.600 [0.573, 0.629] \\
Gun Regulation & 1467 & 1499 & 0.819 & 0.237 & 0.582 [0.552, 0.611] \\
Health Insurance & 1462 & 1487 & 0.834 & 0.272 & 0.562 [0.531, 0.591] \\
Climate Change & 1466 & 1499 & 0.877 & 0.386 & 0.491 [0.473, 0.510] \\
Refugee Allowing & 1472 & 1507 & 0.829 & 0.358 & 0.471 [0.440, 0.503] \\
Gay Marriage & 1642 & 1718 & 0.924 & 0.459 & 0.465 [0.438, 0.493] \\
Race Diversity & 1466 & 1492 & 0.818 & 0.381 & 0.437 [0.406, 0.466] \\
\textit{Drug Addiction} & 1460 & 1497 & 0.794 & 0.615 & 0.179 [0.147, 0.211] \\
\textit{Gender Role} & 1453 & 1500 & 0.082 & 0.030 & 0.052 [0.036, 0.068] \\
\midrule
\multicolumn{6}{l}{\textbf{Party V201231x}} \\
Gun Regulation & 2214 & 1919 & 0.783 & 0.245 & 0.538 [0.511, 0.565] \\
Income Inequality & 2217 & 1925 & 0.704 & 0.202 & 0.503 [0.476, 0.528] \\
Health Insurance & 2211 & 1902 & 0.786 & 0.304 & 0.482 [0.454, 0.509] \\
Climate Change & 2213 & 1921 & 0.850 & 0.417 & 0.433 [0.417, 0.450] \\
Refugee Allowing & 2225 & 1924 & 0.736 & 0.353 & 0.383 [0.353, 0.411] \\
Race Diversity & 2214 & 1912 & 0.735 & 0.383 & 0.352 [0.324, 0.379] \\
Gay Marriage & 2522 & 2198 & 0.855 & 0.509 & 0.347 [0.323, 0.371] \\
\textit{Drug Addiction} & 2203 & 1915 & 0.789 & 0.611 & 0.178 [0.150, 0.204] \\
\textit{Gender Role} & 2204 & 1916 & 0.071 & 0.031 & 0.040 [0.027, 0.053] \\
\bottomrule
\end{tabular}
\end{table}

\begin{table}[t]
\centering
\caption{Left-right difference on the desirability labels in the WVS Wave 7 human data, per country: $\ell = S_{\mathrm{left}} - S_{\mathrm{right}}$ under the left-right split of Q240, with 95\% bootstrap CIs in brackets. Positive $\ell$ values indicate that left respondents sit closer to the socially desirable pole. Items are ordered by the mean of $\ell$ across countries.}
\label{tab:wvs-ideology-loading}
\begin{tabular}{l rr cc c}
\toprule
Item & $n_{\mathrm{left}}$ & $n_{\mathrm{right}}$ & $S_{\mathrm{left}}$ & $S_{\mathrm{right}}$ & Left-right gap $\ell$  \\
\midrule
\multicolumn{6}{l}{\textbf{Netherlands}} \\
Environment vs Economy & 544 & 542 & 0.899 & 0.566 & 0.332 [0.285, 0.382] \\
Immigration Impact & 556 & 595 & 0.557 & 0.391 & 0.166 [0.144, 0.188] \\
Immigration Policy & 515 & 590 & 0.560 & 0.389 & 0.171 [0.147, 0.194] \\
Gay Parenting & 551 & 574 & 0.793 & 0.703 & 0.090 [0.060, 0.118] \\
\textit{Strong Leader} & 543 & 574 & 0.772 & 0.585 & 0.186 [0.146, 0.224] \\
\textit{Government Surveillance} & 569 & 599 & 0.834 & 0.715 & 0.120 [0.086, 0.154] \\
\textit{Refugee Asylum} & 478 & 515 & 0.864 & 0.810 & 0.054 [0.009, 0.103] \\
\textit{Maternal Employment} & 564 & 601 & 0.801 & 0.725 & 0.075 [0.048, 0.102] \\
\midrule
\multicolumn{6}{l}{\textbf{Germany}} \\
Environment vs Economy & 548 & 190 & 0.757 & 0.532 & 0.226 [0.149, 0.309] \\
Immigration Impact & 554 & 196 & 0.544 & 0.386 & 0.157 [0.121, 0.196] \\
Immigration Policy & 530 & 194 & 0.625 & 0.521 & 0.105 [0.073, 0.139] \\
Gay Parenting & 529 & 176 & 0.734 & 0.597 & 0.137 [0.084, 0.188] \\
\textit{Strong Leader} & 547 & 191 & 0.791 & 0.674 & 0.117 [0.067, 0.171] \\
\textit{Government Surveillance} & 557 & 197 & 0.843 & 0.790 & 0.052 [0.009, 0.098] \\
\textit{Refugee Asylum} & 553 & 194 & 0.935 & 0.845 & 0.090 [0.038, 0.150] \\
\textit{Maternal Employment} & 553 & 194 & 0.666 & 0.596 & 0.070 [0.024, 0.118] \\
\midrule
\multicolumn{6}{l}{\textbf{Great Britain}} \\
Environment vs Economy & 779 & 428 & 0.806 & 0.516 & 0.290 [0.237, 0.348] \\
Immigration Impact & 798 & 446 & 0.762 & 0.600 & 0.162 [0.136, 0.188] \\
Immigration Policy & 788 & 444 & 0.682 & 0.520 & 0.163 [0.141, 0.185] \\
Gay Parenting & 776 & 423 & 0.805 & 0.600 & 0.205 [0.173, 0.234] \\
\textit{Strong Leader} & 783 & 438 & 0.823 & 0.712 & 0.111 [0.077, 0.148] \\
\textit{Government Surveillance} & 789 & 445 & 0.779 & 0.616 & 0.164 [0.126, 0.201] \\
\textit{Refugee Asylum} & 789 & 447 & 0.720 & 0.591 & 0.129 [0.074, 0.184] \\
\textit{Maternal Employment} & 774 & 439 & 0.739 & 0.617 & 0.122 [0.094, 0.152] \\
\bottomrule
\end{tabular}
\end{table}

\section{Demographic Background}
\subsection{Demographic shifts between ANES 2020 and 2024}\label{sec:demographic-shifts}
\begin{figure}[!t] 
    \centering
    \begin{subfigure}[]{0.38\textwidth} 
        \centering
        \includegraphics[width=\linewidth]{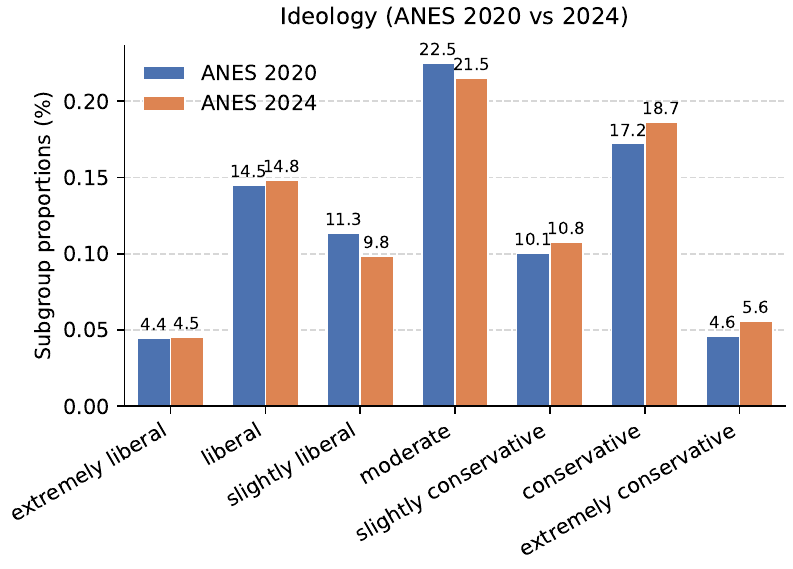}
    \end{subfigure}
     \begin{subfigure}[]{0.38\textwidth} 
        \centering
        \includegraphics[width=\linewidth]{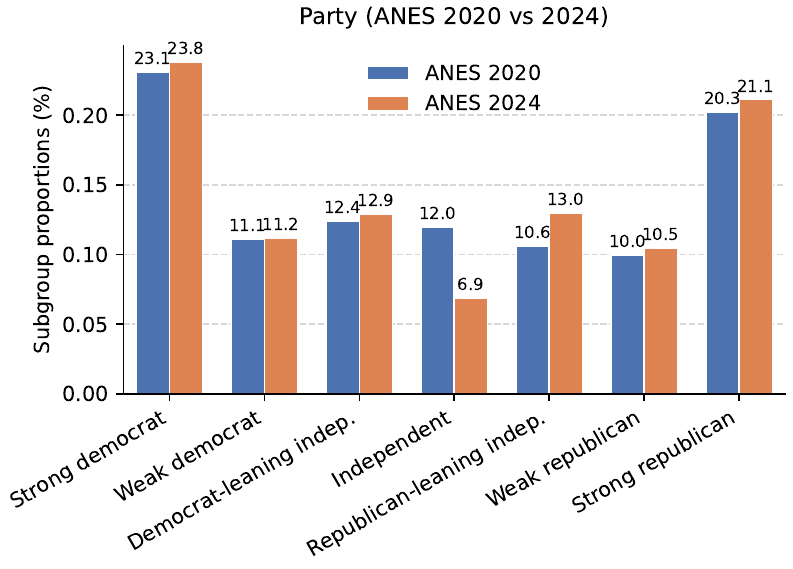}
    \end{subfigure}
    \begin{subfigure}[]{0.215\textwidth} 
        \centering
        \vspace{-28pt}
        \includegraphics[width=\linewidth]{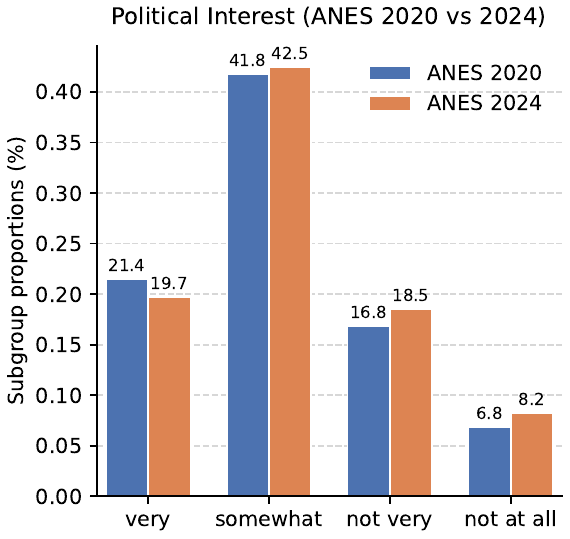}
    \end{subfigure}
    \caption{Subgroup proportions for Ideology, Party, and Political Interest,  ANES 2020 vs. 2024.}
    \label{fig:subgroup-proportions-eg}
\end{figure}

\autoref{fig:subgroup-proportions-eg} illustrates detailed subgroup proportions for \textit{Ideology}, \textit{Party}, and \textit{Political Interest}. We found for \textit{Ideology}, a small portion of respondents became more conservative. For \textit{Party}, more respondents tend to be less concentrated on the most neutral option, ``Independent'', and were more polarized. For \textit{Political Interest}, relatively more respondents became less interested in 2024 compared to 2020.

\subsection{Demographic distributions of the WVS populations} 
\begin{figure}[h] 
    \centering
        \includegraphics[width=0.82\textwidth]{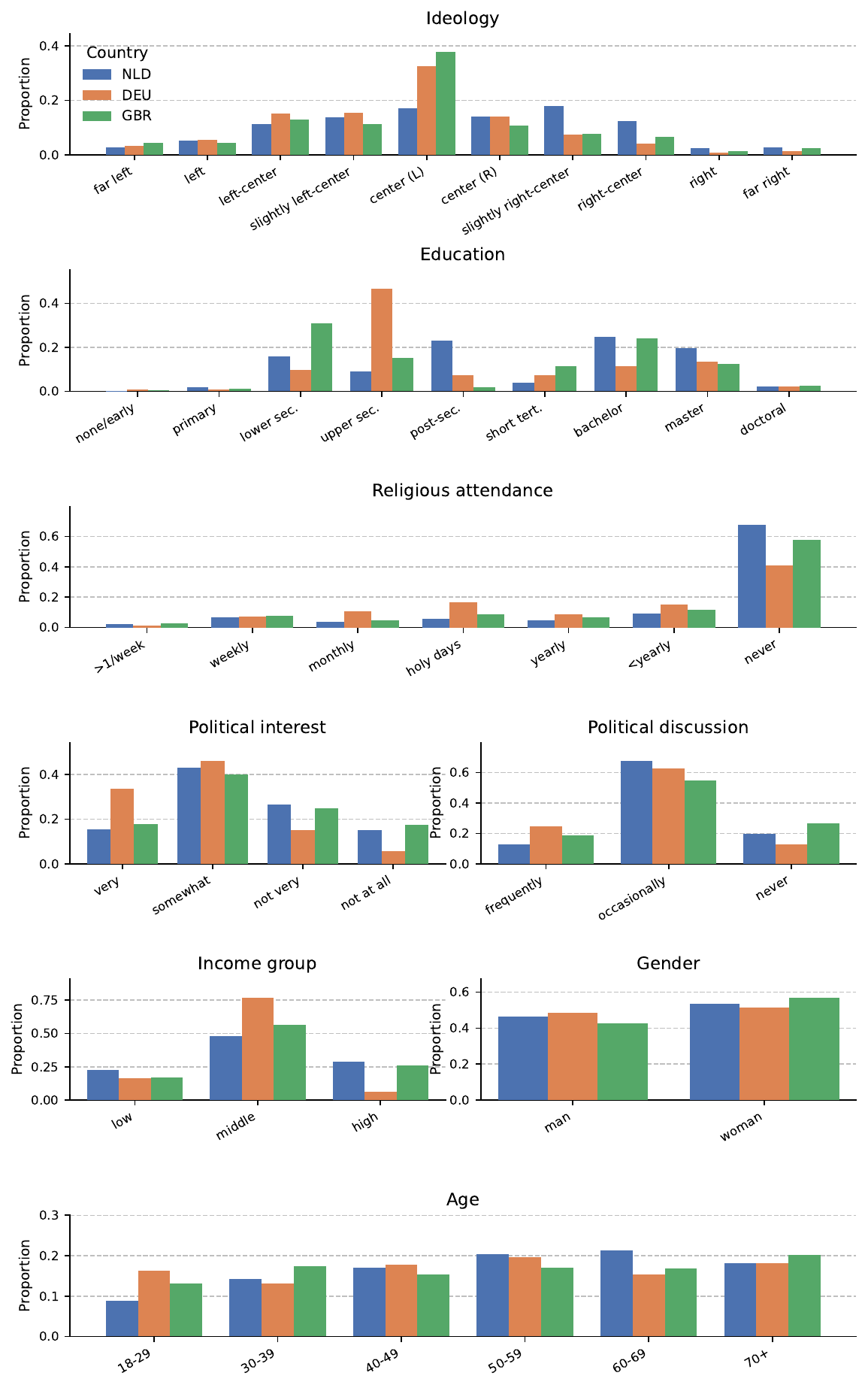}\vspace{-10pt}
    \caption{Distributions of the demographic conditioning variables across the three WVS Wave~7 countries (Netherlands, Germany and Great Britain). Bars show the proportion of respondents in each category per country. Race (Q290) is omitted, as it uses country-specific, non-comparable category schemes, and not used in Germany.}\label{fig:wvs7-demographics}
\end{figure}

\autoref{fig:wvs7-demographics} shows the distributions of the demographic conditioning variables for the three WVS Wave~7 populations. 
The three profiles are broadly similar. Respondents cluster around the ideological center, most report middle household income and upper-secondary education, and the largest share rarely or never attend religious services, with gender close to balanced. Age is spread fairly evenly across decades, with a slight tilt toward respondents aged 50 and over.

 \section{Additional Results}

 \subsection{JS-divergence and desirability gap results for ANES 2020}\label{sec:full-jsd-gap-tables}

\autoref{tab:paired-tests-llama8b}, \autoref{tab:paired-tests-llama70b}, and \autoref{tab:paired-tests-gpt41mini} report per-item JS-divergence at the Replicate baseline and the change under each condition across the three models. \autoref{tab:peritem-absD-8b}, \autoref{tab:peritem-absD-70b}, and \autoref{tab:peritem-absD-gpt} report the corresponding signed desirability gap at baseline and the change in its absolute value. Every cell carries a 95\% bootstrap interval from a paired bootstrap that resamples both silicon and ANES responses ($n = 10{,}000$), and changes relative to the baseline are marked as significant or non-significant using the paired bootstrap test with Benjamini-Hochberg correction at $q = 0.05$ described in Sec.~\ref{sec:evaluation}. Baseline columns report percentile intervals of the statistic itself, while condition columns report percentile intervals of the per-replicate difference. Intervals are unadjusted percentile intervals and are shown for description only, so significance should be read from the markers rather than from interval overlap. $\Delta\mathrm{JSD}$ and $\Delta|D|$ are corrected in separate families, each covering all item-condition comparisons within a given model and dataset. The divergence tables cover all ten items, while the gap tables cover the nine directional items, since \textit{Current Economy} has no socially desirable pole, and Reverse-coded covers only the six eligible items throughout.


\begin{table*}[h]\centering
\caption{Per-item JS-divergence at Replicate and change $\Delta\mathrm{JSD}$ (condition $-$ Replicate), Llama-3.1-8B (ANES~2020). Negative values indicate closer alignment with the human distribution; \textsuperscript{ns} marks a non-significant change. Notation and significance markers as in Appendix \ref{sec:full-jsd-gap-tables}.}
\label{tab:paired-tests-llama8b}
\small
\setlength{\tabcolsep}{3pt}
\renewcommand{\cellalign}{cc}
\begin{tabular}{lccccc}
\toprule
Question & Replicate $\mathrm{JSD}$ & Reformulated & Reverse-coded & Priming & Preamble \\
\midrule
Climate Change    & \dcell{0.3631}{0.3502}{0.3768} & \dcell{-0.0242}{-0.0313}{-0.0170} & \dcell{-0.2489}{-0.2674}{-0.2303} & \dcell{+0.0510}{+0.0453}{+0.0564} & \dcell[\textsuperscript{ns}]{+0.0062}{-0.0010}{+0.0132} \\
Current Economy   & \dcell{0.2566}{0.2450}{0.2685} & \dcell{-0.1343}{-0.1431}{-0.1252} & N/A & \dcell{+0.0392}{+0.0299}{+0.0484} & \dcell{-0.0541}{-0.0632}{-0.0448} \\
Refugee Allowing  & \dcell{0.2215}{0.2109}{0.2322} & \dcell{-0.0267}{-0.0323}{-0.0214} & \dcell{+0.0512}{+0.0342}{+0.0681} & \dcell{-0.0176}{-0.0234}{-0.0118} & \dcell{-0.0375}{-0.0439}{-0.0314} \\
Health Insurance  & \dcell{0.2265}{0.2151}{0.2386} & \dcell{-0.0516}{-0.0602}{-0.0431} & N/A & \dcell{-0.0175}{-0.0253}{-0.0100} & \dcell{-0.0132}{-0.0219}{-0.0045} \\
Income Inequality & \dcell{0.1447}{0.1334}{0.1564} & \dcell[\textsuperscript{ns}]{-0.0014}{-0.0114}{+0.0083} & \dcell{-0.0826}{-0.0977}{-0.0676} & \dcell{+0.0311}{+0.0205}{+0.0415} & \dcell{-0.0142}{-0.0248}{-0.0038} \\
Race Diversity    & \dcell{0.1219}{0.1117}{0.1326} & \dcell{-0.0143}{-0.0244}{-0.0042} & \dcell{+0.0474}{+0.0373}{+0.0576} & \dcell{-0.1198}{-0.1300}{-0.1095} & \dcell{-0.0973}{-0.1088}{-0.0858} \\
Drug Addiction    & \dcell{0.1548}{0.1458}{0.1642} & \dcell{-0.0360}{-0.0444}{-0.0273} & N/A & \dcell{-0.0119}{-0.0189}{-0.0049} & \dcell{-0.0549}{-0.0624}{-0.0473} \\
Gay Marriage      & \dcell{0.1376}{0.1288}{0.1468} & \dcell{-0.1083}{-0.1166}{-0.1002} & \dcell{+0.0345}{+0.0292}{+0.0395} & \dcell{+0.0326}{+0.0269}{+0.0383} & \dcell{+0.0345}{+0.0292}{+0.0395} \\
Gun Regulation    & \dcell{0.0909}{0.0814}{0.1008} & \dcell{-0.0180}{-0.0279}{-0.0084} & N/A & \dcell{+0.1014}{+0.0902}{+0.1127} & \dcell{+0.0908}{+0.0795}{+0.1025} \\
Gender Role       & \dcell{0.0832}{0.0749}{0.0922} & \dcell{+0.0442}{+0.0335}{+0.0550} & \dcell{+0.3546}{+0.3398}{+0.3699} & \dcell{+0.0298}{+0.0198}{+0.0400} & \dcell{+0.0551}{+0.0450}{+0.0651} \\
\bottomrule
\end{tabular}
\end{table*}


\begin{table}[t]\centering
\caption{Per-item change in the absolute desirability gap $\Delta|D|$ (condition $-$ Replicate), Llama-3.1-8B (ANES~2020). The Replicate column reports the signed baseline gap $D$; negative values indicate a narrowing gap; \textsuperscript{ns} marks a non-significant change. Notation and significance markers as in \autoref{sec:full-jsd-gap-tables}.}
\label{tab:peritem-absD-8b}
\small
\setlength{\tabcolsep}{3pt}
\renewcommand{\cellalign}{cc}
\begin{tabular}{lccccc}
\toprule
Item & Replicate $D$ & Reformulated & Reverse-coded & Priming & Preamble \\
\midrule
Gay Marriage      & \dcell{+0.288}{+0.276}{+0.301} & \dcell{-0.188}{-0.199}{-0.176} & \dcell{+0.016}{+0.013}{+0.020} & \dcell{+0.016}{+0.012}{+0.019} & \dcell{+0.016}{+0.013}{+0.020} \\
Refugee Allowing  & \dcell{+0.417}{+0.402}{+0.432} & \dcell{-0.055}{-0.065}{-0.046} & \dcell{-0.220}{-0.251}{-0.188} & \dcell{-0.029}{-0.037}{-0.021} & \dcell{-0.085}{-0.096}{-0.075} \\
Income Inequality & \dcell{+0.404}{+0.387}{+0.421} & \dcell{-0.118}{-0.133}{-0.104} & \dcell{-0.273}{-0.305}{-0.241} & \dcell[\textsuperscript{ns}]{+0.009}{-0.004}{+0.021} & \dcell{-0.028}{-0.041}{-0.015} \\
Gender Role       & \dcell{+0.240}{+0.226}{+0.254} & \dcell{+0.060}{+0.043}{+0.078} & \dcell{+0.443}{+0.426}{+0.460} & \dcell{-0.092}{-0.108}{-0.077} & \dcell{+0.031}{+0.014}{+0.049} \\
Climate Change    & \dcell{-0.083}{-0.093}{-0.072} & \dcell{+0.027}{+0.022}{+0.032} & \dcell{+0.066}{+0.045}{+0.087} & \dcell{-0.027}{-0.031}{-0.022} & \dcell{-0.014}{-0.019}{-0.010} \\
Gun Regulation    & \dcell{-0.325}{-0.343}{-0.307} & \dcell{-0.032}{-0.047}{-0.016} & N/A & \dcell{+0.127}{+0.114}{+0.140} & \dcell{+0.118}{+0.105}{+0.131} \\
Drug Addiction    & \dcell{+0.297}{+0.284}{+0.310} & \dcell{-0.119}{-0.128}{-0.110} & N/A & \dcell{-0.097}{-0.105}{-0.089} & \dcell{-0.074}{-0.082}{-0.066} \\
Race Diversity    & \dcell{+0.352}{+0.336}{+0.368} & \dcell{-0.067}{-0.079}{-0.054} & \dcell{+0.039}{+0.029}{+0.049} & \dcell{-0.336}{-0.350}{-0.321} & \dcell{-0.341}{-0.363}{-0.310} \\
Health Insurance  & \dcell{+0.423}{+0.409}{+0.438} & \dcell{-0.093}{-0.102}{-0.084} & N/A & \dcell{-0.025}{-0.031}{-0.019} & \dcell{-0.009}{-0.014}{-0.003} \\
\bottomrule
\end{tabular}
\end{table}

\begin{table*}[h]\centering
\caption{Per-item JS-divergence at Replicate and change $\Delta\mathrm{JSD}$ (condition $-$ Replicate), Llama-3.1-70B (ANES~2020). Negative values indicate closer alignment with the human distribution; \textsuperscript{ns} marks a non-significant change. Notation and significance markers as in \autoref{sec:full-jsd-gap-tables}.}
\label{tab:paired-tests-llama70b}
\small
\setlength{\tabcolsep}{3pt}
\renewcommand{\cellalign}{cc}
\begin{tabular}{lccccc}
\toprule
Item & Replicate $\mathrm{JSD}$ & Reformulated & Reverse-coded & Priming & Preamble \\
\midrule
Climate Change    & \dcell{0.1391}{0.1288}{0.1505} & \dcell{+0.0776}{+0.0664}{+0.0890} & \dcell{-0.0866}{-0.0991}{-0.0745} & \dcell{+0.0385}{+0.0275}{+0.0495} & \dcell{+0.0149}{+0.0049}{+0.0252} \\
Current Economy   & \dcell{0.1497}{0.1408}{0.1592} & \dcell{+0.0697}{+0.0599}{+0.0797} & N/A & \dcell{+0.0510}{+0.0426}{+0.0599} & \dcell{+0.0183}{+0.0101}{+0.0260} \\
Refugee Allowing  & \dcell{0.1835}{0.1748}{0.1923} & \dcell{-0.1380}{-0.1439}{-0.1321} & \dcell{+0.0156}{+0.0095}{+0.0218} & \dcell{+0.0077}{+0.0025}{+0.0133} & \dcell{-0.0015}{-0.0030}{-0.0002} \\
Health Insurance  & \dcell{0.2010}{0.1915}{0.2107} & \dcell{-0.0209}{-0.0258}{-0.0162} & N/A & \dcell[\textsuperscript{ns}]{-0.0031}{-0.0069}{+0.0007} & \dcell{-0.0175}{-0.0214}{-0.0139} \\
Income Inequality & \dcell{0.1463}{0.1384}{0.1545} & \dcell[\textsuperscript{ns}]{-0.0019}{-0.0053}{+0.0016} & \dcell{-0.0163}{-0.0215}{-0.0109} & \dcell{-0.0092}{-0.0129}{-0.0058} & \dcell{+0.0144}{+0.0106}{+0.0184} \\
Race Diversity    & \dcell{0.2045}{0.1940}{0.2152} & \dcell{+0.0153}{+0.0058}{+0.0246} & \dcell{+0.0553}{+0.0473}{+0.0631} & \dcell{-0.0264}{-0.0351}{-0.0179} & \dcell{+0.0122}{+0.0064}{+0.0179} \\
Drug Addiction    & \dcell{0.1719}{0.1635}{0.1805} & \dcell{-0.0505}{-0.0563}{-0.0445} & N/A & \dcell[\textsuperscript{ns}]{+0.0008}{+0.0000}{+0.0020} & \dcell[\textsuperscript{ns}]{+0.0008}{+0.0000}{+0.0020} \\
Gay Marriage      & \dcell{0.0623}{0.0549}{0.0705} & \dcell{+0.0813}{+0.0712}{+0.0916} & \dcell{+0.0427}{+0.0295}{+0.0557} & \dcell[\textsuperscript{ns}]{-0.0031}{-0.0106}{+0.0043} & \dcell[\textsuperscript{ns}]{-0.0029}{-0.0107}{+0.0046} \\
Gun Regulation    & \dcell{0.1556}{0.1444}{0.1675} & \dcell{-0.0288}{-0.0396}{-0.0182} & N/A & \dcell[\textsuperscript{ns}]{+0.0035}{-0.0076}{+0.0145} & \dcell{+0.0408}{+0.0292}{+0.0523} \\
Gender Role       & \dcell{0.0073}{0.0049}{0.0104} & \dcell{+0.1112}{+0.1013}{+0.1215} & \dcell{+0.7133}{+0.6994}{+0.7275} & \dcell{+0.1253}{+0.1145}{+0.1360} & \dcell[\textsuperscript{ns}]{+0.0034}{+0.0001}{+0.0068} \\
\bottomrule
\end{tabular}
\end{table*}

\begin{table}[t]
\centering
\caption{Per-item change in the absolute desirability gap $\Delta|D|$ (condition $-$ Replicate), Llama-3.1-70B (ANES~2020). The Replicate column reports the signed baseline gap $D$; negative values indicate a narrowing gap; \textsuperscript{ns} marks a non-significant change. Notation and significance markers as in Appendix \ref{sec:full-jsd-gap-tables}.}
\label{tab:peritem-absD-70b}
\small
\setlength{\tabcolsep}{3pt}
\renewcommand{\cellalign}{cc}
\begin{tabular}{lccccc}
\toprule
Item & Replicate $D$ & Reformulated & Reverse-coded & Priming & Preamble \\
\midrule
Gay Marriage      & \dcell{-0.169}{-0.187}{-0.151} & \dcell{+0.187}{+0.168}{+0.205} & \dcell{-0.123}{-0.153}{-0.093} & \dcell{-0.037}{-0.056}{-0.019} & \dcell{-0.069}{-0.088}{-0.051} \\
Refugee Allowing  & \dcell{+0.195}{+0.176}{+0.213} & \dcell{-0.042}{-0.060}{-0.026} & \dcell{+0.182}{+0.168}{+0.196} & \dcell{-0.106}{-0.124}{-0.089} & \dcell{+0.020}{+0.003}{+0.036} \\
Income Inequality & \dcell{+0.261}{+0.243}{+0.280} & \dcell[\textsuperscript{ns}]{-0.000}{-0.017}{+0.016} & \dcell[\textsuperscript{ns}]{-0.005}{-0.022}{+0.012} & \dcell{-0.062}{-0.079}{-0.045} & \dcell{+0.066}{+0.050}{+0.082} \\
Gender Role       & \dcell{+0.052}{+0.042}{+0.063} & \dcell{+0.184}{+0.169}{+0.199} & \dcell{+0.825}{+0.814}{+0.836} & \dcell[\textsuperscript{ns}]{+0.000}{-0.015}{+0.016} & \dcell[\textsuperscript{ns}]{-0.011}{-0.022}{+0.000} \\
Climate Change    & \dcell{-0.105}{-0.116}{-0.093} & \dcell{+0.084}{+0.076}{+0.093} & \dcell{-0.058}{-0.080}{-0.036} & \dcell{-0.048}{-0.056}{-0.039} & \dcell{-0.012}{-0.021}{-0.003} \\
Gun Regulation    & \dcell{+0.092}{+0.073}{+0.111} & \dcell{+0.035}{+0.017}{+0.053} & N/A & \dcell{+0.032}{+0.014}{+0.050} & \dcell{+0.048}{+0.031}{+0.067} \\
Drug Addiction    & \dcell{+0.305}{+0.292}{+0.319} & \dcell{-0.031}{-0.036}{-0.026} & N/A & \dcell[\textsuperscript{ns}]{+0.000}{+0.000}{+0.001} & \dcell[\textsuperscript{ns}]{+0.000}{+0.000}{+0.001} \\
Race Diversity    & \dcell{+0.322}{+0.306}{+0.339} & \dcell{-0.184}{-0.199}{-0.169} & \dcell{+0.118}{+0.109}{+0.126} & \dcell{+0.044}{+0.033}{+0.055} & \dcell{+0.027}{+0.016}{+0.039} \\
Health Insurance  & \dcell{+0.099}{+0.080}{+0.118} & \dcell{+0.158}{+0.142}{+0.175} & N/A & \dcell[\textsuperscript{ns}]{+0.014}{-0.003}{+0.032} & \dcell{+0.110}{+0.093}{+0.127} \\
\bottomrule
\end{tabular}
\end{table}

\begin{table*}[h]\centering
\caption{Per-item JS-divergence at Replicate and change $\Delta\mathrm{JSD}$ (condition $-$ Replicate), GPT-4.1-mini (ANES~2020). Negative values indicate closer alignment with the human distribution; \textsuperscript{ns} marks a non-significant change. Notation and significance markers as in Appendix \ref{sec:full-jsd-gap-tables}.}
\label{tab:paired-tests-gpt41mini}
\small
\setlength{\tabcolsep}{3pt}
\renewcommand{\cellalign}{cc}
\begin{tabular}{lccccc}
\toprule
Item & Replicate $\mathrm{JSD}$ & Reformulated & Reverse-coded & Priming & Preamble \\
\midrule
Climate Change    & \dcell{0.0536}{0.0472}{0.0607} & \dcell{-0.0323}{-0.0406}{-0.0240} & \dcell{+0.2666}{+0.2506}{+0.2822} & \dcell{+0.1460}{+0.1358}{+0.1561} & \dcell{+0.0224}{+0.0150}{+0.0299} \\
Current Economy   & \dcell{0.1958}{0.1848}{0.2071} & \dcell{-0.0494}{-0.0608}{-0.0375} & N/A & \dcell{+0.0679}{+0.0578}{+0.0781} & \dcell{-0.0112}{-0.0203}{-0.0021} \\
Refugee Allowing  & \dcell{0.0816}{0.0730}{0.0907} & \dcell{-0.0624}{-0.0700}{-0.0550} & \dcell{-0.0459}{-0.0534}{-0.0385} & \dcell{+0.0120}{+0.0030}{+0.0211} & \dcell{+0.1331}{+0.1229}{+0.1434} \\
Health Insurance  & \dcell{0.0342}{0.0285}{0.0407} & \dcell{+0.0235}{+0.0169}{+0.0303} & N/A & \dcell{+0.0314}{+0.0241}{+0.0386} & \dcell{+0.1237}{+0.1141}{+0.1332} \\
Income Inequality & \dcell{0.0371}{0.0312}{0.0436} & \dcell{+0.0474}{+0.0390}{+0.0560} & \dcell{-0.0323}{-0.0383}{-0.0265} & \dcell{+0.0344}{+0.0270}{+0.0419} & \dcell{+0.0382}{+0.0307}{+0.0459} \\
Race Diversity    & \dcell{0.2273}{0.2153}{0.2397} & \dcell{-0.0725}{-0.0843}{-0.0612} & \dcell{-0.1486}{-0.1584}{-0.1388} & \dcell{+0.0390}{+0.0318}{+0.0459} & \dcell{-0.0106}{-0.0204}{-0.0009} \\
Drug Addiction    & \dcell{0.1629}{0.1541}{0.1720} & \dcell{-0.0236}{-0.0293}{-0.0181} & N/A & \dcell{+0.0099}{+0.0068}{+0.0128} & \dcell{+0.0099}{+0.0068}{+0.0128} \\
Gay Marriage      & \dcell{0.0363}{0.0308}{0.0424} & \dcell{-0.0069}{-0.0125}{-0.0014} & \dcell{+0.0590}{+0.0541}{+0.0637} & \dcell{-0.0126}{-0.0182}{-0.0070} & \dcell{+0.0107}{+0.0045}{+0.0171} \\
Gun Regulation    & \dcell{0.0018}{0.0007}{0.0035} & \dcell{+0.0141}{+0.0099}{+0.0185} & N/A & \dcell{+0.0067}{+0.0040}{+0.0096} & \dcell{+0.0099}{+0.0070}{+0.0131} \\
Gender Role       & \dcell{0.0696}{0.0616}{0.0782} & \dcell{-0.0587}{-0.0675}{-0.0501} & \dcell{+0.7763}{+0.7580}{+0.7943} & \dcell{+0.1118}{+0.1036}{+0.1199} & \dcell{+0.0460}{+0.0375}{+0.0547} \\
\bottomrule
\end{tabular}
\end{table*}

\begin{table}[t]
\centering
\caption{Per-item change in the absolute desirability gap $\Delta|D|$ (condition $-$ Replicate), GPT-4.1-mini (ANES~2020). The Replicate column reports the signed baseline gap $D$; negative values indicate a narrowing gap; \textsuperscript{ns} marks a non-significant change. Notation and significance markers as in Appendix \ref{sec:full-jsd-gap-tables}.}
\label{tab:peritem-absD-gpt}
\small
\setlength{\tabcolsep}{4pt}
\renewcommand{\cellalign}{cc}
\begin{tabular}{lccccc}
\toprule
Item & Replicate $D$ & Reformulated & Reverse-coded & Priming & Preamble \\
\midrule
Gay Marriage      & \dcell{+0.150}{+0.134}{+0.166} & \dcell{-0.015}{-0.029}{-0.001} & \dcell{-0.027}{-0.041}{-0.013} & \dcell[\textsuperscript{ns}]{+0.001}{-0.012}{+0.015} & \dcell{+0.048}{+0.035}{+0.061} \\
Refugee Allowing  & \dcell{+0.286}{+0.269}{+0.304} & \dcell{-0.152}{-0.168}{-0.136} & \dcell{-0.085}{-0.100}{-0.070} & \dcell{+0.033}{+0.020}{+0.047} & \dcell{+0.143}{+0.132}{+0.154} \\
Income Inequality & \dcell{+0.223}{+0.204}{+0.242} & \dcell{-0.043}{-0.061}{-0.026} & \dcell{-0.168}{-0.186}{-0.150} & \dcell{+0.074}{+0.058}{+0.091} & \dcell{+0.071}{+0.055}{+0.088} \\
Gender Role       & \dcell{-0.048}{-0.055}{-0.041} & \dcell[\textsuperscript{ns}]{+0.009}{-0.006}{+0.025} & \dcell{+0.898}{+0.885}{+0.911} & \dcell{+0.006}{+0.004}{+0.008} & \dcell{+0.004}{+0.002}{+0.006} \\
Climate Change    & \dcell{+0.094}{+0.082}{+0.106} & \dcell{-0.044}{-0.066}{-0.022} & \dcell{+0.205}{+0.196}{+0.213} & \dcell{+0.142}{+0.135}{+0.150} & \dcell{+0.063}{+0.054}{+0.072} \\
Gun Regulation    & \dcell{+0.049}{+0.030}{+0.068} & \dcell[\textsuperscript{ns}]{-0.024}{-0.058}{+0.010} & N/A & \dcell{+0.029}{+0.010}{+0.047} & \dcell{+0.070}{+0.052}{+0.088} \\
Drug Addiction    & \dcell{+0.301}{+0.288}{+0.315} & \dcell{-0.062}{-0.069}{-0.055} & N/A & \dcell{+0.004}{+0.003}{+0.006} & \dcell{+0.004}{+0.003}{+0.006} \\
Race Diversity    & \dcell{+0.429}{+0.414}{+0.443} & \dcell{-0.158}{-0.168}{-0.147} & \dcell{-0.137}{-0.147}{-0.128} & \dcell{+0.014}{+0.011}{+0.017} & \dcell{-0.005}{-0.010}{-0.000} \\
Health Insurance  & \dcell{+0.198}{+0.180}{+0.216} & \dcell{+0.035}{+0.019}{+0.051} & N/A & \dcell{+0.075}{+0.060}{+0.090} & \dcell{+0.185}{+0.172}{+0.198} \\
\bottomrule
\end{tabular}
\end{table}

\subsection{JS-divergence and desirability gap results for ANES 2024}\label{sec:full-jad-gap-2024}
Table~\ref{tab:jsd-gap-gpt41mini-2024} reports the JS-divergence and signed desirability gap at the Replicate baseline for ANES 2024 using GPT-4.1-mini, together with the change in each under Reformulated. Intervals and significance follow the conventions of Appendix \ref{sec:full-jsd-gap-tables}, with Benjamini-Hochberg correction applied within each metric. All eight overlapping items are directional, so both metrics cover the same item set.


\begin{table}[t]\centering
\caption{JS-divergence and signed desirability gap at Replicate, and their changes under Reformulated, GPT-4.1-mini (\textbf{ANES 2024}). Negative $\Delta\mathrm{JSD}$ and $\Delta|D|$ indicate movement toward the human distribution; \textsuperscript{ns} marks a non-significant change. Notation and significance markers as in Appendix \ref{sec:full-jsd-gap-tables}.}
\label{tab:jsd-gap-gpt41mini-2024}
\small
\setlength{\tabcolsep}{4pt}
\renewcommand{\cellalign}{cc}
\begin{tabular}{lcc|cc}
\toprule
Question & Replicate $\mathrm{JSD}$ & $\Delta\mathrm{JSD}$ & Replicate $D$ & $\Delta|D|$ \\
\midrule
Climate Change    & \dcell{0.0811}{0.0739}{0.0890} & \dcell{-0.0514}{-0.0596}{-0.0430} & \dcell{+0.145}{+0.133}{+0.156} & \dcell{-0.142}{-0.150}{-0.126} \\
Health Insurance  & \dcell{0.0230}{0.0184}{0.0284} & \dcell{+0.0053}{+0.0004}{+0.0105} & \dcell{+0.170}{+0.151}{+0.188} & \dcell{+0.017}{+0.001}{+0.034} \\
Income Inequality & \dcell{0.0226}{0.0181}{0.0279} & \dcell{+0.0736}{+0.0645}{+0.0829} & \dcell{+0.173}{+0.154}{+0.192} & \dcell[\textsuperscript{ns}]{-0.010}{-0.028}{+0.008} \\
Race Diversity    & \dcell{0.2700}{0.2569}{0.2839} & \dcell{-0.1022}{-0.1147}{-0.0896} & \dcell{+0.507}{+0.492}{+0.521} & \dcell{-0.186}{-0.198}{-0.175} \\
Drug Addiction    & \dcell{0.1229}{0.1151}{0.1312} & \dcell{-0.0103}{-0.0171}{-0.0037} & \dcell{+0.242}{+0.230}{+0.255} & \dcell{-0.092}{-0.101}{-0.084} \\
Gay Marriage      & \dcell{0.0214}{0.0161}{0.0275} & \dcell{-0.0162}{-0.0207}{-0.0119} & \dcell{+0.156}{+0.134}{+0.178} & \dcell{-0.082}{-0.098}{-0.066} \\
Gun Regulation    & \dcell{0.0009}{0.0002}{0.0023} & \dcell{+0.0224}{+0.0176}{+0.0274} & \dcell{-0.034}{-0.054}{-0.015} & \dcell{+0.056}{+0.037}{+0.075} \\
Gender Role       & \dcell{0.0256}{0.0211}{0.0307} & \dcell[\textsuperscript{ns}]{-0.0071}{-0.0152}{+0.0010} & \dcell{-0.057}{-0.065}{-0.050} & \dcell{+0.030}{+0.013}{+0.047} \\
\bottomrule
\end{tabular}
\end{table}

\subsection{JS-divergence and desirability gap results for WVS Wave 7}\label{sec:full-jad-gap-wvs}
Table~\ref{tab:jsd-gap-wvs7-reformulated} reports the JS-divergence and signed desirability gap at the Replicate baseline for WVS Wave~7 using GPT-4.1-mini, together with the change in each under Reformulated, stacked by country: Netherlands, Germany, and Great Britain. Intervals and significance follow the conventions of Appendix \ref{sec:full-jsd-gap-tables}, with Benjamini-Hochberg correction applied within each metric and country. Government Responsibility and Income Equality are non-directional, so the gap columns cover eight of the ten items.

\begin{table}[p]\centering
\caption{JS-divergence and signed desirability gap at Replicate, and their changes under Reformulated, GPT-4.1-mini (\textbf{WVS Wave 7}). Negative $\Delta\mathrm{JSD}$ and $\Delta|D|$ indicate movement toward the human distribution; \textsuperscript{ns} marks a non-significant change; dashes mark the two non-directional items. Notation and significance markers as in Appendix \ref{sec:full-jsd-gap-tables}.}
\label{tab:jsd-gap-wvs7-reformulated}
\footnotesize
\setlength{\tabcolsep}{4pt}
\renewcommand{\cellalign}{cc}
\begin{tabular}{lcccc}
\toprule
Item & Replicate $\mathrm{JSD}$ & $\Delta\mathrm{JSD}$ & Replicate $D$ & $\Delta|D|$ \\
\midrule
\multicolumn{5}{l}{\textit{Netherlands}} \\
Environment vs Economy & \dcell{0.0114}{0.0066}{0.0178} & \dcell{-0.0084}{-0.0135}{-0.0039} & \dcell{+0.102}{+0.076}{+0.128} & \dcell{-0.056}{-0.080}{-0.032} \\
Refugee Asylum & \dcell{0.0847}{0.0734}{0.0970} & \dcell{-0.0104}{-0.0191}{-0.0012} & \dcell{+0.171}{+0.152}{+0.190} & \dcell{-0.005}{-0.009}{-0.000} \\
Government Responsibility & \dcell{0.0892}{0.0771}{0.1053} & \dcell{-0.0167}{-0.0304}{-0.0031} & --- & --- \\
Income Equality & \dcell{0.0976}{0.0855}{0.1131} & \dcell{-0.0283}{-0.0422}{-0.0142} & --- & --- \\
Immigration Impact & \dcell{0.1646}{0.1470}{0.1843} & \dcell{+0.0482}{+0.0279}{+0.0684} & \dcell{+0.177}{+0.165}{+0.188} & \dcell{+0.023}{+0.013}{+0.033} \\
Gay Parenting & \dcell{0.0735}{0.0616}{0.0878} & \dcell[\textsuperscript{ns}]{-0.0020}{-0.0151}{+0.0105} & \dcell{+0.071}{+0.058}{+0.085} & \dcell{+0.027}{+0.018}{+0.037} \\
Maternal Employment & \dcell{0.0891}{0.0793}{0.0997} & \dcell{-0.0318}{-0.0409}{-0.0222} & \dcell{+0.029}{+0.017}{+0.042} & \dcell{+0.054}{+0.044}{+0.064} \\
Immigration Policy & \dcell{0.2862}{0.2627}{0.3107} & \dcell{-0.0770}{-0.0969}{-0.0570} & \dcell{+0.165}{+0.154}{+0.176} & \dcell{-0.030}{-0.037}{-0.022} \\
Government Surveillance & \dcell{0.3321}{0.3095}{0.3569} & \dcell{-0.0805}{-0.1010}{-0.0602} & \dcell{-0.262}{-0.277}{-0.246} & \dcell[\textsuperscript{ns}]{-0.006}{-0.018}{+0.007} \\
Strong Leader & \dcell{0.3594}{0.3365}{0.3844} & \dcell{-0.1563}{-0.1750}{-0.1375} & \dcell{+0.336}{+0.319}{+0.353} & \dcell{-0.046}{-0.056}{-0.036} \\
\midrule
\multicolumn{5}{l}{\textit{Germany}} \\
Environment vs Economy & \dcell{0.0891}{0.0738}{0.1065} & \dcell{-0.0174}{-0.0325}{-0.0025} & \dcell{+0.267}{+0.240}{+0.294} & \dcell{-0.026}{-0.045}{-0.006} \\
Refugee Asylum & \dcell{0.0528}{0.0431}{0.0636} & \dcell[\textsuperscript{ns}]{-0.0010}{-0.0080}{+0.0056} & \dcell{+0.111}{+0.095}{+0.128} & \dcell[\textsuperscript{ns}]{-0.001}{-0.005}{+0.002} \\
Government Responsibility & \dcell{0.1295}{0.1122}{0.1513} & \dcell[\textsuperscript{ns}]{-0.0189}{-0.0387}{+0.0020} & --- & --- \\
Income Equality & \dcell{0.1943}{0.1742}{0.2190} & \dcell{-0.0312}{-0.0530}{-0.0099} & --- & --- \\
Immigration Impact & \dcell{0.1475}{0.1281}{0.1698} & \dcell{+0.0727}{+0.0493}{+0.0966} & \dcell{+0.172}{+0.158}{+0.187} & \dcell{+0.054}{+0.043}{+0.066} \\
Gay Parenting & \dcell{0.1131}{0.0968}{0.1328} & \dcell[\textsuperscript{ns}]{+0.0036}{-0.0129}{+0.0199} & \dcell{+0.141}{+0.123}{+0.158} & \dcell[\textsuperscript{ns}]{+0.003}{-0.008}{+0.014} \\
Maternal Employment & \dcell{0.1640}{0.1500}{0.1789} & \dcell[\textsuperscript{ns}]{-0.0007}{-0.0127}{+0.0126} & \dcell{+0.140}{+0.124}{+0.155} & \dcell{+0.061}{+0.050}{+0.073} \\
Immigration Policy & \dcell{0.1227}{0.1057}{0.1413} & \dcell{-0.0210}{-0.0354}{-0.0058} & \dcell{+0.062}{+0.050}{+0.074} & \dcell{-0.010}{-0.017}{-0.002} \\
Government Surveillance & \dcell{0.3513}{0.3236}{0.3817} & \dcell{-0.1368}{-0.1631}{-0.1100} & \dcell{-0.184}{-0.200}{-0.169} & \dcell{-0.044}{-0.055}{-0.032} \\
Strong Leader & \dcell{0.2516}{0.2310}{0.2747} & \dcell{-0.0907}{-0.1085}{-0.0729} & \dcell{+0.245}{+0.228}{+0.263} & \dcell{-0.022}{-0.032}{-0.012} \\
\midrule
\multicolumn{5}{l}{\textit{Great Britain}} \\
Environment vs Economy & \dcell{0.0432}{0.0344}{0.0533} & \dcell{-0.0203}{-0.0289}{-0.0122} & \dcell{+0.201}{+0.179}{+0.224} & \dcell{-0.050}{-0.069}{-0.031} \\
Refugee Asylum & \dcell{0.2307}{0.2176}{0.2445} & \dcell{-0.0094}{-0.0144}{-0.0042} & \dcell{+0.394}{+0.375}{+0.413} & \dcell{-0.004}{-0.007}{-0.002} \\
Government Responsibility & \dcell{0.1109}{0.0983}{0.1265} & \dcell{-0.0257}{-0.0391}{-0.0120} & --- & --- \\
Income Equality & \dcell{0.1807}{0.1648}{0.1993} & \dcell{-0.0393}{-0.0558}{-0.0228} & --- & --- \\
Immigration Impact & \dcell{0.1096}{0.0959}{0.1252} & \dcell{-0.0321}{-0.0455}{-0.0185} & \dcell{+0.034}{+0.024}{+0.045} & \dcell[\textsuperscript{ns}]{-0.006}{-0.014}{+0.003} \\
Gay Parenting & \dcell{0.0954}{0.0827}{0.1097} & \dcell[\textsuperscript{ns}]{+0.0061}{-0.0073}{+0.0191} & \dcell{+0.066}{+0.054}{+0.078} & \dcell{+0.021}{+0.012}{+0.029} \\
Maternal Employment & \dcell{0.1203}{0.1108}{0.1304} & \dcell{-0.0113}{-0.0199}{-0.0020} & \dcell{+0.092}{+0.080}{+0.104} & \dcell{+0.055}{+0.046}{+0.064} \\
Immigration Policy & \dcell{0.0895}{0.0801}{0.0999} & \dcell{-0.0104}{-0.0168}{-0.0044} & \dcell{+0.014}{+0.005}{+0.024} & \dcell[\textsuperscript{ns}]{-0.013}{-0.018}{+0.002} \\
Government Surveillance & \dcell{0.3468}{0.3248}{0.3699} & \dcell{-0.1022}{-0.1227}{-0.0815} & \dcell{-0.309}{-0.324}{-0.294} & \dcell{-0.037}{-0.048}{-0.027} \\
Strong Leader & \dcell{0.2664}{0.2496}{0.2846} & \dcell{-0.1409}{-0.1558}{-0.1262} & \dcell{+0.257}{+0.243}{+0.270} & \dcell{-0.051}{-0.060}{-0.042} \\
\bottomrule
\end{tabular}
\end{table}

\clearpage
\subsection{Response distributions for ANES 2020}
\begin{figure}[!h] 
    \centering
    \begin{subfigure}[t]{0.32\textwidth} 
        \centering
        \includegraphics[width=\linewidth]{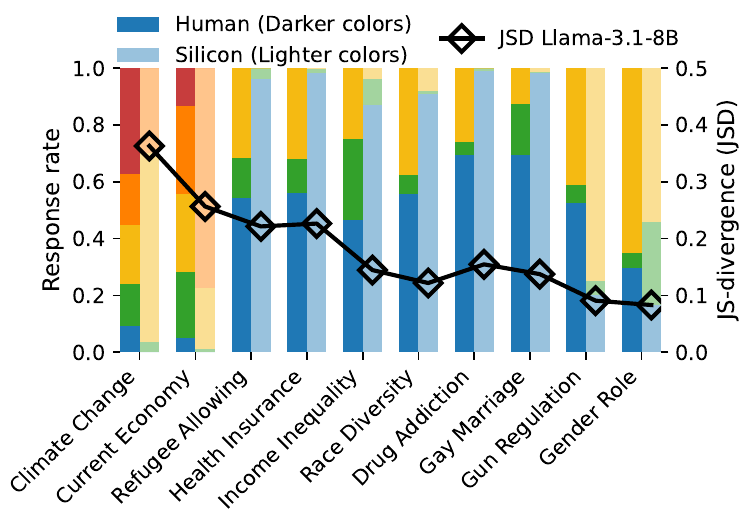}\vspace{-5pt}
        \caption{Human vs.\ Llama-3.1-8B 
        }
        \label{fig:replicate-llama8b}
    \end{subfigure}
    \begin{subfigure}[t]{0.32\textwidth} 
        \centering
        \includegraphics[width=\linewidth]{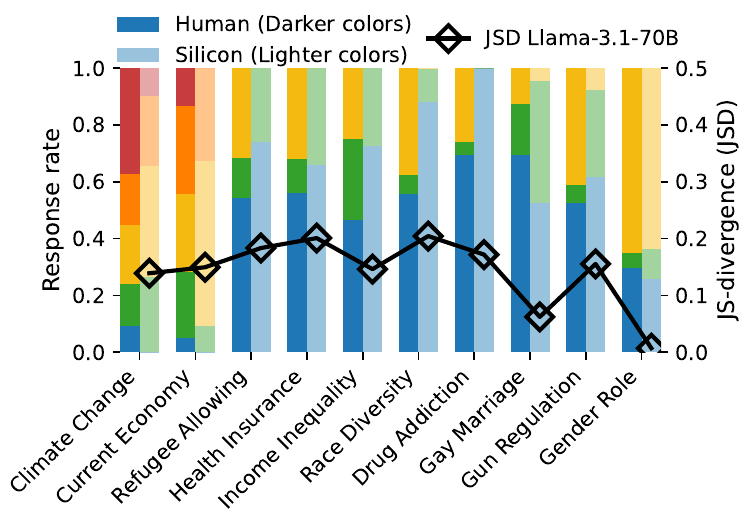}\vspace{-5pt}
        \caption{Human vs.\ Llama-3.1-70B 
        }
        \label{fig:replicate-llama70b}
    \end{subfigure}
    \begin{subfigure}[t]{0.32\textwidth}
        \centering
        \includegraphics[width=\linewidth]{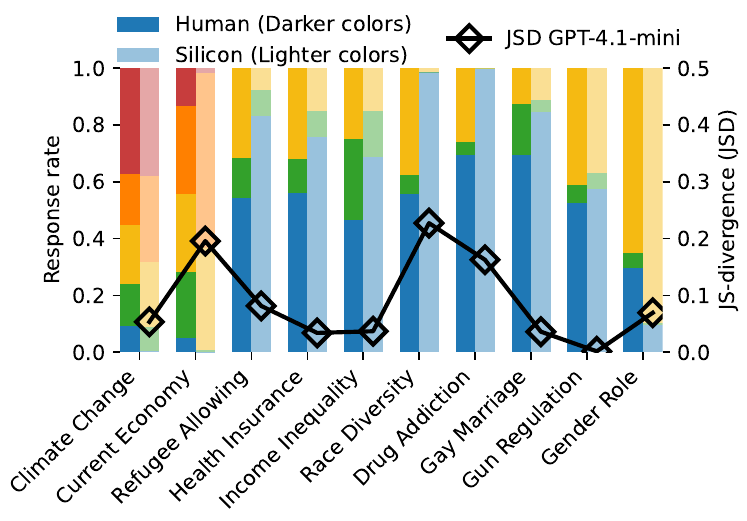}\vspace{-5pt}
        \caption{Human vs.\ GPT-4.1-mini 
        }
        \label{fig:replicate-gpt4.1}
    \end{subfigure}\vspace{-5pt}
    \caption{Baseline ({Replicate}) response distributions for the ten items, ANES~2020. For each item, bar color denotes the answer option in its fixed survey order and does not encode social desirability; darker and lighter shades of the same color denote human and silicon respondents, respectively. The black markers (right axis) give the JS-divergence between the human and silicon distributions.}
    \label{fig:combined-replication-baseline}
\end{figure}

\paragraph{Replicate}
\label{sec:full-distr-replicate}
The full response distributions for the Replicate condition are shown in \autoref{fig:combined-replication-baseline}.
The smallest model, Llama-8B, shows the largest departure from the human
distributions. On sensitive items such as \textit{Refugee Allowing},
\textit{Health Insurance}, \textit{Drug Addiction}, and \textit{Gay Marriage},
its responses concentrate on the socially desirable option (lighter bars), where
human responses are more dispersed. This concentration on a single desirable
option, rather than the spread humans show, is consistent with a strong
desirability effect in this model.

Llama-70B has generally lower JS-divergence than Llama-8B (Table~\ref{tab:baseline}), but its distributions depart from humans in a different way. On several items (e.g., \textit{Refugee Allowing}, \textit{Health Insurance}), it places substantial mass on both the socially desirable (lighter blue) and the socially undesirable (lighter green)  options, where human respondents more often choose the moderate (yellow) response (e.g., ``No change'' or ``Neither favor nor oppose''). This bimodal pattern, rather than concentration on a single option, distinguishes 70B from 8B: the model places mass on both poles instead of the moderate response, a polarized rather than one-sided departure from the human distribution.

GPT-4.1-mini shifts toward the socially desirable option on most items, producing distributions that are narrower and more concentrated on the desirable option than those in ANES 2020. On sensitive items such as \textit{Race Diversity} and \textit{Drug Addiction}, its simulated respondents concentrate on a single option almost entirely, whereas the human responses are spread across several options.

\begin{figure}[!t] 
    \centering
    \begin{subfigure}[]{0.48\textwidth} 
        \centering
        \includegraphics[width=\linewidth]{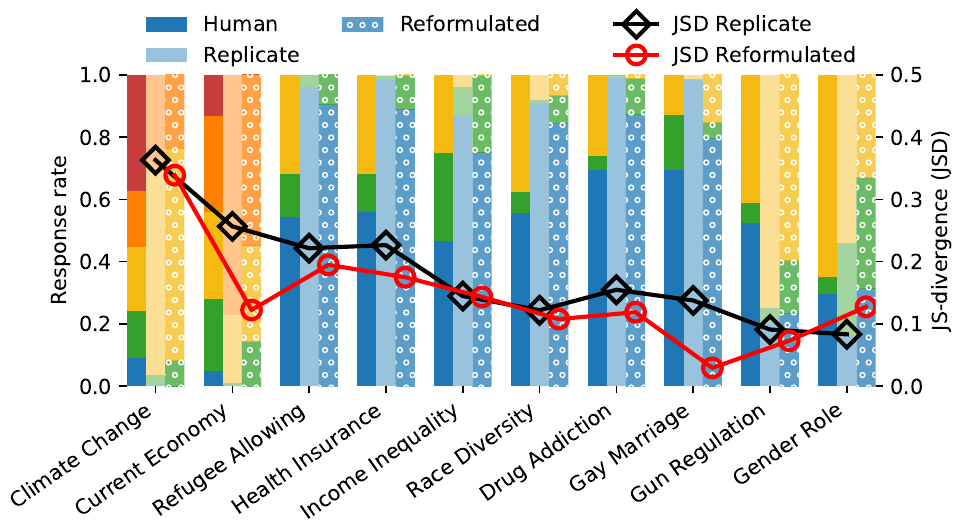}\vspace{-5pt}
        \caption{Llama-3.1-8B
        }
        \label{fig:reformulated-bars-llama8b}
    \end{subfigure}
    \begin{subfigure}[]{0.48\textwidth} 
        \centering
        \includegraphics[width=\linewidth]{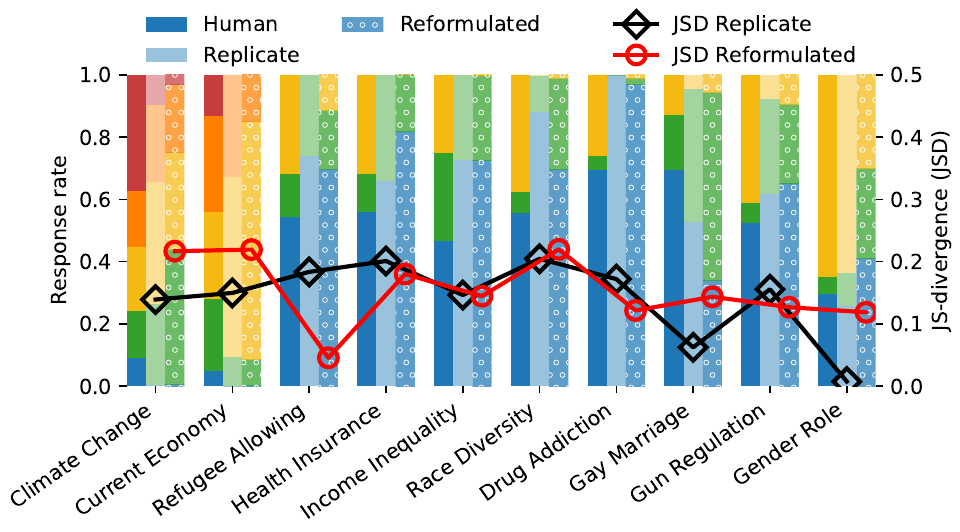}\vspace{-5pt}
        \caption{Llama-3.1-70B
        }
        \label{fig:reformulated-bars-llama70b}
    \end{subfigure}
    \hfill 
    \begin{subfigure}[]{0.48\textwidth}
        \centering
        \includegraphics[width=\linewidth]{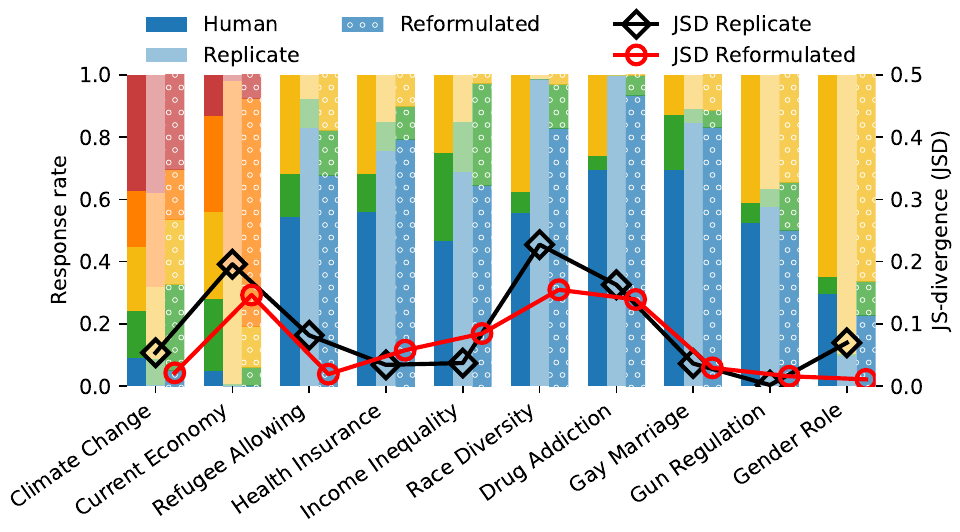}\vspace{-5pt}
        \caption{GPT-4.1-mini
        }
        \label{fig:reformulated-bars-gpt4.1}
    \end{subfigure}\vspace{-5pt}
    \caption{Human, Replicate, and \textbf{Reformulated} responses for all items on three models.}
    \label{fig:combined-reformulated-bars}
\end{figure}

\begin{figure}[!t] 
    \centering
    \begin{subfigure}[]{0.33\textwidth} 
        \centering
        \includegraphics[width=\linewidth]{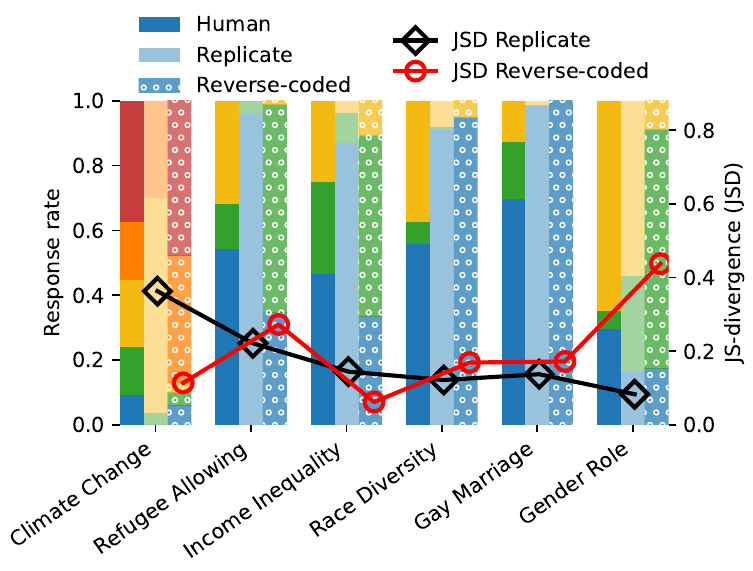}\vspace{-5pt}
        \caption{Llama-3.1-8B
        }
        \label{fig:reversed-bars-llama8b}
    \end{subfigure}
    \begin{subfigure}[]{0.32\textwidth} 
        \centering
        \includegraphics[width=\linewidth]{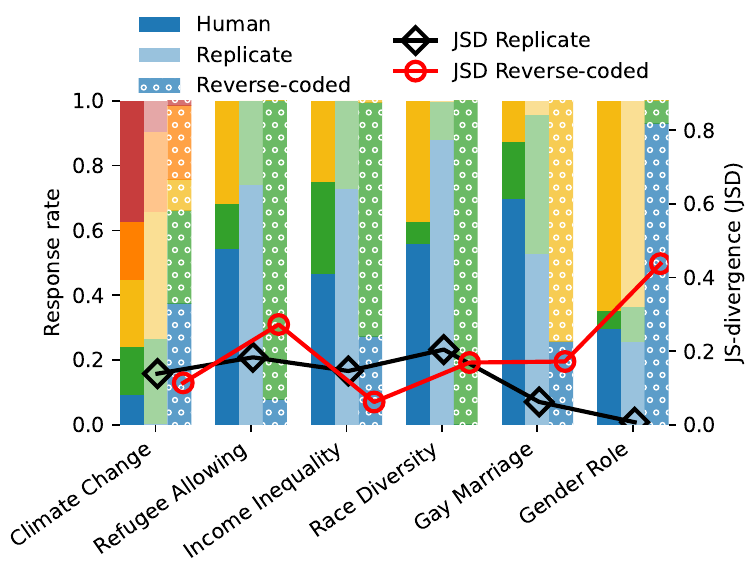}\vspace{-5pt}
        \caption{Llama-3.1-70B
        }
        \label{fig:reversed-bars-llama70b}
    \end{subfigure}
    \begin{subfigure}[]{0.33\textwidth}
        \centering
        \includegraphics[width=\linewidth]{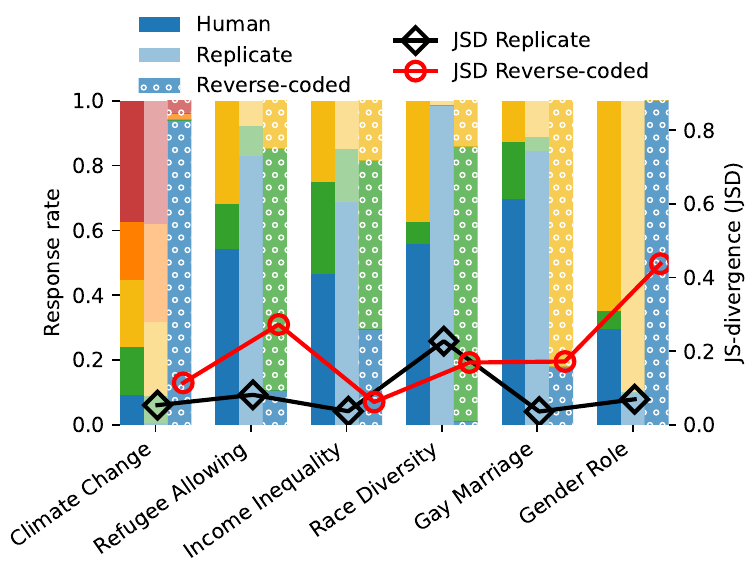}\vspace{-5pt}
        \caption{GPT-4.1-mini
        }
        \label{fig:reversed-bars-gpt4.1}
    \end{subfigure}\vspace{-5pt}
    \caption{Human, Replicate, and \textbf{Reverse-coded} responses for all items on three models.}
    \label{fig:combined-reversed-bars}
\end{figure}

\begin{figure}[!t] 
    \centering
    \begin{subfigure}[]{0.48\textwidth} 
        \centering
        \includegraphics[width=\linewidth]{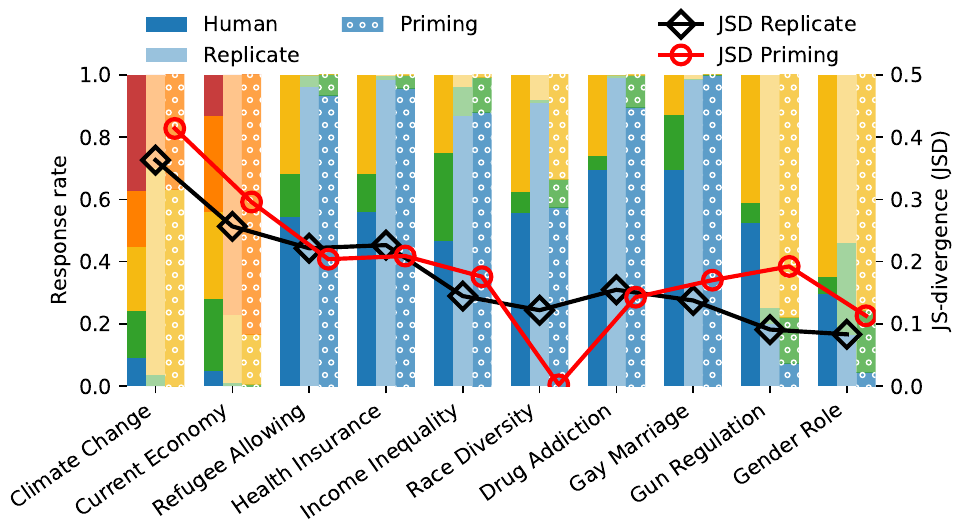}\vspace{-5pt}
        \caption{Llama-3.1-8B
        }
        \label{fig:priming-bars-llama8b}
    \end{subfigure}
    \begin{subfigure}[]{0.48\textwidth} 
        \centering
        \includegraphics[width=\linewidth]{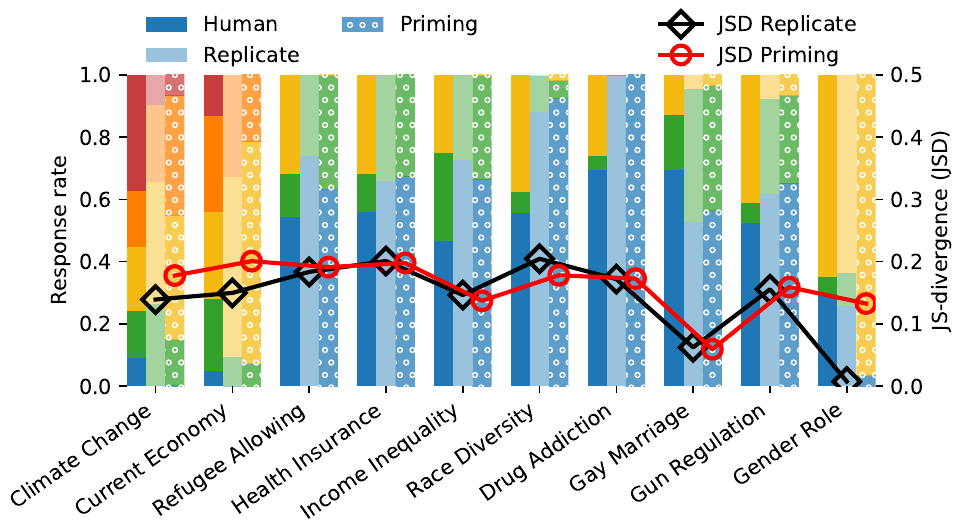}\vspace{-5pt}
        \caption{Llama-3.1-70B
        }
        \label{fig:priming-bars-llama70b}
    \end{subfigure}
    \hfill 
    \begin{subfigure}[]{0.48\textwidth}
        \centering
        \includegraphics[width=\linewidth]{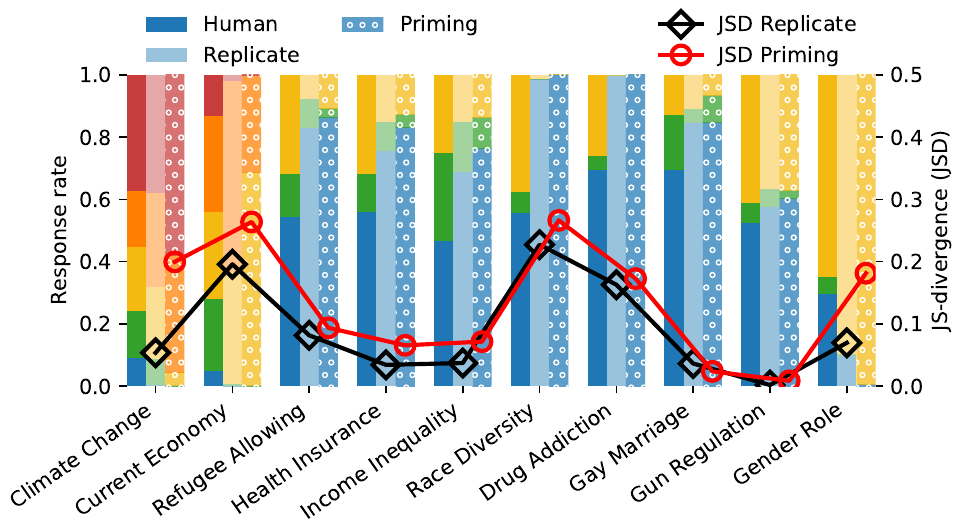}\vspace{-5pt}
        \caption{GPT-4.1-mini
        }
        \label{fig:priming-bars-gpt4.1}
    \end{subfigure}\vspace{-5pt}
    \caption{Human, Replicate, and \textbf{Priming} responses for all items on three models.}
    \label{fig:combined-priming-bars}
\end{figure}

\begin{figure}[!t] 
    \centering
    \begin{subfigure}[]{0.48\textwidth} 
        \centering
        \includegraphics[width=\linewidth]{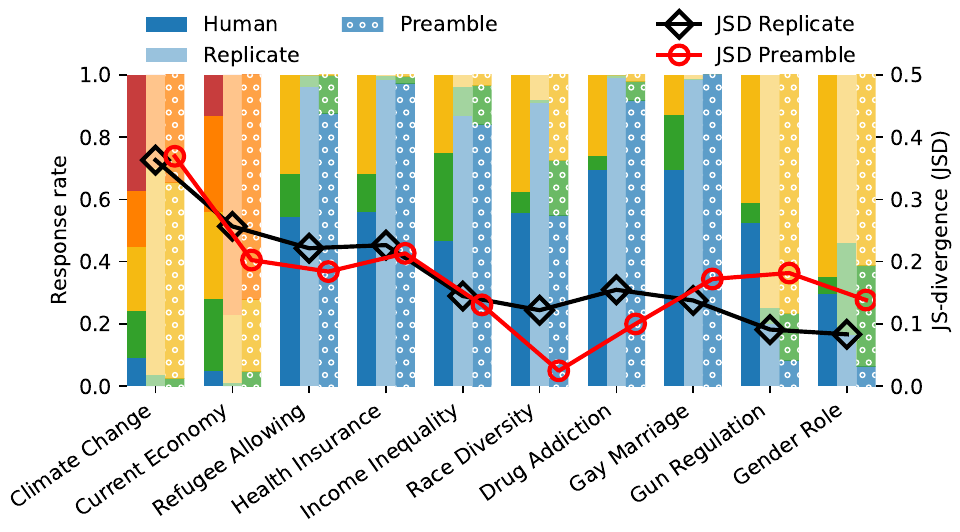}\vspace{-5pt}
        \caption{Llama-3.1-8B
        }
        \label{fig:preamble-bars-llama8b}
    \end{subfigure}
    \begin{subfigure}[]{0.48\textwidth} 
        \centering
        \includegraphics[width=\linewidth]{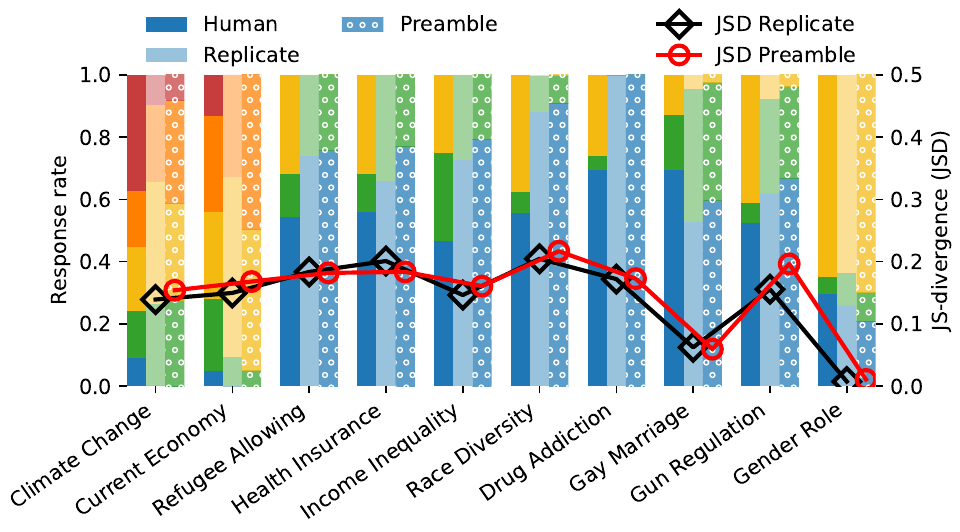}\vspace{-5pt}
        \caption{Llama-3.1-70B
        }
        \label{fig:preamble-bars-llama70b}
    \end{subfigure}
    \hfill 
    \begin{subfigure}[]{0.48\textwidth}
        \centering
        \includegraphics[width=\linewidth]{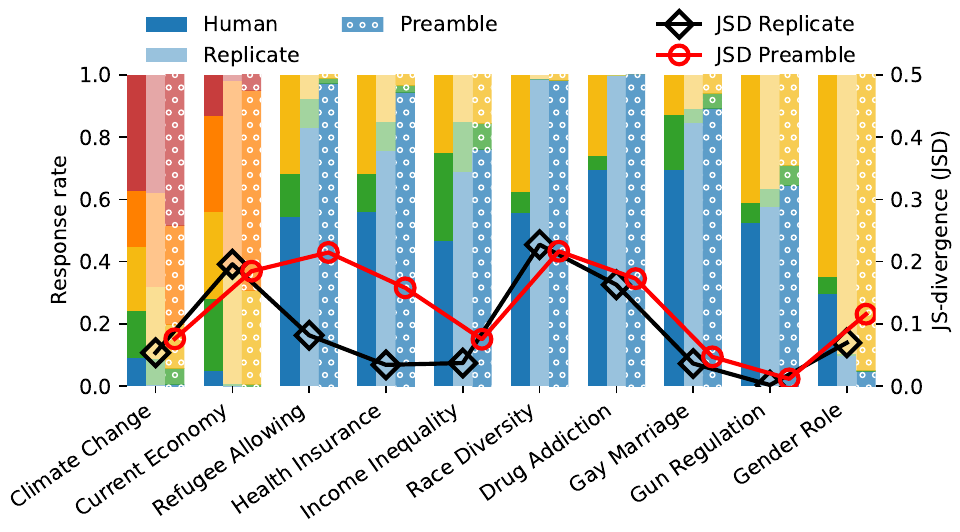}\vspace{-5pt}
        \caption{GPT-4.1-mini
        }
        \label{fig:preamble-bars-gpt4.1}
    \end{subfigure}\vspace{-5pt}
    \caption{Human, Replicate, and \textbf{Preamble} responses for all items on three models.}
    \label{fig:combined-preamble-bars}
\end{figure}

\paragraph{Reformulated, Reverse-coded, Priming and Preamble}
\label{sec:full-distr-4cond}
The full response distributions across the four conditions are shown in \autoref{fig:combined-reformulated-bars}, \autoref{fig:combined-reversed-bars}, \autoref{fig:combined-priming-bars}, and \autoref{fig:combined-preamble-bars}.

\subsection{Response distributions for WVS Wave 7 }\label{sec:wvs-distributions}
\begin{figure}[h] 
    \centering
    \begin{subfigure}[]{0.48\textwidth} 
        \centering
        \includegraphics[width=\linewidth]{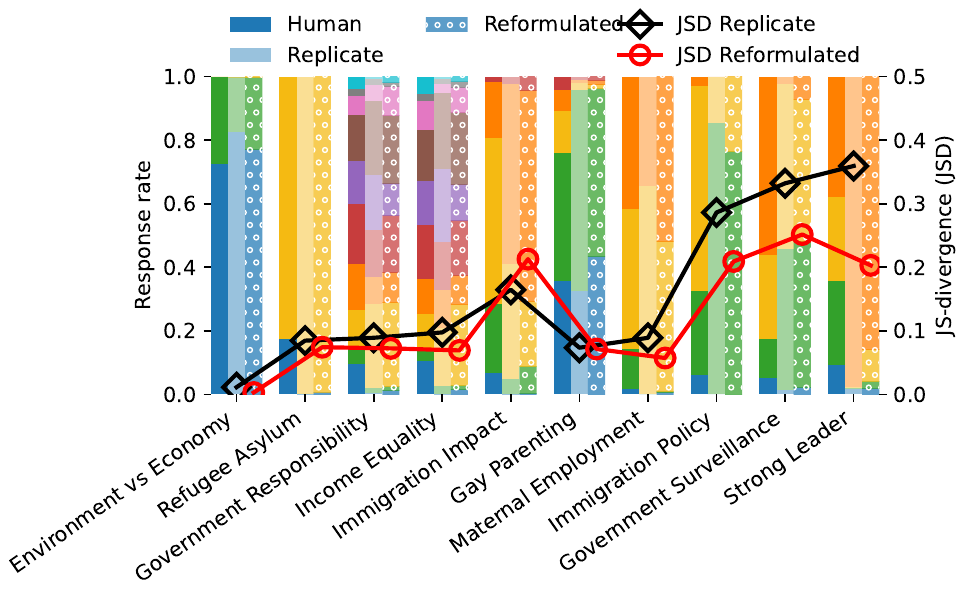}\vspace{-10pt} 
        \caption{Netherlands
        }
    \end{subfigure}
    \begin{subfigure}[]{0.48\textwidth} 
        \centering
        \includegraphics[width=\linewidth]{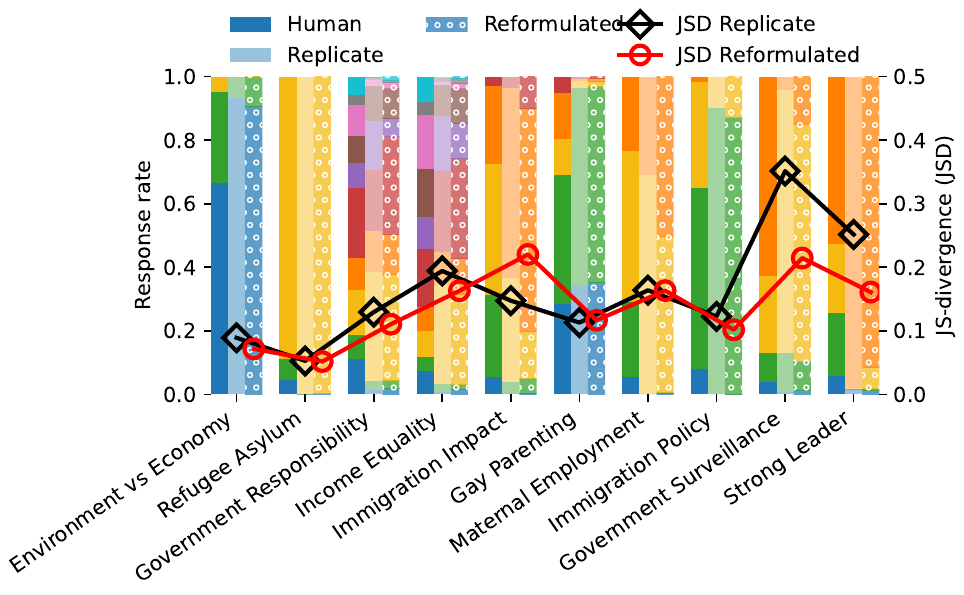}\vspace{-10pt} 
        \caption{Germany
        }
    \end{subfigure}
    \begin{subfigure}[]{0.48\textwidth}
        \centering
        \includegraphics[width=\linewidth]{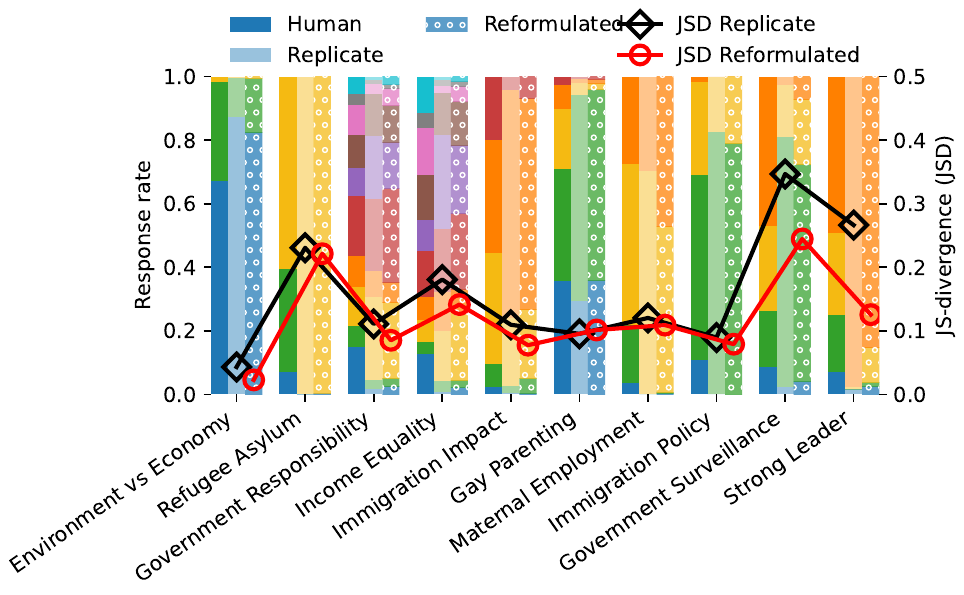}\vspace{-10pt} 
        \caption{Great Britain
        }
    \end{subfigure} 
\caption{Human, \textbf{Replicate}, and \textbf{Reformulated} response distributions for the ten WVS Wave~7 items, using GPT-4.1-mini, for the Netherlands, Germany, and Great Britain.} \label{fig:wvs-jsd-country}
\end{figure}

Figure~\ref{fig:wvs-jsd-country} shows the human, Replicate, and Reformulated response distributions for the ten WVS Wave~7 items, with the per-item JS-divergence under each condition, for the Netherlands, Germany, and Great Britain, complementing the analysis in the main text (Sec. \ref{sec:generalization-analysis}).

\clearpage
\subsection{Conditioning by demographic attribute}
\label{app:conditioning}

Section~\ref{sec:conditioning} shows that the demographic profile as a whole shapes the response distribution. Here we test which individual attributes carry that signal. We treat each of the eight ANES 2020 demographic variables separately, grouping Age into decades so it can be treated as categorical. For variable $A_k$, let $P_{k,a}$ and $Q_{k,a}$ denote the human and silicon response distributions when $A_k = a$, for instance Race with $a$ = White. If the model conditions on $A_k$, a silicon subgroup should resemble its own human counterpart more closely than it resembles the other human subgroups. We therefore define the conditioning gain
\begin{equation}
\label{eq:gain}
G_k = \frac{1}{|A_k|} \sum_{a} \left( \frac{1}{|A_k| - 1} \sum_{a' \neq a} \mathrm{JSD}\big(Q_{k,a} \,\|\, P_{k,a'}\big) - \mathrm{JSD}\big(Q_{k,a} \,\|\, P_{k,a}\big) \right),
\end{equation}
averaged over the ten items, which is positive when the model differentiates subgroups in a way that tracks the human data and zero when the silicon distribution is unrelated to the level of $A_k$. Alongside $G_k$ we report the divergence between the human subgroups themselves,

\begin{equation}
\label{eq:human-sep}
B_k = \frac{2}{|A_k| (|A_k| - 1)} \sum_{a < a'} \mathrm{JSD}\big(P_{k,a} \,\|\, P_{k,a'}\big),
\end{equation}

which bounds what conditioning could recover. Where the human subgroups barely differ, there is little for the model to reproduce. Inference uses a non-parametric bootstrap that resamples responses within each subgroup, keeping the number of respondents per subgroup at its observed value, with the false discovery rate controlled across variables by the Benjamini-Hochberg procedure as in Section~\ref{sec:evaluation}. Each subgroup contains several hundred respondents, so even a very small gain can be statistically significant. We run this experiment under the Replicate condition, since the mitigation conditions modify the prompt but not the profile.

\autoref{tab:conditioning} reports the gain for each variable. Only Ideology and Party produce substantial signal, with $G_k = 0.085$ and $0.062$, and every other gain is an order of magnitude smaller ($G_k \leq 0.008$). For Race and Discuss Politics the gain is indistinguishable from zero, and for Political Interest, Church Attendance, Gender, and Age it is small but non-zero. A large $B_k$ does not by itself produce a large gain: Race and Political Interest have the same $B_k$ of $0.013$, yet gains of $0.000$ and $0.008$. Gender and Age show negligible gains for different reasons. Gender has almost nothing to recover ($B_k = 0.003$), while Age has more human variation ($B_k = 0.007$) and the model recovers almost none of it.
A null gain shows only that the model fails to track the human subgroup differences. Two cases remain indistinguishable: the model may return the same distribution at every attribute level, or different distributions that do not match the human ones.
\begin{table}[!h]
\centering
\caption{Demographic conditioning for GPT-4.1-mini under the Replicate condition, averaged over the ten ANES 2020 items. $B_k$ is the divergence between the human subgroups and bounds what conditioning could recover; $G_k$ is the conditioning gain, with a percentile bootstrap interval. Values marked \textsuperscript{ns} are not significant after Benjamini-Hochberg correction.
Variables are ordered by $G_k$, and the rule separates the two with substantial conditioning signal from the remainder.}
\label{tab:conditioning}
\begin{tabular}{lccc}
\toprule
Demographic variable & $|A_k|$ & $B_k$ & $G_k$ [95\% CI] \\
\midrule
Ideology & 7 & 0.132 & 0.085 [0.082, 0.088] \\
Party & 7 & 0.078 & 0.062 [0.060, 0.063] \\
\midrule
Political Interest & 4 & 0.013 & 0.008 [0.007, 0.009] \\
Church Attendance & 2 & 0.018 & 0.003 [0.002, 0.004] \\
Gender & 2 & 0.003 & 0.001 [0.001, 0.002] \\
Age & 6 & 0.007 & 0.001 [0.000, 0.001] \\
Race & 5 & 0.013 & 0.000\textsuperscript{ns} [$-0.002$, 0.001] \\
Discuss Politics & 2 & 0.016 & 0.000\textsuperscript{ns} [$-0.002$, 0.001] \\
\bottomrule
\end{tabular}
\end{table}

\subsection{Ordinal weighting sensitivity}\label{app:weight-sensitivity}
The signed desirability gap weights each option by its position toward the desirable pole (Section~\ref{sec:evaluation}). For categorical items this weighting is already binary, one on the desirable option and zero otherwise, so distinguishing neutral from undesirable options would require three-way labels we do not collect. For ordinal items the scale orders the options, so the weighting can be varied directly.
We test how much the classification depends on this choice, using the WVS Wave 7 ordinal items.

For each WVS ordinal item we recompute the signed gap and the change in its absolute value under a most-desirable-only weighting that places all weight on the single most desirable option, collapsing the neutral options into the undesirable side and mirroring the weighting used for categorical items. 
Table~\ref{tab:ordinal-sensitivity} reports the desirability gap under this weighting alongside the scale-position weighting used in the main results. The main-text pattern (whether they satisfy the joint criterion) is largely unchanged, including \textit{Strong Leader}, which continues to satisfy the criterion in all three countries.
The exception is \textit{Immigration Policy}: its baseline signed gap turns negative once the weight is concentrated on the single most desirable option, so it no longer satisfies the joint criterion in the Netherlands and Germany. 

\begin{table}[h]
\centering
\caption{Ordinal weighting sensitivity for the six WVS Wave 7 ordinal items, GPT-4.1-mini. $D$ is the signed baseline gap (Reformulated $-$ Replicate) under the scale-position weighting used in the main results and under a most-desirable-only weighting. Bold marks items satisfying the joint criterion under the corresponding weighting. $\Delta$JSD does not depend on the weighting and is unchanged from \autoref{tab:reformulated-wvs7}, so only $D$ is shown here. Both weightings are tested by the paired bootstrap with Benjamini-Hochberg correction of \autoref{tab:reformulated-wvs7}; \textsuperscript{ns} marks a change that is not significant.}
\label{tab:ordinal-sensitivity}
\begin{tabular}{l cc cc}
\toprule
 & \multicolumn{2}{c}{Scale position} & \multicolumn{2}{c}{Most desirable only} \\
\cmidrule(lr){2-3}\cmidrule(lr){4-5}
Item & $D$ & $\Delta|D|$ & $D$ & $\Delta|D|$ \\
\midrule
\multicolumn{5}{l}{\textit{Netherlands}} \\
Immigration Impact & $+0.177$ & $+0.023$ & $+0.005$ & $+0.021$ \\
Gay Parenting & $+0.071$ & $+0.027$ & $-0.030$ & $+0.049$\textsuperscript{ns} \\
Maternal Employment & $+0.029$ & $+0.054$ & $-0.071$ & $+0.034$\textsuperscript{ns} \\
Immigration Policy & $\mathbf{+0.165}$ & $\mathbf{-0.030}$ & $-0.063$ & $+0.000$\textsuperscript{ns} \\
Government Surveillance & $-0.262$ & $-0.006$\textsuperscript{ns} & $-0.539$ & $-0.049$ \\
Strong Leader & $\mathbf{+0.336}$ & $\mathbf{-0.046}$ & $\mathbf{+0.597}$ & $\mathbf{-0.117}$ \\
\midrule
\multicolumn{5}{l}{\textit{Germany}} \\
Immigration Impact & $+0.172$ & $+0.054$ & $+0.006$ & $+0.065$ \\
Gay Parenting & $+0.141$ & $+0.003$\textsuperscript{ns} & $+0.058$ & $+0.010$\textsuperscript{ns} \\
Maternal Employment & $+0.140$ & $+0.061$ & $+0.073$ & $+0.196$ \\
Immigration Policy & $\mathbf{+0.062}$ & $\mathbf{-0.010}$ & $-0.080$ & $-0.001$\textsuperscript{ns} \\
Government Surveillance & $-0.184$ & $-0.044$ & $-0.583$ & $-0.108$ \\
Strong Leader & $\mathbf{+0.245}$ & $\mathbf{-0.022}$ & $\mathbf{+0.453}$ & $\mathbf{-0.066}$ \\
\midrule
\multicolumn{5}{l}{\textit{Great Britain}} \\
Immigration Impact & $+0.034$ & $-0.006$\textsuperscript{ns} & $-0.154$ & $-0.025$ \\
Gay Parenting & $+0.066$ & $+0.021$ & $-0.065$ & $-0.064$ \\
Maternal Employment & $+0.092$ & $+0.055$ & $+0.021$ & $+0.175$ \\
Immigration Policy & $+0.014$ & $-0.013$\textsuperscript{ns} & $-0.109$ & $+0.000$\textsuperscript{ns} \\
Government Surveillance & $-0.309$ & $-0.037$ & $-0.443$ & $-0.044$ \\
Strong Leader & $\mathbf{+0.257}$ & $\mathbf{-0.051}$ & $\mathbf{+0.483}$ & $\mathbf{-0.124}$ \\
\bottomrule
\end{tabular}
\end{table}

\subsection{Decoding robustness}
\label{app:decoding}
Our main experiments use stochastic decoding, but we also evaluate the methods under deterministic (greedy) decoding, which provides greater reproducibility.
Table~\ref{tab:decoding-stochastic} compares the two decodings for GPT-4.1-mini on ANES~2020, reporting the change in mean JS-divergence and mean absolute desirability gap relative to Replicate under each setting.
The two decodings yield nearly identical results. Baseline alignment is slightly worse under deterministic decoding (higher $\overline{\mathrm{JSD}}$ and $\overline{|D|}$), because always selecting the single most likely answer discards the response variation that better matches the spread of human responses. The effect of each prompt condition is nonetheless stable across settings. Reformulated improves alignment on both metrics under both decodings, while Reverse-coded, Priming, and Preamble
do not. The reformulation effect therefore appears to be a property of the prompt design rather than a consequence of the decoding strategy.
\begin{table}[!h]\centering
\caption{Effect of each prompt condition relative to Replicate for GPT-4.1-mini under stochastic (T=1) and deterministic (T=0) decoding
(ANES~2020). Values report changes in the mean JS-divergence
$\overline{\Delta \mathrm{JSD}}$ and mean absolute desirability gap
$\overline{\Delta |D|}$ across items (condition $-$ baseline); negative denotes improvement. 
Reformulated, Priming, and Preamble are averaged over the nine directional items; Reverse-coded over the six eligible items against its own six-item baseline.}
\label{tab:decoding-stochastic}
\resizebox{\textwidth}{!}{
\begin{tabular}{ll cc cc cc cc}
\toprule
& & \multicolumn{2}{c}{Reformulated} & \multicolumn{2}{c}{Reverse-coded}
  & \multicolumn{2}{c}{Priming} & \multicolumn{2}{c}{Preamble} \\
\cmidrule(lr){3-4}\cmidrule(lr){5-6}\cmidrule(lr){7-8}\cmidrule(lr){9-10}
Temperature & Base $\overline{\mathrm{JSD}}$\,/\,$\overline{|D|}$ & $\overline{\Delta \mathrm{JSD}}$ & $\overline{\Delta |D|}$ & $\overline{\Delta \mathrm{JSD}}$ & $\overline{\Delta |D|}$
      & $\overline{\Delta \mathrm{JSD}}$ & $\overline{\Delta |D|}$ & $\overline{\Delta \mathrm{JSD}}$ & $\overline{\Delta |D|}$ \\
\midrule
$T=1$  & 0.078\,/\,0.198 & $-0.019$ & $-0.050$ & $+0.146$ & $+0.114$ & $+0.042$ & $+0.042$ & $+0.043$ & $+0.065$\\
$T=0$  & 0.089\,/\,0.208 & $-0.021$ & $-0.051$ & $+0.144$ & $+0.119$ & $+0.039$ & $+0.043$ & $+0.051$ & $+0.067$ \\
\bottomrule
\end{tabular}}
\end{table}

\subsection{Demographic profile duplication}
\label{sec:duplication}
In silicon sampling, each synthetic respondent is assigned a demographic profile drawn from the target population distribution. Because profiles are sampled, the same combination of demographic attributes may be drawn more than once, producing duplicate profiles. This duplication is a property of the sampling process itself, independent of the decoding strategy. It becomes a concern under greedy (deterministic) decoding, where duplicate profiles are queried with identical prompts and therefore yield identical responses, effectively sampling the same individual multiple times. To assess how often this occurs, we measured the rate of duplicated demographic profiles produced by our sampling procedure.

We define two respondents as duplicates if they are identical across all demographic variables used to construct the profile (Appendix \ref{sec:demographic-vars})
We sampled 5,441 synthetic respondents (matching the size of the ANES 2020 sample), using GPT-4.1-mini, and computed the proportion of respondents whose profile coincided with that of at least one other respondent. Table~\ref{tab:duplication} reports these rates across the ten survey questions under both the Replicate and Reformulated conditions.

\begin{table}[h]
\centering
\caption{Percentage of duplicated demographic profiles across survey items, for 5{,}441 sampled synthetic respondents under the Replicate and Reformulated conditions.}
\label{tab:duplication}
\resizebox{\textwidth}{!}{%
\begin{tabular}{l*{11}{c}}
\toprule
 & Current & Gay & Refugee & Income & Gender & Climate & Gun & Drug & Race & Health & \textbf{Average} \\
 & Economy & Marriage & Allowing & Inequality & Role & Change & Regulation & Addiction & Diversity & Insurance & \\
\midrule
Replicate    & 2.55 & 2.30 & 2.57 & 2.57 & 2.33 & 2.26 & 2.41 & 2.22 & 2.33 & 2.72 & \textbf{2.43} \\
Reformulated & 2.54 & 2.78 & 2.46 & 2.44 & 3.18 & 2.61 & 2.83 & 2.87 & 2.68 & 2.50 & \textbf{2.69} \\
\bottomrule
\end{tabular}%
}
\end{table}

Duplication rates are low and stable across questions, averaging 2.43\% under Replicate and 2.69\% under Reformulated, corresponding to roughly 132 and 146 of 5,441 respondents, respectively. Comparable rates were observed in the remaining experimental conditions. 
The vast majority of sampled profiles are therefore unique, and the small number of duplicates is unlikely to meaningfully affect our results, even under greedy decoding. Our main results rely on stochastic decoding, where duplicate profiles need not yield identical responses, which addresses this concern directly.
We further note that some degree of duplication is expected and not inherently problematic: identical profiles arise when a demographic combination is genuinely common in the target population, so their repetition reflects realistic population frequencies rather than a sampling artifact.

\subsection{Non-response rates and their effects}\label{app:non-response-analysis}
Here, we examine whether the non-response handling in Appendix~\ref{sec:non-response} affects our results. 
Of the $5{,}441$ ANES 2020 respondents, $83.8\%$ answer all ten items substantively, and non-response is concentrated in the remaining $16.2\%$ rather than spread across the sample.
Because non-response is concentrated in this subset rather than uniform across respondents,
listwise deletion would remove a non-random group and could bias the composition of the
retained sample, which available-case handling avoids.

As shown in~\autoref{tab:nr-rates}, item non-response is moderate and similar across the eight policy items, between $12.8\%$ and $13.7\%$, and near zero on \textit{Gay Marriage} ($1.3\%$) and \textit{Current Economy} ($0.3\%$), items on which respondents rarely decline to answer. 
These rates tell us how much is missing from each item, but not whether it is missing evenly across groups. 
Excluding non-response could distort demographic comparisons if some groups skipped questions more often than others. To check this, for each item we classify every respondent as substantive or non-substantive and test with a Chi-Square ($\chi^2$) test whether the non-response rate differs across the levels of each conditioning variable. Because the sample is large, significance is easy to reach, so alongside the count of items with $p<0.05$ we report
Cram\'er's $V$, which measures how much the non-response rate differs across
groups, from $0$ (no difference) to $1$ (a strong difference). Table~\ref{tab:nr-diff} shows these differences are small. Non-response is essentially uniform across Gender, and statistically detectable across Party and Race on nine and six of the ten items, but weak throughout. The largest $V$ across all $70$ variable-by-item tests
is $0.078$, below $0.1$, the conventional threshold for a small effect.
Several of the remaining significant results come from sparse tables and are discounted, as the $\chi^2$ test is unreliable when expected counts are very low. 
The excluded non-response is thus only mildly patterned across groups, and the contrasts are preserved up to small shifts on the party and racial dimensions.

As a further check, we recompute the JS-divergence with human non-response reintroduced as one additional category on both distributions. The silicon distribution then contains zero non-response, whereas the human distribution retains its non-responses. Given this asymmetric design, we compare only how the two treatments rank the group-item pairs by divergence.
That ordering is stable: the rank correlation between the substantive-only and augmented divergences across the 290 group-item pairs is $0.90$ (Spearman), so the groups and items the model reproduces relatively well or poorly are the same whether or not non-response is included. The substantive-only results therefore do not depend on the exclusion of non-response.
\begin{table}[t]
\centering
\caption{Item non-response in ANES 2020, as a percentage of the $5{,}441$ respondents. }
\label{tab:nr-rates}
\begin{tabular}{lr}
\toprule
Item & Non-response (\%) \\
\midrule
Drug Addiction     & 13.7 \\
Health Insurance   & 13.7 \\
Race Diversity     & 13.4 \\
Gender Role        & 13.3 \\
Gun Regulation     & 13.2 \\
Climate Change     & 13.1 \\
Income Inequality  & 13.0 \\
Refugee Allowing   & 12.8 \\
\midrule
Gay Marriage       & 1.3 \\
Current Economy    & 0.3 \\
\bottomrule
\end{tabular}
\end{table}

\begin{table}[t]
\centering
\caption{Differential item non-response across the levels of each conditioning variable (ANES 2020). ``Items differential'' counts the items (of ten) with $p<0.05$; Max $V$ is the largest Cram\'er's $V$ over the ten items. $^{\dagger}$For these variables the differential items come mostly from sparse tables (minimum expected count below five) and are not treated as reliable.}
\label{tab:nr-diff}
\begin{tabular}{lrr}
\toprule
Conditioning variable & Items differential & Max $V$ \\
\midrule
Party                          & 9/10 & 0.078 \\
Age                            &7/10  & 0.054 \\
Race                           & 6/10 & 0.051 \\
Discuss Politics$^{\dagger}$   & 3/10 & 0.049 \\
Political Interest$^{\dagger}$ & 2/10 & 0.042 \\
Church                         & 1/10 & 0.044 \\
Ideology                       & 1/10 & 0.055 \\
Gender                         & 0/10 & 0.016 \\
\bottomrule
\end{tabular}
\end{table}

\subsection{Full stratified analysis by demographic subgroup}\label{sec:full-stratified-results}
We present the full demographic subgroup analysis for GPT-4.1-mini, reporting the change in JS-divergence ($\Delta\mathrm{JSD}$) and in the absolute desirability gap ($\Delta|D|$) for each prompting condition relative to Replicate. For every condition, the two metrics are shown as a pair of tables sharing the same column order, so a given subgroup and item can be read across both. \textit{Current Economy} is excluded throughout, as it has no desirability direction and therefore no gap to report. The Reverse-coded condition was collected only for its six-item subset, so \textit{Drug Addiction}, \textit{Gun Regulation}, and \textit{Health Insurance} appear as dashes in its tables and its row averages span six items.

The signed baseline gap is not uniformly positive at subgroup granularity: 43 of the 315 subgroup-item cells have $D<0$ under Replicate, comprising 34 of the 35 subgroups on \textit{Gender Role}, 7 on \textit{Gun Regulation}, and 2 on \textit{Climate Change}. 
$\Delta|D|<0$ denotes a desirability gap narrowing toward zero regardless of its baseline sign.
Where $D<0$ at baseline, a narrowing gap means the model selects the socially desirable option more often than it did under Replicate.


\begin{table}[]
\centering
\caption{Difference in JS-divergence ($\Delta$JSD) between the \textbf{Reformulated} and Replicate conditions on stratified subgroups using \textbf{GPT-4.1-mini} (ANES~2020). Darker blue signifies greater improvement ($\Delta$JSD$<0$), darker red greater worsening ($\Delta$JSD$>0$). }
\label{tab:full-reformulated-deltajsd}
\resizebox{\linewidth}{!}{
\begin{tabular}{llcccccccccc}
\toprule
 &  & \makecell{Gay Marriage} & \makecell{Refugee\\ Allowing} & \makecell{Income\\ Inequality} & \makecell{Gender Role} & \makecell{Climate\\ Change} & \makecell{Gun\\ Regulation} & \makecell{Drug\\ Addiction} & \makecell{Race\\ Diversity} & \makecell{Health\\ Insurance} & \makecell{Average} \\
\midrule
\multirow{7}{*}{Ideology} & Extremely liberal & \colorcell{-0.016} & \colorcell{-0.011} & \colorcell{0.037} & \colorcell{0.251} & \colorcell{-0.005} & \colorcell{0.02} & \colorcell{0} & \colorcell{-0.021} & \colorcell{0.014} & \colorcell{0.03} \\
 & Liberal & \colorcell{0} & \colorcell{-0.029} & \colorcell{0.049} & \colorcell{0} & \colorcell{0.026} & \colorcell{0.027} & \colorcell{-0.005} & \colorcell{-0.016} & \colorcell{0.014} & \colorcell{0.007} \\
 & Slightly liberal & \colorcell{-0.005} & \colorcell{-0.01} & \colorcell{0.064} & \colorcell{-0.062} & \colorcell{0.06} & \colorcell{0.028} & \colorcell{-0.015} & \colorcell{-0.046} & \colorcell{0.013} & \colorcell{0.003} \\
 & Moderate & \colorcell{-0.022} & \colorcell{-0.096} & \colorcell{0.027} & \colorcell{-0.098} & \colorcell{-0.013} & \colorcell{0.011} & \colorcell{-0.015} & \colorcell{-0.064} & \colorcell{0.032} & \colorcell{-0.026} \\
 & Slightly conservative & \colorcell{-0.012} & \colorcell{-0.103} & \colorcell{0.001} & \colorcell{-0.082} & \colorcell{-0.109} & \colorcell{0.005} & \colorcell{-0.04} & \colorcell{-0.125} & \colorcell{0.058} & \colorcell{-0.045} \\
 & Conservative & \colorcell{0.007} & \colorcell{-0.063} & \colorcell{0.051} & \colorcell{-0.07} & \colorcell{-0.099} & \colorcell{0.027} & \colorcell{-0.04} & \colorcell{-0.154} & \colorcell{0.016} & \colorcell{-0.036} \\
 & Extremely conservative & \colorcell{0.024} & \colorcell{-0.01} & \colorcell{0.057} & \colorcell{-0.032} & \colorcell{-0.139} & \colorcell{0.139} & \colorcell{0.022} & \colorcell{0.004} & \colorcell{0.093} & \colorcell{0.018} \\
\bottomrule
\end{tabular}}
\end{table}

\begin{table}[t]
\centering
\caption{Difference in absolute desirability gap ($\Delta|D|$) between the \textbf{Reformulated} and Replicate conditions on stratified subgroups using \textbf{GPT-4.1-mini} (ANES~2020). Darker blue signifies greater improvement ($\Delta|D|$$<0$), darker red greater worsening ($\Delta|D|$$>0$). }
\label{tab:full-reformulated-deltaabsD}
\resizebox{\linewidth}{!}{\begin{tabular}{llcccccccccc}
\toprule
 &  & \makecell{Gay Marriage} & \makecell{Refugee\\ Allowing} & \makecell{Income\\ Inequality} & \makecell{Gender Role} & \makecell{Climate\\ Change} & \makecell{Gun\\ Regulation} & \makecell{Drug\\ Addiction} & \makecell{Race\\ Diversity} & \makecell{Health\\ Insurance} & \makecell{Average} \\
\midrule
\multirow{7}{*}{Ideology} & Extremely liberal & \colorcell{-0.011} & \colorcell{-0.074} & \colorcell{-0.031} & \colorcell{0.601} & \colorcell{0.001} & \colorcell{0.005} & \colorcell{0} & \colorcell{-0.065} & \colorcell{0.022} & \colorcell{0.05} \\
 & Liberal & \colorcell{0} & \colorcell{-0.077} & \colorcell{-0.049} & \colorcell{0.183} & \colorcell{0.042} & \colorcell{0.023} & \colorcell{-0.001} & \colorcell{-0.019} & \colorcell{0.023} & \colorcell{0.014} \\
 & Slightly liberal & \colorcell{-0.026} & \colorcell{-0.104} & \colorcell{-0.068} & \colorcell{-0.023} & \colorcell{0.144} & \colorcell{0.086} & \colorcell{-0.02} & \colorcell{-0.059} & \colorcell{0.034} & \colorcell{-0.004} \\
 & Moderate & \colorcell{-0.056} & \colorcell{-0.182} & \colorcell{-0.013} & \colorcell{-0.008} & \colorcell{0.046} & \colorcell{0.065} & \colorcell{-0.011} & \colorcell{-0.034} & \colorcell{0.053} & \colorcell{-0.016} \\
 & Slightly conservative & \colorcell{-0.045} & \colorcell{-0.171} & \colorcell{-0.152} & \colorcell{-0.009} & \colorcell{-0.181} & \colorcell{-0.089} & \colorcell{-0.058} & \colorcell{-0.24} & \colorcell{0.081} & \colorcell{-0.096} \\
 & Conservative & \colorcell{0.049} & \colorcell{-0.178} & \colorcell{-0.014} & \colorcell{-0.009} & \colorcell{-0.2} & \colorcell{0.028} & \colorcell{-0.166} & \colorcell{-0.393} & \colorcell{-0.015} & \colorcell{-0.1} \\
 & Extremely conservative & \colorcell{0.11} & \colorcell{-0.064} & \colorcell{0.047} & \colorcell{-0.027} & \colorcell{-0.155} & \colorcell{0.092} & \colorcell{-0.265} & \colorcell{-0.392} & \colorcell{0.11} & \colorcell{-0.06} \\
\bottomrule
\end{tabular}}
\end{table}

\begin{table}[t]
\centering
\caption{Difference in JS-divergence ($\Delta$JSD) between the \textbf{Reverse-coded} and Replicate conditions on stratified subgroups using \textbf{GPT-4.1-mini} (ANES~2020). Darker blue signifies greater improvement ($\Delta$JSD$<0$), darker red greater worsening ($\Delta$JSD$>0$). Dashes mark items not collected under this condition.}
\label{tab:full-reversed-deltajsd}
\resizebox{\linewidth}{!}{\begin{tabular}{llcccccccccc}
\toprule
 &  & \makecell{Gay Marriage} & \makecell{Refugee\\ Allowing} & \makecell{Income\\ Inequality} & \makecell{Gender Role} & \makecell{Climate\\ Change} & \makecell{Gun\\ Regulation} & \makecell{Drug\\ Addiction} & \makecell{Race\\ Diversity} & \makecell{Health\\ Insurance} & \makecell{Average} \\
\midrule
\multirow{7}{*}{Ideology} & Extremely liberal & \colorcell{0} & \colorcell{-0.018} & \colorcell{0.153} & \colorcell{0.715} & \colorcell{0.087} & -- & -- & \colorcell{-0.019} & -- & \colorcell{0.153} \\
 & Liberal & \colorcell{-0.002} & \colorcell{0.002} & \colorcell{0.128} & \colorcell{0.688} & \colorcell{0.162} & -- & -- & \colorcell{-0.021} & -- & \colorcell{0.16} \\
 & Slightly liberal & \colorcell{0.02} & \colorcell{-0.003} & \colorcell{0.062} & \colorcell{0.734} & \colorcell{0.209} & -- & -- & \colorcell{-0.072} & -- & \colorcell{0.158} \\
 & Moderate & \colorcell{0.036} & \colorcell{-0.096} & \colorcell{-0.044} & \colorcell{0.743} & \colorcell{0.336} & -- & -- & \colorcell{-0.171} & -- & \colorcell{0.134} \\
 & Slightly conservative & \colorcell{0.1} & \colorcell{-0.135} & \colorcell{-0.04} & \colorcell{0.788} & \colorcell{0.427} & -- & -- & \colorcell{-0.185} & -- & \colorcell{0.159} \\
 & Conservative & \colorcell{0.159} & \colorcell{-0.026} & \colorcell{0.169} & \colorcell{0.854} & \colorcell{0.406} & -- & -- & \colorcell{-0.278} & -- & \colorcell{0.214} \\
 & Extremely conservative & \colorcell{0.073} & \colorcell{-0.02} & \colorcell{0.315} & \colorcell{0.836} & \colorcell{0.182} & -- & -- & \colorcell{-0.23} & -- & \colorcell{0.193} \\
\bottomrule
\end{tabular}}
\end{table}

\begin{table}[t]
\centering
\caption{Difference in absolute desirability gap ($\Delta|D|$) between the \textbf{Reverse-coded} and Replicate conditions on stratified subgroups using \textbf{GPT-4.1-mini} (ANES~2020). Darker blue signifies greater improvement ($\Delta|D|$$<0$), darker red greater worsening ($\Delta|D|$$>0$). Dashes mark items not collected under this condition.}
\label{tab:full-reversed-deltaabsD}
\resizebox{\linewidth}{!}{\begin{tabular}{llcccccccccc}
\toprule
 &  & \makecell{Gay Marriage} & \makecell{Refugee\\ Allowing} & \makecell{Income\\ Inequality} & \makecell{Gender Role} & \makecell{Climate\\ Change} & \makecell{Gun\\ Regulation} & \makecell{Drug\\ Addiction} & \makecell{Race\\ Diversity} & \makecell{Health\\ Insurance} & \makecell{Average} \\
\midrule
\multirow{7}{*}{Ideology} & Extremely liberal & \colorcell{0} & \colorcell{-0.027} & \colorcell{0.208} & \colorcell{0.903} & \colorcell{0.038} & -- & -- & \colorcell{-0.008} & -- & \colorcell{0.186} \\
 & Liberal & \colorcell{-0.003} & \colorcell{-0.039} & \colorcell{0.175} & \colorcell{0.83} & \colorcell{0.087} & -- & -- & \colorcell{-0.013} & -- & \colorcell{0.173} \\
 & Slightly liberal & \colorcell{-0.028} & \colorcell{-0.081} & \colorcell{0.048} & \colorcell{0.862} & \colorcell{0.124} & -- & -- & \colorcell{-0.05} & -- & \colorcell{0.146} \\
 & Moderate & \colorcell{-0.029} & \colorcell{-0.155} & \colorcell{-0.122} & \colorcell{0.912} & \colorcell{0.243} & -- & -- & \colorcell{-0.145} & -- & \colorcell{0.117} \\
 & Slightly conservative & \colorcell{0.073} & \colorcell{-0.212} & \colorcell{-0.085} & \colorcell{0.921} & \colorcell{0.284} & -- & -- & \colorcell{-0.128} & -- & \colorcell{0.142} \\
 & Conservative & \colorcell{-0.059} & \colorcell{-0.041} & \colorcell{0.305} & \colorcell{0.958} & \colorcell{0.319} & -- & -- & \colorcell{-0.276} & -- & \colorcell{0.201} \\
 & Extremely conservative & \colorcell{-0.21} & \colorcell{-0.049} & \colorcell{0.486} & \colorcell{0.914} & \colorcell{0.173} & -- & -- & \colorcell{-0.404} & -- & \colorcell{0.152} \\
\bottomrule
\end{tabular}}
\end{table}

\begin{table}[t]
\centering
\caption{Difference in JS-divergence ($\Delta$JSD) between the \textbf{Priming} and Replicate conditions on stratified subgroups using \textbf{GPT-4.1-mini} (ANES~2020). Darker blue signifies greater improvement ($\Delta$JSD$<0$), darker red greater worsening ($\Delta$JSD$>0$). }
\label{tab:full-priming-deltajsd}
\resizebox{\linewidth}{!}{\begin{tabular}{llcccccccccc}
\toprule
 &  & \makecell{Gay Marriage} & \makecell{Refugee\\ Allowing} & \makecell{Income\\ Inequality} & \makecell{Gender Role} & \makecell{Climate\\ Change} & \makecell{Gun\\ Regulation} & \makecell{Drug\\ Addiction} & \makecell{Race\\ Diversity} & \makecell{Health\\ Insurance} & \makecell{Average} \\
\midrule
\multirow{7}{*}{Ideology} & Extremely liberal & \colorcell{0} & \colorcell{0.008} & \colorcell{0.047} & \colorcell{0.051} & \colorcell{0.066} & \colorcell{0.02} & \colorcell{0} & \colorcell{0} & \colorcell{0.034} & \colorcell{0.025} \\
 & Liberal & \colorcell{0.004} & \colorcell{0.017} & \colorcell{0.056} & \colorcell{0.018} & \colorcell{0.065} & \colorcell{-0.011} & \colorcell{0} & \colorcell{0.008} & \colorcell{0.029} & \colorcell{0.021} \\
 & Slightly liberal & \colorcell{-0.013} & \colorcell{0.032} & \colorcell{0.082} & \colorcell{0.051} & \colorcell{0.043} & \colorcell{0.007} & \colorcell{0} & \colorcell{0} & \colorcell{0.056} & \colorcell{0.029} \\
 & Moderate & \colorcell{-0.015} & \colorcell{0.032} & \colorcell{0.061} & \colorcell{0.065} & \colorcell{0.122} & \colorcell{0.009} & \colorcell{0} & \colorcell{0.016} & \colorcell{0.054} & \colorcell{0.038} \\
 & Slightly conservative & \colorcell{0.013} & \colorcell{0.05} & \colorcell{0.103} & \colorcell{0.093} & \colorcell{0.272} & \colorcell{0.021} & \colorcell{0} & \colorcell{0.008} & \colorcell{0.087} & \colorcell{0.072} \\
 & Conservative & \colorcell{-0.017} & \colorcell{0.057} & \colorcell{0.078} & \colorcell{0.204} & \colorcell{0.355} & \colorcell{0.007} & \colorcell{0.003} & \colorcell{0.063} & \colorcell{0.028} & \colorcell{0.087} \\
 & Extremely conservative & \colorcell{-0.075} & \colorcell{-0.022} & \colorcell{0.026} & \colorcell{0.26} & \colorcell{0.395} & \colorcell{-0.021} & \colorcell{0.069} & \colorcell{0.226} & \colorcell{0.051} & \colorcell{0.101} \\
\bottomrule
\end{tabular}}
\end{table}

\begin{table}[t]
\centering
\caption{Difference in absolute desirability gap ($\Delta|D|$) between the \textbf{Priming} and Replicate conditions on stratified subgroups using \textbf{GPT-4.1-mini} (ANES~2020). Darker blue signifies greater improvement ($\Delta|D|$$<0$), darker red greater worsening ($\Delta|D|$$>0$). }
\label{tab:full-priming-deltaabsD}
\resizebox{\linewidth}{!}{\begin{tabular}{llcccccccccc}
\toprule
 &  & \makecell{Gay Marriage} & \makecell{Refugee\\ Allowing} & \makecell{Income\\ Inequality} & \makecell{Gender Role} & \makecell{Climate\\ Change} & \makecell{Gun\\ Regulation} & \makecell{Drug\\ Addiction} & \makecell{Race\\ Diversity} & \makecell{Health\\ Insurance} & \makecell{Average} \\
\midrule
\multirow{7}{*}{Ideology} & Extremely liberal & \colorcell{0} & \colorcell{0.011} & \colorcell{0.057} & \colorcell{0.082} & \colorcell{0.035} & \colorcell{0.096} & \colorcell{0} & \colorcell{0} & \colorcell{0.043} & \colorcell{0.036} \\
 & Liberal & \colorcell{0.004} & \colorcell{0.014} & \colorcell{0.089} & \colorcell{0.006} & \colorcell{0.066} & \colorcell{-0.056} & \colorcell{0} & \colorcell{0.003} & \colorcell{0.038} & \colorcell{0.018} \\
 & Slightly liberal & \colorcell{-0.018} & \colorcell{0.054} & \colorcell{0.143} & \colorcell{0.003} & \colorcell{0.051} & \colorcell{-0.029} & \colorcell{0} & \colorcell{0} & \colorcell{0.106} & \colorcell{0.035} \\
 & Moderate & \colorcell{-0.033} & \colorcell{0.024} & \colorcell{0.088} & \colorcell{0} & \colorcell{0.125} & \colorcell{0.027} & \colorcell{0} & \colorcell{0.004} & \colorcell{0.063} & \colorcell{0.033} \\
 & Slightly conservative & \colorcell{0.04} & \colorcell{0.044} & \colorcell{0.134} & \colorcell{0} & \colorcell{0.154} & \colorcell{0.036} & \colorcell{0} & \colorcell{0.002} & \colorcell{0.115} & \colorcell{0.058} \\
 & Conservative & \colorcell{0.024} & \colorcell{0.082} & \colorcell{0.091} & \colorcell{0} & \colorcell{0.251} & \colorcell{-0.008} & \colorcell{0.001} & \colorcell{0.023} & \colorcell{0.064} & \colorcell{0.059} \\
 & Extremely conservative & \colorcell{0.025} & \colorcell{-0.042} & \colorcell{0.043} & \colorcell{0} & \colorcell{0.316} & \colorcell{-0.074} & \colorcell{0.08} & \colorcell{0.128} & \colorcell{0.141} & \colorcell{0.069} \\
\bottomrule
\end{tabular}}
\end{table}

\begin{table}[t]
\centering
\caption{Difference in JS-divergence ($\Delta$JSD) between the \textbf{Preamble} and Replicate conditions on stratified subgroups using \textbf{GPT-4.1-mini} (ANES~2020). Darker blue signifies greater improvement ($\Delta$JSD$<0$), darker red greater worsening ($\Delta$JSD$>0$).}
\label{tab:full-preamble-deltajsd}
\resizebox{\linewidth}{!}{\begin{tabular}{llcccccccccc}
\toprule
 &  & \makecell{Gay Marriage} & \makecell{Refugee\\ Allowing} & \makecell{Income\\ Inequality} & \makecell{Gender Role} & \makecell{Climate\\ Change} & \makecell{Gun\\ Regulation} & \makecell{Drug\\ Addiction} & \makecell{Race\\ Diversity} & \makecell{Health\\ Insurance} & \makecell{Average} \\
\midrule
\multirow{7}{*}{Ideology} & Extremely liberal & \colorcell{0} & \colorcell{0.031} & \colorcell{0.051} & \colorcell{0.017} & \colorcell{0.046} & \colorcell{0.035} & \colorcell{0} & \colorcell{0} & \colorcell{0.043} & \colorcell{0.025} \\
 & Liberal & \colorcell{-0.001} & \colorcell{0.037} & \colorcell{0.042} & \colorcell{0.014} & \colorcell{0.031} & \colorcell{-0.006} & \colorcell{0} & \colorcell{0.003} & \colorcell{0.054} & \colorcell{0.019} \\
 & Slightly liberal & \colorcell{0.004} & \colorcell{0.097} & \colorcell{0.043} & \colorcell{0.051} & \colorcell{-0.002} & \colorcell{-0.003} & \colorcell{0} & \colorcell{-0.01} & \colorcell{0.093} & \colorcell{0.03} \\
 & Moderate & \colorcell{0.021} & \colorcell{0.156} & \colorcell{0.072} & \colorcell{0.044} & \colorcell{0.016} & \colorcell{0.005} & \colorcell{0} & \colorcell{-0.002} & \colorcell{0.138} & \colorcell{0.05} \\
 & Slightly conservative & \colorcell{0.036} & \colorcell{0.197} & \colorcell{0.098} & \colorcell{0.041} & \colorcell{0.081} & \colorcell{0.013} & \colorcell{0} & \colorcell{-0.012} & \colorcell{0.242} & \colorcell{0.077} \\
 & Conservative & \colorcell{0.016} & \colorcell{0.259} & \colorcell{0.096} & \colorcell{0.043} & \colorcell{0.053} & \colorcell{0.013} & \colorcell{0.003} & \colorcell{-0.04} & \colorcell{0.253} & \colorcell{0.077} \\
 & Extremely conservative & \colorcell{-0.012} & \colorcell{0.153} & \colorcell{0.054} & \colorcell{0.039} & \colorcell{0.041} & \colorcell{0.037} & \colorcell{0.069} & \colorcell{-0.055} & \colorcell{0.19} & \colorcell{0.057} \\
\bottomrule
\end{tabular}}
\end{table}

\begin{table}[t]
\centering
\caption{Difference in absolute desirability gap ($\Delta|D|$) between the \textbf{Preamble} and Replicate conditions on stratified subgroups using \textbf{GPT-4.1-mini} (ANES~2020). Darker blue signifies greater improvement ($\Delta|D|$$<0$), darker red greater worsening ($\Delta|D|$$>0$). }
\label{tab:full-preamble-deltaabsD}
\resizebox{\linewidth}{!}{\begin{tabular}{llcccccccccc}
\toprule
 &  & \makecell{Gay Marriage} & \makecell{Refugee\\ Allowing} & \makecell{Income\\ Inequality} & \makecell{Gender Role} & \makecell{Climate\\ Change} & \makecell{Gun\\ Regulation} & \makecell{Drug\\ Addiction} & \makecell{Race\\ Diversity} & \makecell{Health\\ Insurance} & \makecell{Average} \\
\midrule
\multirow{7}{*}{Ideology} & Extremely liberal & \colorcell{0} & \colorcell{0.023} & \colorcell{0.058} & \colorcell{0.056} & \colorcell{0.031} & \colorcell{0.127} & \colorcell{0} & \colorcell{0} & \colorcell{0.047} & \colorcell{0.038} \\
 & Liberal & \colorcell{0.001} & \colorcell{0.023} & \colorcell{0.076} & \colorcell{0.005} & \colorcell{0.048} & \colorcell{-0.035} & \colorcell{0} & \colorcell{0.001} & \colorcell{0.053} & \colorcell{0.019} \\
 & Slightly liberal & \colorcell{0.013} & \colorcell{0.098} & \colorcell{0.086} & \colorcell{0.003} & \colorcell{-0.001} & \colorcell{-0.045} & \colorcell{0} & \colorcell{-0.003} & \colorcell{0.135} & \colorcell{0.032} \\
 & Moderate & \colorcell{0.028} & \colorcell{0.101} & \colorcell{0.085} & \colorcell{0} & \colorcell{0.066} & \colorcell{0.053} & \colorcell{0} & \colorcell{-0.001} & \colorcell{0.151} & \colorcell{0.054} \\
 & Slightly conservative & \colorcell{0.084} & \colorcell{0.13} & \colorcell{0.098} & \colorcell{-0.002} & \colorcell{0.082} & \colorcell{0.07} & \colorcell{0} & \colorcell{-0.004} & \colorcell{0.245} & \colorcell{0.078} \\
 & Conservative & \colorcell{0.102} & \colorcell{0.304} & \colorcell{0.086} & \colorcell{0} & \colorcell{0.082} & \colorcell{0.053} & \colorcell{0.001} & \colorcell{-0.02} & \colorcell{0.342} & \colorcell{0.106} \\
 & Extremely conservative & \colorcell{0.127} & \colorcell{0.339} & \colorcell{0.106} & \colorcell{0} & \colorcell{0.063} & \colorcell{0.088} & \colorcell{0.08} & \colorcell{-0.052} & \colorcell{0.345} & \colorcell{0.122} \\
\bottomrule
\end{tabular}}
\end{table}

\clearpage

\clearpage
\section{Datasets and Prompts}\label{sec:data-prompt}

\subsection{Selected demographic variables}\label{sec:demographic-vars}
\begin{table}[h]
\centering
\caption{Selected demographic variables of ANES 2020 respondents. `Code' indicates the variable names used in the ANES 2020 dataset. }
\begin{tabular}{ll p{11cm}}
\toprule
\textbf{Variable}  & \textbf{Code} & \textbf{Answer choices}  \\ \midrule
Race               & V201549x      & 1: white, 2: black, 3: hispanic, 4: asian, 5: native american   \\[1ex]
Gender             & V201600       & 1: man, 2: woman \\[1ex]
Age                & V201507x      & \textless{}Open-ended\textgreater{}\\[1ex]
Ideology           & V201200       & 1: extremely liberal, 2: liberal, 3: slightly liberal, 4: moderate, 5: slightly conservative, 6: conservative, 7: extremely conservative\\[1ex]
Party           & V201231x      & 1: a strong democrat, 2: a weak Democrat, 3: an independent who leans Democratic, 4: an independent, 5: an independent who leans Republican, 6: a weak Republican, 7: a strong Republican \\[1ex]
Political interest & V202406       & 1: very, 2: somewhat, 3: not very, 4: not at all\\[1ex]
Church attendance  & V201452       & 1: attend church, 2: do not attend church\\[1ex]
Political discuss  & V202022       & 1: I like to discuss politics with my family and friends., 2: I never discuss politics with my family or friends.\\ 
\bottomrule
\end{tabular}
\end{table}

\begin{table}[h]
\centering
\caption{Selected demographic variables of WVS Wave 7 respondents. `Code' indicates the variable names used in the  WVS Wave 7 dataset. }\label{tab:wvs-demo}
\begin{tabular}{ll p{11cm}}
\toprule
\textbf{Variable}  & \textbf{Code} & \textbf{Answer choices}  \\ \midrule
Race & Q290 & Country-specific codes; not applicable for Germany. \newline
\textit{Netherlands:} 528001: white, 528002: black, 528003: South Asian, 528004: East Asian, 528005: Arabic or Central Asian, 528999: another ethnic group. \newline 
\textit{Great Britain:} 826100: White British, 826101: Irish, 826102: Gypsy or Irish Traveller, 826103: any other White background, 826002: Black Caribbean, 826003: Black African, 826004: any other Black background, 826005: Indian, 826006: Pakistani, 826007: Bangladeshi, 826008: Chinese, 826107: any other Asian background, 826015: Arabic or Central Asian, 826016: mixed race, 826104 \& 826105: mixed White and Black, 826106: mixed White and Asian, 826999: another ethnic background. \\[1ex]
Gender             & Q260       
& 1: man, 2: woman \\[1ex]
Age                & Q262      & \textless{}Open-ended\textgreater{}\\[1ex]
Education & Q275 & 0: early childhood education or no formal education, 1: primary education, 2: lower secondary education, 3: upper secondary education, 4: post-secondary non-tertiary education, 5: short-cycle tertiary education, 6: a bachelor's degree or equivalent, 7: a master's degree or equivalent, 8: a doctoral degree or equivalent \\[1ex]
Income & Q288R &  1: low, 2: medium, 3: high \\[1ex]
Ideology           & Q240       & 1: on the far left,
            2: on the left,
            3: left of center,
            4: slightly left of center,
            5: at the center,
            6: at the center,
            7: slightly right of center,
            8: right of center,
            9: on the right,
            10: on the far right
\\[1ex]
Political interest & Q199       & 1: very, 2: somewhat, 3: not very, 4: not at all\\
\bottomrule
\end{tabular}
\end{table}

\subsection{All questions and prompt conditions of ANES 2020}\label{sec:all-questions}

\textbf{Race diversity V202371}
\\[1ex]
\textit{Original}:
Does the increasing number of people of many different races and ethnic groups in the United States make this country a better place to live, a worse place to live, or does it make no difference?
\\[1ex]
\underline{1. Better} \ 2. Worse \ 3. Makes no difference
\\[1ex]
\textit{Reformulated:}
How would this respondent describe the impact of the increasing number of people of many different races and ethnic groups on the United States as a place to live?
\\[1ex]
1. Improves it as a place to live \ 2. Worsens it as a place to live \ 3. Makes no difference
\\[1ex]
\textit{Reverse-coded:}
Does limiting the number of people of many different races and ethnic groups in the United States make this country a better place to live, a worse place to live, or does it make no difference?
\\[1ex]
1. Better \ 2. Worse \ 3. Makes no difference
\\[1ex]
\textbf{Gender role V202287}
\\[1ex]
\textit{Original}:
Do you think it is better, worse, or makes no difference for the family as a whole if the man works outside the home and the woman takes care of the home and family?
\\[1ex]
1. Better \ \underline{2. Worse} \ 3. Makes no difference
\\[1ex]
\textit{Reformulated:}
How would this respondent assess whether it is better, worse, or makes no difference for the family as a whole if the man works outside the home and the woman takes care of the home and family?
\\[1ex]
1. Better \ 2. Worse \ 3. Makes no difference
\\[1ex]
\textit{Reverse-coded:}
Do you think it is better, worse, or makes no difference for the family as a whole if both the man and the woman share work outside the home and take care of the home and family
\\[1ex]
1. Better \ 2. Worse \ 3. Makes no difference
\\[1ex]
\textbf{Current Economy V201324}
\\[1ex]
\textit{Original}:
What do you think about the state of the economy these days in the United States?
\\[1ex]
1. Very good \ 2. Good \ 3. Neither good nor bad \ 4. Bad \ 5. Very bad
\\[1ex]
\textit{Reformulated:}
How would this respondent describe the state of the economy these days in the United States?
\\[1ex]
1. Very strong \ 2. Strong \ 3. Neither strong nor weak \ 4. Weak \ 5. Very weak
\\[1ex]
\textit{Reverse-coded:}
Not applicable
\\[1ex]
\textbf{Drug addiction V202348}
\\[1ex]
\textit{Original}:
Do you think the federal government should be doing more about the opioid drug addiction issue, should be doing less, or is it currently doing the right amount?
\\[1ex]
\underline{1. Should be doing more} \ 2. Should be doing less \ 3. Is doing the right amount
\\[1ex]
\textit{Reformulated:}
How would this respondent assess whether the federal government should be doing more about the opioid drug addiction issue, should be doing less, or is it currently doing the right amount?
\\[1ex]
1. Should be doing more \ 2. Should be doing less \ 3. Is doing the right amount
\\[1ex]
\textit{Reverse-coded:}
Not applicable
\\[1ex]
\textbf{Climate change V202332}
\\[1ex]
\textit{Original}:
How much, if at all, do you think climate change is currently affecting severe weather events or temperature patterns in the United States?
\\[1ex]
1. Not at all \ 2. A little \ 3. A moderate amount \ 4. A lot \ \underline{5. A great deal}
\\[1ex]
\textit{Reformulated:}
How would this respondent assess how much, if at all, climate change is currently affecting severe weather events or temperature patterns in the United States?
\\[1ex]
1. Not at all \ 2. A little \ 3. A moderate amount \ 4. A lot \ 5. A great deal
\\[1ex]
\textit{Reverse-coded:}
How much do you agree with the statement that climate change is unrelated to severe weather events or temperature patterns in the United States?
\\[1ex]
1. Not at all \ 2. A little \ 3. A moderate amount \ 4. A lot \ 5. A great deal
\\[1ex]
\textbf{Gay marriage V201416}
\\[1ex]
\textit{Original}:
Which comes closest to your view? You can just tell me the number of your choice.
\\[1ex]
\underline{1. Gay and lesbian couples should be allowed to legally marry.} \ 2. Gay and lesbian couples should be allowed to form civil unions but not legally marry. \ 3. There should be no legal recognition of gay or lesbian couples’ relationship.
\\[1ex]
\textit{Reformulated:}
Which comes closest to this respondent’s view?
\\[1ex]
1. Gay and lesbian couples should be allowed to legally marry. \ 2. Gay and lesbian couples should be allowed to form civil unions but not legally marry. \ 3. There should be no legal recognition of gay or lesbian couples’ relationship.
\\[1ex]
\textit{Reverse-coded:}
Which one disagrees the most with your view?
\\[1ex]
1. Gay and lesbian couples should be allowed to legally marry. \ 2. Gay and lesbian couples should be allowed to form civil unions but not legally marry. \ 3. There should be no legal recognition of gay or lesbian couples’ relationship.
\\[1ex]
\textbf{Refugee allowing V202234}
\\[1ex]
\textit{Original}:
Do you favor, oppose, or neither favor nor oppose allowing refugees who are fleeing war, persecution, or natural disasters in other countries to come to live in the U.S.?
\\[1ex]
\underline{1. Favor} \ 2. Oppose \ 3. Neither favor nor oppose
\\[1ex]
\textit{Reformulated:}
What is this respondent’s position on whether refugees who are fleeing war, persecution, or natural disasters in other countries should be allowed to come to live in the U.S.?
\\[1ex]
1. Should be allowed \ 2. Should not be allowed \ 3. No clear position
\\[1ex]
\textit{Reverse-coded:}
Do you favor, oppose, or neither favor nor oppose prohibiting refugees who are fleeing war, persecution, or natural disasters in other countries from coming to live in the U.S.?
\\[1ex]
1. Favor \ 2. Oppose \ 3. Neither favor nor oppose
\\[1ex]
\textbf{Health insurance V202378}
\\[1ex]
\textit{Original}:
Do you favor an increase, decrease, or no change in government spending to help people pay for health insurance when people cannot pay for it all themselves?
\\[1ex]
\underline{1. Increase} \ 2. Decrease \ 3. No change
\\[1ex]
\textit{Reformulated:}
How would this respondent assess if there should be an increase, decrease, or no change in government spending to help people pay for health insurance when people cannot pay for it all themselves?
\\[1ex]
1. Increase \ 2. Decrease \ 3. No change
\\[1ex]
\textit{Reverse-coded:}
Not applicable
\\[1ex]
\textbf{Gun regulation V202337}
\\[1ex]
\textit{Original}:
Do you think the federal government should make it more difficult for people to buy a gun than it is now, make it easier for people to buy a gun, or keep these rules about the same as they are now?
\\[1ex]
\underline{1. More difficult} \ 2. Easier \ 3. Keep these rules about the same
\\[1ex]
\textit{Reformulated:}
What is this respondent’s position on whether the federal government should make it more difficult for people to buy a gun than it is now, make it easier for people to buy a gun, or keep these rules about the same as they are now?
\\[1ex]
1. More difficult \ 2. Easier \ 3. Keep these rules about the same
\\[1ex]
\textit{Reverse-coded:}
Not applicable
\\[1ex]
\textbf{Income inequality V202257}
\\[1ex]
\textit{Original}:
Do you favor, oppose, or neither favor nor oppose the government trying to reduce the difference in incomes between the richest and poorest households?
\\[1ex]
\underline{1. Favor} \ 2. Oppose \ 3. Neither favor nor oppose
\\[1ex]
\textit{Reformulated:}
How would this respondent assess whether the government should be trying to reduce the difference in incomes between the richest and poorest households?
\\[1ex]
1. Should be trying \ 2. Should not be trying \ 3. Neither of these
\\[1ex]
\textit{Reverse-coded:}
Do you favor, oppose, or neither favor nor oppose the government to stop trying to reduce the difference in incomes between the richest and poorest households?
\\[1ex]
1. Favor \ 2. Oppose \ 3. Neither favor nor oppose

\subsection{All questions and prompt conditions of WVS Wave 7}\label{sec:all-questions-wvs}

\textbf{Gay Parenting Q36}
\\[1ex]
\textit{Original}:
How would you feel about the following statement? Do you agree or disagree with it? Homosexual couples are as good parents as other couples.
\\[1ex]
\underline{1. Agree strongly} 2. Agree 3. Neither agree nor disagree 4. Disagree 5. Disagree strongly
\\[1ex]
\textit{Reformulated:}
How would this respondent feel about the following statement? Does this respondent agree or disagree with it? Homosexual couples are as good parents as other couples.
\\[1ex]
1. Agree fully 2. Agree 3. Neither agree nor disagree 4. Disagree 5. Disagree fully
\\[1ex]
\textbf{Refugee Asylum Q125}
\\[1ex]
\textit{Original}:
From your point of view, what have been the effects of immigration on the development of this country? For the following statement about the effects of immigration, please, tell me whether you agree or disagree with it. Gives asylum to political refugees who are persecuted elsewhere.
\\[1ex]
\underline{2. Agree} 1. Hard to say 0. Disagree
\\[1ex]
\textit{Reformulated:}
For the following statement about the effects of immigration, does this respondent agree or disagree with it? Gives asylum to political refugees who are persecuted elsewhere.	
\\[1ex]
2. Agree 1. Hard to say 0. Disagree
\\[1ex]
\textbf{Income Equality Q106}
\\[1ex]
\textit{Original}:
Now I'd like you to tell me your view on an issue. How would you place your view on this scale? 1 means you agree completely with the statement on the left; 10 means you agree completely with the statement on the right; and if your views fall somewhere in between, you can choose any number in between.
\\[1ex]
1. Incomes should be made more equal 10. There should be greater incentives for individual effort
\\[1ex]
\textit{Reformulated:}
How would this respondent place their view on this scale? 1 means complete agreement with the statement on the left; 10 means complete agreement with the statement on the right; and if their views fall somewhere in between, any number in between is possible.
\\[1ex]
1. Incomes should be made more equal 10. There should be greater incentives for individual effort
\\[1ex]
\textbf{Maternal Employment Q28}
\\[1ex]
\textit{Original}:
For the following statement I read out, can you tell me how strongly you agree or disagree with it. Do you strongly agree, agree, disagree, or strongly disagree? When a mother works for pay, the children suffer.	
\\[1ex]
1. Strongly agree 2. Agree 3. Disagree \underline{4. Strongly disagree}
\\[1ex]
\textit{Reformulated:}
For the following statement, how strongly does this respondent agree or disagree with it? Does this respondent fully agree, agree, disagree, or fully disagree? When a mother works for pay, the children suffer.
\\[1ex]
1. Fully agree 2. Agree 3. Disagree 4. Fully disagree
\\[1ex]
\textbf{Environment vs Economy Q111}
\\[1ex]
\textit{Original}:
Here are two statements people sometimes make when discussing the environment and economic growth. Which of them comes closer to your own point of view?
\\[1ex]
\underline{1. Protecting the environment should be given priority, even if it causes slower economic growth and some} \underline{loss of jobs.} 2. Economic growth and creating jobs should be the top priority, even if the environment suffers to some extent. 3. Other answer
\\[1ex]
\textit{Reformulated:}
Here are two statements people sometimes make when discussing the environment and economic growth. Which of them comes closer to this respondent's point of view?
\\[1ex]
1. Protecting the environment should be given priority, even if it causes slower economic growth and some loss of jobs. 2. Economic growth and creating jobs should be the top priority, even if the environment suffers to some extent. 3. Other answer
\\[1ex]
\textbf{Immigration Impact Q121}
\\[1ex]
\textit{Original}:
Now we would like to know your opinion about the people from other countries who come to live in your country - the immigrants. How would you evaluate the impact of these people on the development of your country?	
\\[1ex]
\underline{5. Very good} 4. Quite good 3. Neither good nor bad 2. Quite bad 1. Very bad
\\[1ex]
\textit{Reformulated:} Now we would like to know this respondent's opinion about the people from other countries who come to live in their country - the immigrants. How would this respondent evaluate the impact of these people on the development of their country?
\\[1ex]
5. Highly beneficial 4. Beneficial 3. Neither beneficial nor detrimental 2. Detrimental 1. Highly detrimental
\\[1ex]
\textbf{Government Responsibility Q108}
\\[1ex]
\textit{Original}:
Now I'd like you to tell me your view on an issue. How would you place your view on this scale? 1 means you agree completely with the statement on the left; 10 means you agree completely with the statement on the right; and if your views fall somewhere in between, you can choose any number in between.	
\\[1ex]
1. Government should take more responsibility to ensure that everyone is provided for 10. People should take more responsibility to provide for themselves
\\[1ex]
\textit{Reformulated:}
How would this respondent place their view on this scale? 1 means complete agreement with the statement on the left; 10 means complete agreement with the statement on the right; and if their views fall somewhere in between, any number in between is possible.
\\[1ex]
1. Government should take more responsibility to ensure that everyone is provided for 10. People should take more responsibility to provide for themselves
\\[1ex]
\textbf{Immigration Policy Q130}
\\[1ex]
\textit{Original}:
How about people from other countries coming here to work. Which one of the following do you think the government should do?	
\\[1ex]
\underline{1. Let anyone come who wants to} 2. Let people come as long as there are jobs available 3. Place strict limits on the number of foreigners who can come here 4. Prohibit people coming here from other countries
\\[1ex]
\textit{Reformulated:} How about people from other countries coming here to work. Which one of the following does this respondent think the government should do?
\\[1ex]
1. Let anyone come who wants to 2. Let people come as long as there are jobs available 3. Place strict limits on the number of foreigners who can come here 4. Prohibit people coming here from other countries
\\[1ex]
\textbf{Government Surveillance Q198}
\\[1ex]
\textit{Original}:
Do you think that the your country’s government should or should not have the right to keep people under video surveillance in public areas?	
\\[1ex]
1. Definitely should have the right 2. Probably should have the right 3. Probably should not have the right \underline{4. Definitely should not have the right}
\\[1ex]
\textit{Reformulated:} Does this respondent think that the your country’s government should or should not have the right to keep people under video surveillance in public areas?
\\[1ex]
1. Definitely should have the right 2. Probably should have the right 3. Probably should not have the right 4. Definitely should not have the right
\\[1ex]
\textbf{Strong Leader Q235}
\\[1ex]
\textit{Original}:
I'm going to describe a type of political system and ask what you think about it as a way of governing this country. Would you say it is a very good, fairly good, fairly bad or very bad way of governing this country? Having a strong leader who does not have to bother with parliament and elections.	
\\[1ex]
1. Very good 2. Fairly good 3. Fairly bad \underline{4. Very bad}
\\[1ex]
\textit{Reformulated:} I'm going to describe a type of political system and ask what this respondent thinks about it as a way of governing this country. Would this respondent say it is a highly beneficial, beneficial, detrimental, or highly detrimental way of governing this country? Having a strong leader who does not have to bother with parliament and elections.
\\[1ex]
1. Highly beneficial 2. Beneficial 3. Detrimental 4. Highly detrimental

\subsection{Example of prompts}\label{sec:prompt-eg}
We provide two prompt examples of GPT-4.1-mini simulations for user responses to the multiple-choice health insurance question (ANES 2020), utilizing randomly sampled demographic variables. These examples illustrate a comparison under the Replicate and Reformulated question conditions.
\begin{tcolorbox}[colback=gray!0, colframe=black, boxrule=0.5pt, left=2pt,right=2pt,top=2pt,bottom=2pt,
  enlarge top by=-2pt,
  enlarge bottom by=-2pt]
\begin{center}
    \textbf{\underline{Replicate Condition}} 
\end{center}
\textbf{System} 
\\
Today is November 3, 2020. Racially, I am white. I like to discuss politics with my family and friends. Ideologically, I am moderate. Politically, I am a strong Republican. I attend church. I am 36 years old. I am a man.
\\[1.5ex]
\textbf{User}
\\
Question: \textit{Do you favor} an increase, decrease, or no change in government spending to help people pay for health insurance when people can’t pay for it all themselves? Answer choices: 1. Increase 2. Decrease 3. No change When answering, respond ONLY with a single number that corresponds to the option you choose. Do not include any additional text, punctuation or explanation. My answer is
\\[1.5ex]
\textbf{Output}
\\
1

\end{tcolorbox}

\begin{tcolorbox}[colback=gray!0, colframe=black, boxrule=0.5pt, left=2pt,right=2pt,top=2pt,bottom=2pt,
  enlarge top by=-2pt,
  enlarge bottom by=-2pt]
\begin{center}
    \textbf{\underline{Reformulated Condition}}
\end{center}
\textbf{System} 
\\
Today is November 3, 2020. Racially, \textit{the respondent} is white. The respondent likes to discuss politics with their family and friends. Ideologically, the respondent is slightly liberal. Politically, the respondent is an independent who leans Democratic. The respondent does not attend church. The respondent is 37 years old. The respondent is a woman. The respondent is somewhat interested in politics.
\\[2ex]
\textbf{User}
\\
Question: \textit{How would this respondent assess} if there should be an increase, decrease, or no change in government spending to help people pay for health insurance when people cannot pay for it all themselves? Answer choices: 1. Increase 2. Decrease 3. No change When answering, respond ONLY with a single number that corresponds to the option you choose. Do not include any additional text, punctuation or explanation. My answer is
\\[1.5ex]
\textbf{Output}
\\
1

\end{tcolorbox}

\end{document}